\documentclass{article}

\PassOptionsToPackage{numbers, compress}{natbib}

\usepackage[preprint]{neurips_2026}

\usepackage[textsize=tiny]{todonotes}
\usepackage{amsmath}
\usepackage{amssymb}
\usepackage{mathtools}
\usepackage{amsthm}
\usepackage{wrapfig}
\usepackage{enumitem}

\usepackage{microtype}
\usepackage{graphicx}
\usepackage{subcaption}
\usepackage{booktabs}
\usepackage{amsfonts}
\usepackage{booktabs}
\usepackage{siunitx}
\usepackage{stmaryrd}
\usepackage[hidelinks]{hyperref}

\usepackage{hyperref}

\usepackage[capitalize,noabbrev]{cleveref}

\usepackage[utf8]{inputenc} 
\usepackage[T1]{fontenc}    
\usepackage{hyperref}       
\usepackage{url}            
\usepackage{booktabs}       
\usepackage{amsfonts}       
\usepackage{nicefrac}       
\usepackage{microtype}      
\usepackage{xcolor}         

\title{\textit{MapFormers}: Self-Supervised Learning of Cognitive Maps with Input-Dependent Positional Encoding}

\author{%
    Victor Rambaud$^{1,2,3}$ \quad
    Salvador Mascarenhas$^{1,3}$ \quad
    Yair Lakretz$^{2,3}$ \\
    {}\\
    $^{1}$Institut Jean Nicod \quad
    $^{2}$LSCP \\
    $^{3}$Département d’Études Cognitives - ENS, EHESS, CNRS, PSL University \\
    \texttt{victor.rambaud@gmail.com, salvador.mascarenhas@ens.fr, yair.lakretz@gmail.com}
}

\begin{document}

\maketitle

\begin{abstract}
  A cognitive map is an internal model which encodes the abstract relationships among entities in the world, giving humans and animals the flexibility to adapt to new situations, with a strong out-of-distribution (OOD) generalization that current AI systems still do not possess. To bridge this gap, we introduce $\textit{MapFormers}$, new Transformer-based architectures, which can learn cognitive maps from observational data and perform path-integration without supervision. Cognitive maps are learned in the model by disentangling structural relationships in the inputs from their specific content, a property that can be achieved by updating position encodings with input-dependent matrices, built as exponentials of learned combinations of Lie-algebra generators. We developed two variants of $\textit{MapFormers}$ that unify absolute and relative positional encoding to model episodic (EM) and working memory (WM), respectively. We tested $\textit{MapFormers}$ on several formal tasks targeting distinct cognitive capacities, including gating, 2D navigation and nested hierarchies (Dyck Languages). Our results demonstrate that $\textit{MapFormers}$ significantly outperform current AI architectures, achieving near-perfect OOD generalization where standard models fail. Furthermore, we show that $\textit{MapFormers}$ are scalable; evaluations on naturalistic data yield perplexity improvements over baselines, suggesting that these principles extend to large-scale, real-world domains. These results are obtained through efficient parallel computation on commutative maps, though our models can also learn non-commutative cognitive maps via sequential path-integration. Overall, these results suggest that input-dependent matrices provide a critical structural bias, by disentangling abstract relations from content in order to drive robust OOD generalization.
\end{abstract}

\section{Introduction}    
\begin{wrapfigure}{r}{0.52\textwidth}  
    \vspace{-1.5em}
    \centering
    \includegraphics[width=1.\linewidth]{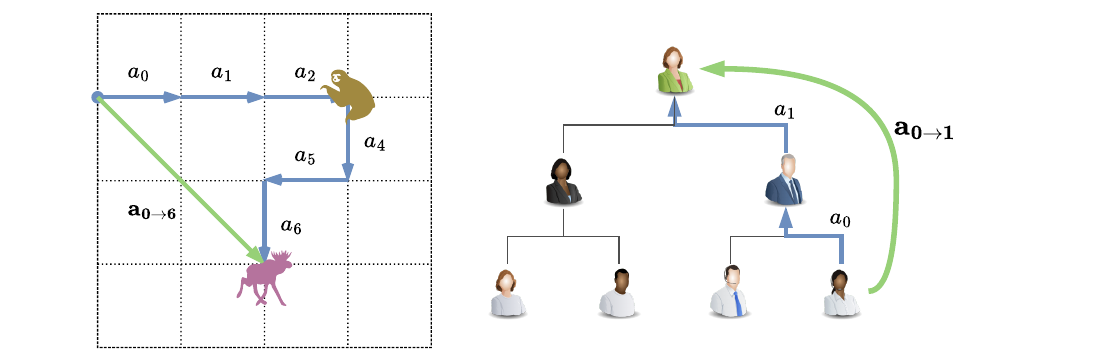}
    \caption{
        \textbf{Cognitive maps factorize content from position.} 
        Actions (blue) update position consistently across environments, and path integration composes them $a_{0\to t} = a_{t-1} \circ \cdots \circ a_0$ (green).
    }
    \label{fig:intro:PI}
    \vspace{-1.5em}
\end{wrapfigure}
Transformer-based models \cite{NIPS2017_3f5ee243} now underpin most modern AI systems, achieving impressive performance across a wide range of tasks. Yet, beneath their impressive performance, these models remain surprisingly brittle: when carefully tested, they often fail to generalize to out-of-distribution tasks that humans and animals can handle with ease \cite{nezhurina2024alice, zhang2024outofdistributiongeneralizationmultimodallarge}. This gap suggests that Transformers might still lack the necessary structural biases that allow biological brains to navigate unknown scenarios and reason through novel problems effectively.

A potential explanation for this gap lies in the ability to construct \textit{cognitive maps}~\cite{BEHRENS2018490}: internal models that separate the abstract, invariant \textit{structure} of an environment from its specific, variable \textit{content}. This factorization of \textit{what} from \textit{where} allows known structures to be applied to entirely new observations, supporting both spatial navigation (e.g., finding shortcuts or detours) and non-spatial reasoning (e.g., inferring family relations). Such different domains can be unified by viewing structure as a graph: nodes represent states and edges relationships between them—spatial or conceptual. In a world where observations are unbounded but the underlying structure remains stable, this factorization is fundamental.
To navigate a structure, an agent must learn the set of \textit{actions} that affect its state and, crucially, how they combine. This process is called \textit{path integration} (fig.~\ref{fig:intro:PI}, green arrows)—the process of integrating a sequence of actions—which is essential for tracking state changes and mental planning \cite{olaf2018}.
Path integration also enables to plan shortcuts or bypass obstacles by recognizing that distinct action sequences, such as $(\uparrow, \rightarrow, \rightarrow, \downarrow)$ and $(\rightarrow, \rightarrow)$, are functionally equivalent. Humans and animals are highly proficient at this type of relational composition \cite{Hafting2005, HOWARD20141331, Wehner1981}, a capability for robust out-of-distribution generalization that current AI systems still struggle to replicate \cite{vafa2024evaluatingworldmodelimplicit, kaesberg-etal-2025-sparc}.

Previous work made progress in modeling cognitive maps using RNNs with \textit{action-dependent} matrices \cite{Whittington2025, Whittington770495}, but these models do not scale since (1) their matrices must be multiplied sequentially, preventing parallel sequence processing, and (2) they require the underlying structure (e.g., the set of actions) to be specified in advance rather than learned. 

To address this, we introduce \textit{MapFormers}, a class of transformers that learn cognitive maps via next-token prediction and disentangle structure from content without supervision. For commutative maps, \textit{MapFormers} compute path integration \textit{in parallel}, enabling efficient scaling to more complex problems. We evaluate \textit{MapFormers} on three tasks probing distinct structural capacities -- navigation, gating, and nested hierarchies -- and show systematic generalization, including perfect extrapolation to longer and deeper nested structures than those seen during training. More broadly, we show that \textit{MapFormers} can scale to real-world problems, such as natural language, where we also observe improved performance over standard baselines.

We introduce two variants of \textit{MapFormers}, which unify relative rotary (RoPE; \cite{su2023roformerenhancedtransformerrotary}) and absolute positional embeddings, respectively as models of working and episodic memory. 
Finally, we show how our principled Lie-group formalism, applicable to both transformers and State Space Models (SSMs) \cite{gu2024mambalineartimesequencemodeling, dao2024transformersssmsgeneralizedmodels}, extends to non-commutative problems, albeit not in parallel. We validate this empirically against alternative input-dependent positional encoding methods, demonstrating the advantage of \textit{MapFormers}. Taken together, our results open new pathways toward AI systems that achieve systematic OOD generalization by capturing latent structures via path integration rather than relying on heuristics, while remaining scalable to real-world domains.
\section{Related Literature}

The Tolman-Eichenbaum Machine (TEM)~\cite{Whittington770495} introduced a set of spatial (navigation) and non-spatial (e.g., family trees) tasks that require learning a cognitive map, and proposed an RNN using \emph{action-dependent} matrices to update its position before querying a Hopfield memory module \cite{ramsauer2021hopfieldnetworksneed}. This framework was extended to model the prefrontal cortex in working memory~\cite{Whittington2025}, representing structure \emph{implicitly} by rotating the full neural activity into structured activity slots with action-dependent matrices. It was also shown that TEM can be seen as a Transformer (TEM-t) \cite{whittington2022relatingtransformersmodelsneural}, where positional embeddings represent position in a cognitive map. However, TEM-t lacks \textit{scalability}: it requires sequential updates of positional embeddings via action-dependent matrices, preventing parallel sequence processing and making it impractical for large-scale domains.

Other work has demonstrated the generalization power of sequence-dependent positional encoding in transformers \cite{zhu2025rethinkingaddressinglanguagemodels, golovneva2024contextualpositionencodinglearning, gopalakrishnan2025decouplingwhatwherepolar, yang2026pathattentionpositionencoding}, yielding superior results in length generalization and a selection of formal tasks, but their representation power is poorly understood. Specifically, little is known about what kind of structural features these models can or cannot learn. Moreover, no connections have been made to cognitive maps, which is one of the main goal of this paper.

Another line of work, Clone-Structured Causal Graphs (CSCGs) \cite{George2021, doi:10.1126/sciadv.adm8470}, learn cognitive maps through discrete latent states (clones). While promising, these models are not parallelizable and are trained via Expectation-Maximization (EM), making them incompatible with standard next-token prediction and cross-entropy training. Moreover, their discrete formulation prevents them from learning smooth maps or modeling continuous transformations such as head direction \cite{Taube436} or mental rotations \cite{doi:10.1126/science.171.3972.701}.

\section{Theory and Models}



\subsection{Learning Cognitive Maps and Path-Integration using Lie-group Theory}\label{ssec:additive_PI}



We define a cognitive map as a multirelational graph $G = (N, E, \mathcal{R})$, where nodes (\textit{i.e.} states) are connected by edges $E$, typed by a set of cardinal relations $\mathcal{R}$—such as cardinal directions on a grid or family relations in a family tree. Each node $n\in N = (\mathcal{P}, \mathcal{O})$ is characterized by both its \emph{position} $p\in \mathcal{P}$ and an associated \emph{content} (or observation) $o \in \mathcal{O}$, independent of the map's topology—such that the same map can host arbitrary content assignments (fig.\ref{fig:intro:PI}). An agent navigating this map occupies a position $p_t \in N$, 
and updates it using a transition function (\textit{i.e.} an action) $a_t: N \to N$. To efficiently learn the map, the agent must recover the hidden topology by aligning each action with its corresponding relation. Specifically, for any relation $r_{j \to i} \in R$ connecting $n_j$ to $n_i$, the agent needs to learn an action $a_t \in \mathcal{A}$ that satisfies the condition $p_{t+1} = a_t(p_t=n_j) = n_i$.

Once the map is learned, the agent can perform \textit{path integration} (fig.~\ref{fig:intro:PI}, green arrows) by composing a sequence of actions. The resulting position $p_i$ reached from an initial position $p_j$ through a sequence of actions is defined by: 
\[p_i = a_{j \to i} p_j = (a_{i-1} \circ \dotsb \circ a_j)p_j\]
where we denoted integrated actions as $a_{j \to i} = a_{i-1} \circ \dotsb \circ a_j$. Note that position is always defined by integrated action $a_{j \to i}$, representing either relative or, when referenced to the origin $p_0$ ($j=0$), absolute position.

In previous work \cite{Whittington2025, Whittington770495, whittington2022relatingtransformersmodelsneural}, cognitive maps were modeled in RNNs with a \emph{fixed} set of \textit{action-dependent} recurrent weights, to compute integrated actions $a_{j\to i}$ via matrix multiplication: $\mathbf{W_{j\to i}}:=\prod_{s=j}^{i-1} W_{a_s}$. We extend this line of work by dynamically constructing input-dependent matrices as exponentials of learned Lie-algebra generators (sec.~\ref{ssec:groups}).
Rather than hard-coding each action matrix separately, we learn generators of a Lie algebra that implicitly represent all possible actions. This reduces the inductive bias from the number of actions to the algebra's structural complexity. Moreover, for abelian groups, we can replace the costly sequential product with \textit{linear} accumulation in the Lie algebra, followed by a single exponential map (sec.~\ref{ssec:algebra}).
For example, in the case of rotations, action-dependent 
matrices are rotation matrices $R_\theta$ that lie on the manifold of the special 
orthogonal group $\mathrm{SO}(2)$. Its Lie algebra $\mathfrak{so}(2)$ is one-dimensional, 
possessing a single generator matrix $S$ (the unit skew-symmetric matrix; $S^T = -S$). Any matrix in $\mathfrak{so}(2)$ can 
be expressed as $A = \omega S$, where $\omega$ is the angular velocity. A finite 
rotation $R_\theta$ can then be generated by integrating the infinitesimal 
angular velocities $\omega$ over a duration $\Delta_t$. In this case, path integration can be computed at the level of the Lie algebra by summing the integration 
durations $\Delta_t$:
\begin{equation}\label{eq:additive_PI}
W_{j \to i} := \prod_{t=j}^{i-1}\exp \left( \omega \Delta_t S \right) = \exp \left( \omega S\sum_{t=j}^{i-1} \Delta_t \right) := R(\theta_{j \to i}),
\end{equation}
Most of the computational cost thus boils down to a simple sum $\sum_{t=j}^{i-1} \Delta_t$. 
Computing the exponential map needs to be done only once, after the summation.
Furthermore, in the case of rotation, computing the exponential is not even necessary, as the closed-form 
solution for the exponential map is already known to be the standard rotation matrix 
$R_\theta$ (sec.~\ref{ssec:groups}, \ref{ssec:algebra}).

In practice, in the case of navigation on a finite 2D grid, \textit{i.e.} a torus, action-dependent matrices are $2\times 2$ block-diagonal rotations, where each block lies in $\mathrm{SO}(2)$, defined by a constant \textit{angular velocity} $\omega_b$. They represent rotations at different scales, and are learned by the model during training, adapting themselves to the size of the grid --- e.g., larger grids require a smaller $\omega$ to avoid wrapping around. Path integration can be computed along each of the $n_b$ blocks, in parallel.

Interestingly, there is a connection between the exponential map in Lie group formulations of cognitive maps and its role in the discretization of continuous State Space Models (SSMs). In SSMs, the exponential arises when discretizing continuous dynamics with step size $\Delta_t$ \cite{gu2024mambalineartimesequencemodeling}, mapping a linear operator in continuous time to its discrete-time counterpart. This parallels the Lie algebra–group mapping, where exponentiation maps infinitesimal generators to finite transformations. The continuous and discrete SSM equations (eq.~\ref{eq:ssms_cont} \& \ref{eq:ssm_discrete}) can thus be interpreted as operating in a 1-dimensional Lie algebra and Lie group, respectively. Because Linear Transformers are equivalent to SSMs \cite{dao2024transformersssmsgeneralizedmodels}, in this work, we mainly focus on $\mathrm{SO}(2)^k$, as it is compact and 1-dimensional—matching the SSM framework with a single generator. This ensures bounded neural activity and commutativity, hence linear path-integration, and a closed-form solution of the exponential map.

We further develop in sec.~\ref{ssec:PImba}, \ref{ssec:pimba_EMWM} \& \ref{ssec:map_are_ssms} how Lie-group theory enables the learning of cognitive maps and its connection to both SSMs and Transformers. Finally, the framework can be extended to non-commutative cognitive maps (sec.~\ref{sec:non_commute}) through the introduction of multiple generator matrices, at the cost of losing the additive path-integration introduced in eq.~\ref{eq:additive_PI}.

\subsection{\textit{MapFormers}: Path-Integrating Transformers for learning Cognitive Maps}\label{sec:mapformers}
\begin{figure*}[hbt!]
    \centering
    \begin{subfigure}[b]{0.45\textwidth}
        \centering
        \includegraphics[width=\textwidth]{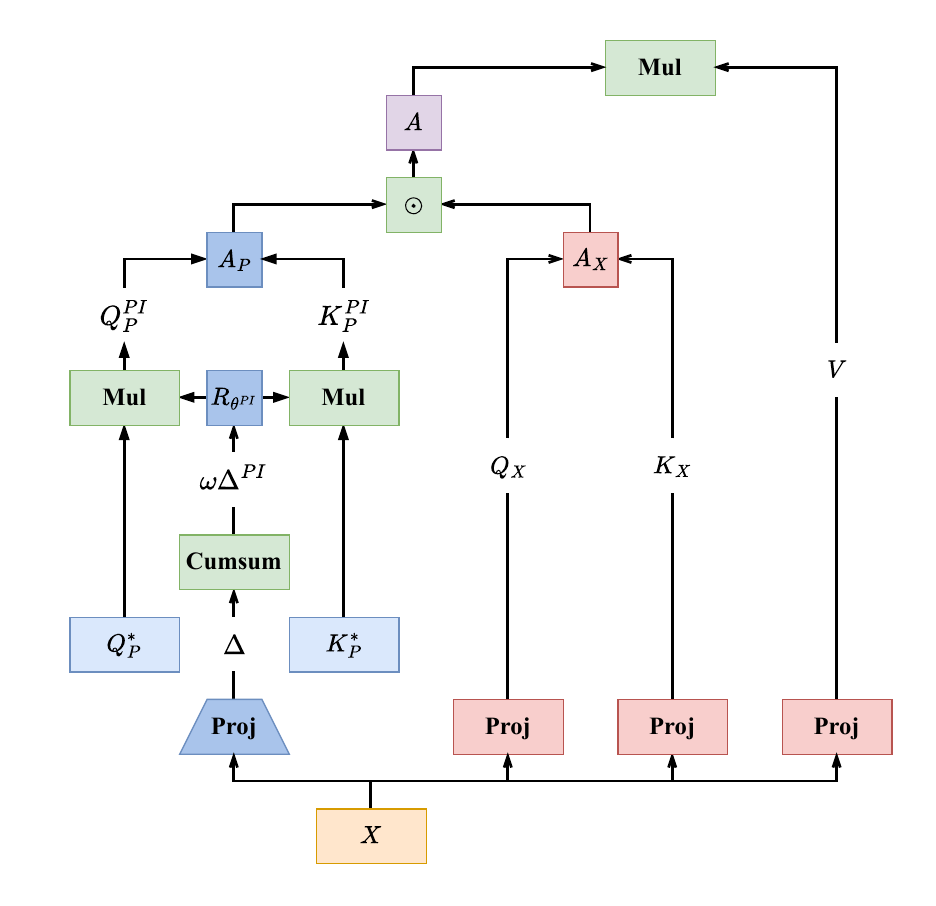}
        \caption{}
        \label{fig:arch:EM}
    \end{subfigure}
    \hfill
    \begin{subfigure}[b]{0.4\textwidth}
        \centering
        \includegraphics[width=\textwidth]{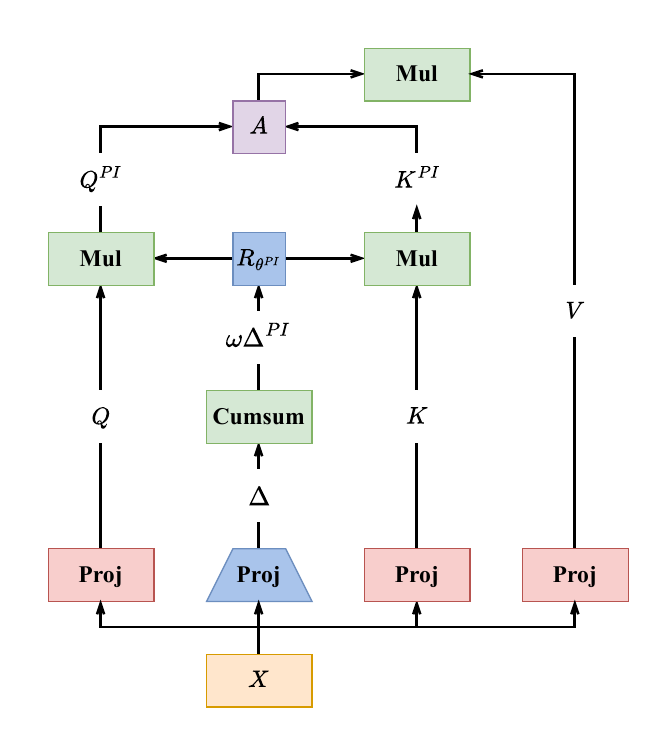}
        \caption{}
        \label{fig:arch:WM}
    \end{subfigure}
    \caption{
        \textbf{Disentangling Structure and Content for Path Integration in Episodic and Working Memory Transformers.} Overview of our \textit{MapFormers} architectures for Episodic (EM; panel a) and Working Memory (WM; panel b). In both models, rotation angle $\theta=\omega\Delta$ is obtained via a low-rank projection of input $X$, before computing a cumsum along the temporal dimension to perform \textit{path integration} (PI) in parallel: $\theta^{PI}=\omega \Delta^{PI}=\omega \cdot \mathrm{cumsum}_t(\Delta)$. \textit{Input-dependent}, block-diagonal, rotation matrices $R_{\theta^{PI}}$ update the position on the cognitive map. \textbf{(a)} \textbf{Episodic Memory \textit{MapFormer}} (\textbf{\textit{Map}EM}) with two parallel attentions $A_X$ and $A_P$. The former captures raw similarities between external inputs and the latter encodes the problem's structure, by path-integrating an initial position $P^*$ \textbf{(b)} \textbf{Working Memory \textit{MapFormer}} (\textbf{\textit{Map}WM}) with relative positional encoding, where $Q, K$ are rotated via input-dependent matrix $R_{\theta^{PI}}$. In \textit{Map}WM, $\theta$ is learned, but keeping it fixed gives exactly RoPE.
    }
    \label{fig:arch}
\end{figure*}
In this section, we introduce two new Transformer-based architectures, which we call \textit{\textbf{MapFormers}}, designed to learn path integration and, consequently, cognitive maps. The fundamental mechanism in both \textit{MapFormers} is the use of \textit{input-dependent matrices} to update positional encoding as in sec.~\ref{ssec:additive_PI}, which, as explained, represents position in a \emph{context-dependent} cognitive map instead of ordinal order used in standard Transformers.

Transformers achieve parallel and large-scale  sequence processing via the attention mechanism. An input sequence $X\in\mathbb{R}^{l\times d}$ is projected into keys, queries and values, $Q, K, V \in \mathbb{R}^{l\times d}$, and token mixing is achieved via a weighted sum of the values based on the key-query similarity matrix $A\in\mathbb{R}^{l\times l}$:
\[
    \textbf{Att}(Q, K, V) = \mathrm{softmax}\left(\frac{QK^T}{\sqrt{d}}\right)V
\]
Because the attention mechanism is permutation-invariant, positional information must be provided, either through absolute positional embeddings added to the input \cite{NIPS2017_3f5ee243}, or relative positional encoding to model the distance between keys and queries \cite{su2023roformerenhancedtransformerrotary}.


The two proposed architectures (fig.~\ref{fig:arch:EM} \& ~\ref{fig:arch:WM}) follow this distinction: the first architecture extracts and represents structural information \emph{separately} from the content, using absolute positional embeddings to form a cognitive map; this is akin to a model of episodic memory \cite{Whittington770495}. The second architecture, conversely, represents structure \emph{implicitly} by rotating neural activity into distinct subspaces, akin to the models of working memory in \cite{Whittington2025}.

To learn a cognitive map, the model has to identify that action tokens ($s_t=a_t$) update the internal position $p_t$, while ensuring that observations ($s_t=o_t$) leave the structure untouched. This is achieved through a two-stage projection process: first, an inner projection $W_{\Delta}^{\mathrm{in}}\in \mathbb{R}^{d\times r}$ maps each token representation $x_t \in \mathbb{R}^d$ to a low-dimensional action representation ${\Delta}^{\mathrm{in}}_t\in \mathbb{R}^r$ (where $r=1,2$ in 1D/2D navigation, capturing the actual motion to perform). Then, an outer projection $W_{\Delta}^{\mathrm{out}}\in \mathbb{R}^{r\times n_b}$ projects the inner action representation to a per-block integration duration:
\begin{equation}\label{eq:lowrank_proj}
    \Delta_t^{\mathrm{in}} = W_{\Delta}^{\mathrm{in}} x_t\in \mathbb{R}^r;\quad\Delta_t =W_{\Delta}^{\mathrm{out}}\Delta_t^{\mathrm{in}} = W_{\Delta}^{\mathrm{out}} \left(W_{\Delta}^{\mathrm{in}} x_t\right) \in \mathbb{R}^{n_b}
\end{equation}
The final rotation angles are then computed as $\theta_t = \omega \odot \Delta_t$, where $\omega\in \mathbb{R}^{n_b}$ is a learnable vector controlling the \textit{angular velocity} on each block. Path-integration is finally computed in parallel along the temporal dimension via a cumulative summation: $\theta_{0\to t}:= \mathrm{cumsum}_t(\theta_t)$, allowing to build the final path-integrated rotation matrix $R_{\theta^{PI}}:=\{R_{\theta_0\to t}\}_{t\leq l}$ (fig.~\ref{fig:arch}). See sections~\ref{sec:implem_details} and \ref{ssec:freqs} for implementation details.



\paragraph{\textit{Map}EM: \textit{MapFormers} with Absolute Positional Embeddings as Models of Episodic Memory}

The first architecture of \textit{MapFormers} makes use of absolute positional embeddings to learn a cognitive map (fig.~\ref{fig:arch:EM}), inspired by the links established between TEM and transformers \cite{whittington2022relatingtransformersmodelsneural}.
Once $R_{\theta^{PI}}$ is computed, one only needs to learn an embedding $p_0\in \mathbb{R}^d$, representing the \emph{origin} of the cognitive map, such that:
\[P:=R_{\theta^{PI}}P^*;\quad P^*=[p_0,\dots,p_0]_{l}\in \mathbb{R}^{l\times d}\]
Attention is computed on $Q_G:=\{q^g_t = \mathrm{vec}(q_t^\top.p_t)\}_{t\leq l}\in \mathbb{R}^{l\times d^2}$ and $K_G$, the conjunctions of \textbf{what} has been seen and \textbf{where}. This can be reframed as the element-wise product of content and position attentions $A_X=Q_XK_X^T/\sqrt{d}$ and $A_P=P.P^T/\sqrt{d}$, computed separately (still in parallel by concatenating the heads, see sec.~\ref{sec:implem_details}), such that $A_P$ acts as an attention mask on $A_X$:
\begin{equation}\label{eq:absolute_pe}
    \textbf{Att}(Q_g, K_g, V) = \mathrm{softmax}\left(\mathbf{A}_{X}\odot \mathbf{A}_{P}\right) V
\end{equation}
This allows modeling Episodic Memory in a transformer that we call \textit{MapFormer}-EM (or \textit{Map}EM, in short, see sec.~\ref{ssec:pimba_EMWM} for details on EM models).

\paragraph{\textit{Map}WM: \textit{MapFormers} with Relative Positional Embeddings as Models of Working Memory}

In \cite{Whittington2025}, Working Memory RNNs represent structure \emph{implicitly} by shifting neural activity into structured activity slots with action-dependent matrices. This directly models the logic of RoPE \cite{su2023roformerenhancedtransformerrotary}, where a fixed rotation matrix modeling relative distance is directly applied on the keys and queries, \textit{i.e.} the state. In \textit{MapFormers}-WM (\textit{Map}WM;~\ref{fig:arch:WM})), keys and queries are rotated by $R_{\theta^{PI}}$, which encodes their relative distance in the cognitive map. We further formalize the links between \textit{Map}WM and models of working memory in sec.~\ref{ssec:map_are_ssms}.

\section{Experimental Setup}
\subsection{Tasks}\label{ssec:task}
We evaluate our models on three formal next-token prediction tasks, each targeting an important cognitive ability: gating, navigating a cognitive map, and handling nested hierarchies. Each task admits a hidden structure that can be manipulated to create challenging OOD samples, implying that generalization is only possible by explicitly learning the underlying structure rather than relying on heuristics. 
Solving all three would demonstrate that the same mechanism initially developed for cognitive maps also supports a broader range of formal tasks. More details about data generation for each of these tasks can be found in sections \ref{ssec:selec_copy}, \ref{ssec:navi_detail} \& \ref{ssec:dyck_explained} of the appendix.

\textbf{Selective Copy } Introduced in Mamba~\cite{gu2024mambalineartimesequencemodeling} and used in CoPE~\cite{golovneva2024contextualpositionencodinglearning}, this task requires the model to copy a sequence while ignoring a blank distractor token $B$: $CFCFEABBCBF \to CFCFEACF$. Solving it requires learning a gating mechanism to dynamically filter out distractors—something which transformers with static positional encoding like RoPE~\cite{su2023roformerenhancedtransformerrotary} cannot do.


\textbf{Forced-Navigation Task } Following \cite{Whittington770495}, we cast navigation as next-token prediction— \textit{i.e.} predicting the upcoming observation. The model receives a sequence of alternating tokens $s = \left(a_1, o_1, \dotsc, a_T, o_T\right)$, where $a_t$
is an action and $o_t$ an observation sampled from a finite set of $K$ objects, plus an empty token $B$ when locations are empty, with probability $p_{empty}$. Crucially, the model is not told which tokens $s_t$ are actions and which are observations. Since observations are sampled randomly, the only way to achieve perfect accuracy when revisiting a location is for the model to recover the latent topology of the environment—i.e., to build a cognitive map by inferring which tokens act as transitions (actions in 2D: $\left\{ \uparrow, \downarrow, \rightarrow, \leftarrow\right\}$) and how they compose.

\textbf{Dyck-2 valid continuations } Dyck-2 is a context-free language over two bracket types, $()$ and $[]$, designed to test a model's ability to handle nested hierarchies, a core feature of human language -- By the Chomsky–Schützenberger theorem \cite{CHOMSKY1963118}, any context-free language can be constructed from Dyck languages via homomorphisms and intersections with regular languages, making them a central testbed for hierarchical generalization. To predict the next token, the model must identify the \emph{valid continuations}, which always include opening either bracket or closing the last opened one. This requires the model to track the stack of unclosed brackets through the sequence, which can be emulated via memory-augmented RNNs \cite{suzgun2019memoryaugmentedrecurrentneuralnetworks}. Conversely, transformers and standard LSTMs fail to generalize to longer and deeper sequences unseen during training \cite{bhattamishra-etal-2020-practical}.

\subsection{Baseline models}\label{ssec:baselines}

In all experiments, we restricted ourselves to Transformer-based models, since our goal is to introduce a generic architecture, trainable at scale. This rules out previous actionable models like TEM \cite{Whittington2025, Whittington770495}, since they are not parallelizable on commutative maps, and pre-suppose what the actions should be instead of discovering them. We also rule-out CSCGs \cite{George2021, doi:10.1126/sciadv.adm8470}, because they are neither parallelizable nor trainable via cross-entropy and next-token prediction. Moreover, since Transformers and SSMs are related \cite{dao2024transformersssmsgeneralizedmodels} and because our Lie-group formalism also applies to SSMs (sec.~\ref{ssec:PImba} \& \ref{ssec:pimba_EMWM}), we removed SSM-based models from our analysis because some architectural variations irrelevant to our Lie-group formalism might alter our empirical results. Nevertheless, we show in sec.~\ref{sec:limba} that the gating matrix of Mamba prevents it from learning a cognitive map.

We therefore focus our analysis on transformer architectures with \textit{input-dependent} positional encoding, known to generalize well on structured problems, apart from \textbf{RoPE} \cite{su2023roformerenhancedtransformerrotary}, the only architecture in our analysis without any bias for structure learning. Our baselines with \textit{input-dependent} positional encoding are \textbf{TAPE} \cite{zhu2025rethinkingaddressinglanguagemodels}, \textbf{CoPE} \cite{golovneva2024contextualpositionencodinglearning} and \textbf{PathAtt} \cite{yang2026pathattentionpositionencoding}, which will be compared to \textbf{\textit{Map}WM} and \textbf{\textit{Map}EM}. As controls, we introduce variations of our architectures: \textbf{\textit{Map}EM}-s/o/os that use structure, observation, or both in eq.~\ref{eq:absolute_pe} to validate the role that both observation and structure play in distinct tasks. In \textbf{\textit{Map}WM}-r, we ablate the rank $r \in \{1,2\}$ of the inner projection $W^{\text{in}}_\Delta$ (eq.~\ref{eq:lowrank_proj}) to verify that the rank of this projection does control movement in the cognitive map. Further details about each architecture are given in sec.~\ref{ssec:baseline_models}.
\section{Results}

\subsection{\textit{MapFormers} and other Path-Integrating baselines solve the Selective-Copy Task}\label{ssec:selective_copy}

\begin{wraptable}{r}{0.52\textwidth}
    \vspace{-12pt}
    \centering
    \small
    \setlength{\tabcolsep}{3pt}
    \begin{tabular}{
        l
        S[table-format=1.2(1)]
        S[table-format=1.2(1)]
        S[table-format=1.2(1)]
    }
    \toprule
    & \multicolumn{1}{c}{\textbf{IID}}
    & \multicolumn{1}{c}{\textbf{OOD Dense}}
    & \multicolumn{1}{c}{\textbf{OOD Sparse}} \\
    \midrule
        \textit{Map}EM-o  & {$0.11_{\pm0.02}$} & {$0.08_{\pm0.03}$} & {$0.07_{\pm0.02}$} \\
        \textit{Map}EM-s  & {$0.10_{\pm0.01}$} & {$0.11_{\pm0.01}$} & {$0.10_{\pm0.01}$} \\
        \midrule
        RoPE    & {$\textbf{1.00}_{\pm0.00}$} & {$0.48_{\pm0.04}$} & {$0.22_{\pm0.05}$} \\
        TAPE    & {$\textbf{1.00}_{\pm0.00}$} & {$0.88_{\pm0.05}$} & {$0.10_{\pm0.07}$} \\
        CoPE    & {$\textbf{1.00}_{\pm0.00}$} & {$\textbf{1.00}_{\pm0.00}$} & {$\textbf{1.00}_{\pm0.00}$} \\
        PathAtt & {$\textbf{1.00}_{\pm0.00}$} & {$\textbf{1.00}_{\pm0.00}$} & {$\textbf{1.00}_{\pm0.00}$} \\
        \midrule
        \textbf{\textit{Map}WM}    & {$\textbf{1.00}_{\pm0.00}$} & {$\textbf{1.00}_{\pm0.00}$} & {$\textbf{1.00}_{\pm0.00}$} \\
        \textbf{\textit{Map}EM-os} & {$\textbf{1.00}_{\pm0.00}$} & {$\textbf{1.00}_{\pm0.00}$} & {$\textbf{1.00}_{\pm0.00}$} \\
        \bottomrule
    \end{tabular}
    \caption{\textbf{Selective copy accuracy.} IID: 128/128 blank/non-blank. OOD dense: 64/128,OOD sparse: 256/128.}
    \label{tab:selective_copy}
    \vspace{-12pt}
\end{wraptable}

We start with the selective-copy task to test whether \textit{MapFormers} can learn a gating mechanism required to ignore distractors. Following \cite{golovneva2024contextualpositionencodinglearning}, we train 2-layer, 2-head transformers ($h=64$) on sequences of 128 non-blank tokens (from $K=16$ objects) interleaved with 128 blanks. OOD samples only vary the amount of blank tokens $B$: 64 for OOD Dense and 256 for OOD Sparse. As we can see in tab.~\ref{tab:selective_copy}, only CoPE, PathAtt, and \textit{MapFormers}—all three relying on a form of path integration—generalize OOD. TAPE improves performances in shorter sequences over RoPE, by learning a form of position interpolation, but performances deteriorate on longer sequences. Since solving this tasks involves understanding its structure (the tokens to ignore) and its semantic content (the tokens to copy), \textit{Map}EM-s/\textit{Map}EM-o, which relies on structure/observation only, fail.

\subsection{\textit{MapFormers} Solve the Grid Navigation Tasks but not Baseline Models}\label{ssec:grid_nav}

\begin{table}[h]
    \centering
    \label{tab:combined_tasks}
    \begin{tabular}{
        l |
        S[table-format=2.2(1)] S[table-format=2.2(1)] S[table-format=2.2(1)] |
        S[table-format=2.2(1)] S[table-format=2.2(1)] S[table-format=2.2(1)]
    } 
        \toprule
        \textbf{} & \multicolumn{3}{c|}{\textbf{1D navigation}} & \multicolumn{3}{c}{\textbf{2D Navigation}} \\
        \cmidrule(lr){2-4} \cmidrule(lr){5-7} 
        \textbf{} & \textbf{IID} & \multicolumn{2}{c|}{\textbf{OOD}} & \textbf{IID} & \multicolumn{2}{c}{\textbf{OOD}}\\
        \cmidrule(lr){3-4} \cmidrule(lr){6-7} 
        \textbf{} & \textbf{} & \textbf{D} & \textbf{S} & \textbf{} & \textbf{D} & \textbf{S} \\
        \midrule
        \textit{Map}EM-o & {$0.14_{\pm0.03}$} & {$0.20_{\pm0.01}$} & {$0.12_{\pm0.03}$} & {$0.16_{\pm0.02}$} & {$0.08_{\pm0.01}$} & {$0.08_{\pm0.01}$} \\
        \midrule
        RoPE (1L) & {$0.25_{\pm 0.03}$} & {$0.40_{\pm 0.05}$} & {$0.16_{\pm 0.02}$} & {$0.46_{\pm 0.01}$} & {$0.52_{\pm 0.02}$} & {$0.39_{\pm 0.01}$} \\
        RoPE (4L) & {$0.95_{\pm 0.01}$} & {$0.97_{\pm 0.01}$} & {$0.70_{\pm 0.03}$} & {$0.82_{\pm 0.05}$} & {$0.87_{\pm 0.03}$} & {$0.68_{\pm 0.03}$} \\
        TAPE (1L) & {$0.23_{\pm 0.01}$} & {$0.40_{\pm 0.02}$} & {$0.15_{\pm 0.03}$} & {$0.37_{\pm 0.01}$} & {$0.45_{\pm 0.02}$} & {$0.35_{\pm 0.02}$} \\
        TAPE (4L) & {$0.93_{\pm 0.02}$} & {$0.96_{\pm 0.03}$} & {$0.67_{\pm 0.01}$} & {$0.65_{\pm 0.02}$} & {$0.68_{\pm 0.01}$} & {$0.53_{\pm 0.02}$} \\
        CoPE (1L) & {$0.63_{\pm 0.21}$} & {$0.71_{\pm 0.21}$} & {$0.47_{\pm 0.22}$} & {$0.65_{\pm 0.03}$} & {$0.71_{\pm 0.04}$} & {$0.56_{\pm 0.04}$} \\
        CoPE (4L) & {$0.88_{\pm 0.01}$} & {$0.91_{\pm 0.02}$} & {$0.65_{\pm 0.05}$} & {$0.76_{\pm 0.01}$} & {$0.86_{\pm 0.01}$} & {$0.70_{\pm 0.01}$} \\
        PathAtt (1L) & {$\textbf{0.99}_{\pm0.00}$} & {$\textbf{0.99}_{\pm0.00}$} & {$\textbf{0.98}_{\pm0.00}$} & {$0.64_{\pm 0.04}$} & {$0.70_{\pm 0.05}$} & {$0.53_{\pm 0.05}$} \\
        \midrule
        \textbf{\textit{Map}WM-r1} & {$\textbf{1.00}_{\pm0.00}$} & {$\textbf{1.00}_{\pm0.00}$} & {$\textbf{1.00}_{\pm0.00}$} & {$0.66_{\pm 0.02}$} & {$0.74_{\pm 0.01}$} & {$0.62_{\pm 0.03}$} \\
        \textbf{\textit{Map}WM-r2} & {$\textbf{1.00}_{\pm0.00}$} & {$\textbf{1.00}_{\pm0.00}$} & {$\textbf{1.00}_{\pm0.00}$} & {$\textbf{1.00}_{\pm0.00}$} & {$\textbf{1.00}_{\pm0.00}$} & {$\textbf{0.99}_{\pm0.01}$} \\
        \textbf{\textit{Map}EM-os} & {$\textbf{1.00}_{\pm0.00}$} & {$\textbf{1.00}_{\pm0.00}$} & {$\textbf{1.00}_{\pm0.00}$} & {$\textbf{1.00}_{\pm0.00}$} & {$\textbf{1.00}_{\pm0.00}$} & {$\textbf{1.00}_{\pm0.00}$} \\
        \textbf{\textit{Map}EM-s} & {$\textbf{1.00}_{\pm0.00}$} & {$\textbf{1.00}_{\pm0.00}$} & {$\textbf{1.00}_{\pm0.00}$} & {$\textbf{1.00}_{\pm0.00}$} & {$\textbf{1.00}_{\pm0.00}$} & {$\textbf{1.00}_{\pm0.00}$} \\
        \bottomrule
    \end{tabular}
    \vspace{5pt}
    \caption{\textbf{1D-2D grid navigation accuracy.} IID: sequence length 128, grid width 64, $p_{empty}=0.5$. OOD-dense (D): 64/32/0.2. OOD-sparse (S): 256/128/0.8. We use single layer models with one head ($h=64$), or when specified, 2 layers and 2 heads. \textit{Map}WM-r1/\textit{Map}WM-r2 use a rank-1/2 matrix for encoding movement on the cognitive map $\Delta^{\mathrm{in}}_t$, while \textit{Map}EM-o computes attention on content only, without positional encoding.}
    \label{tab:navigation_perf}
    \vspace{-0.5cm}
\end{table}

For 1D-2D navigation, observations are sampled from $K=16$ objects and actions from the cardinal directions (2 in 1D, 4 in 2D). We train single layer, single head models ($h=64$) via next token prediction, but compute accuracy only when coming back to previously visited locations, which are the only tokens predictable with certainty. For additional baselines, we increase model's depth up to 4 layers (and 2 heads). Additional details about training are given in sec.~\ref{ssec:navi_detail} of the appendix.

Results in tab.~\ref{tab:navigation_perf} confirm our theory, with baselines failing for distinct but predictable reasons: because observations are random, the model must learn to disentangle content from structure and update its position from incoming actions, something that RoPE cannot do. Moreover, the model must path-integrate its position—ruling out TAPE—and be able to represent an action (moving forward) and its inverse (moving backwards), preventing CoPE from solving the task, even in 1D. The structural bias for path-integration should also match the underlying topology, limiting PathAtt's ability to solve the task in 2D. Finally, as expected, the rank $r=1,2$ of the integration time $\Delta_t\in\mathbb{R}^r$ directly influences our model's ability to learn a cognitive map—since \textit{Map}WM-r1 generalizes in 1D but not in 2D—while models relying on observation alone (\textit{Map}EM-o) perform worse than RoPE.

\textbf{EM scale better than WM models in recall } We further show in sec.~\ref{ssec:recall_scaling} the superiority of EM models (absolue PE) over WM (relative PE) on recall tasks due to their dedicated neural population for encoding position.

\textbf{High-dimensional and non-commutative navigation } We extend the analysis to higher-dimensional and non-abelian navigation: on commutative maps, each independent dimension is encoded as a circle, so the same mechanism can be used but requires more neurons to encode each axis efficiently (sec.~\ref{sec:3D}); \textit{MapFormers} can also learn non-commutative maps such as family trees, but require multiple generators per diagonal block, making path-integration non-parallelizable (sec~\ref{sec:non_commute}).


\subsection{MapFormers learn to generalize to longer and deeper Dyck-2 sequences}\label{ssec:dyck_results}

\begin{figure*}[ht!]
    \centering

    \vspace{1em} 

    \begin{subfigure}[c]{0.32\textwidth}
        \centering
        \includegraphics[width=\textwidth]{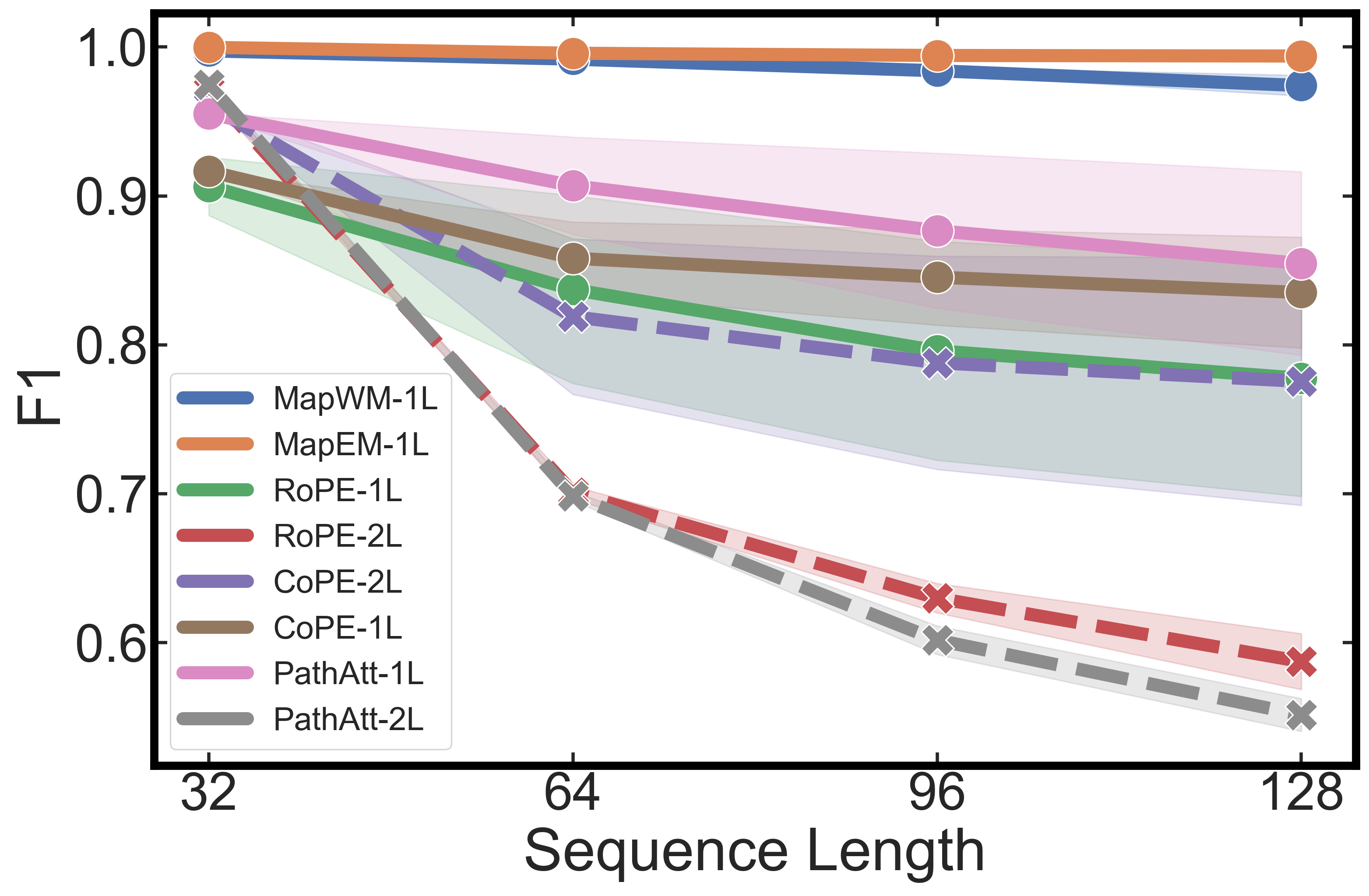}
        \caption{}
        \label{fig:dyck_gen:len}
    \end{subfigure}
    \begin{subfigure}[c]{0.32\textwidth}
        \centering
        \includegraphics[width=\textwidth]{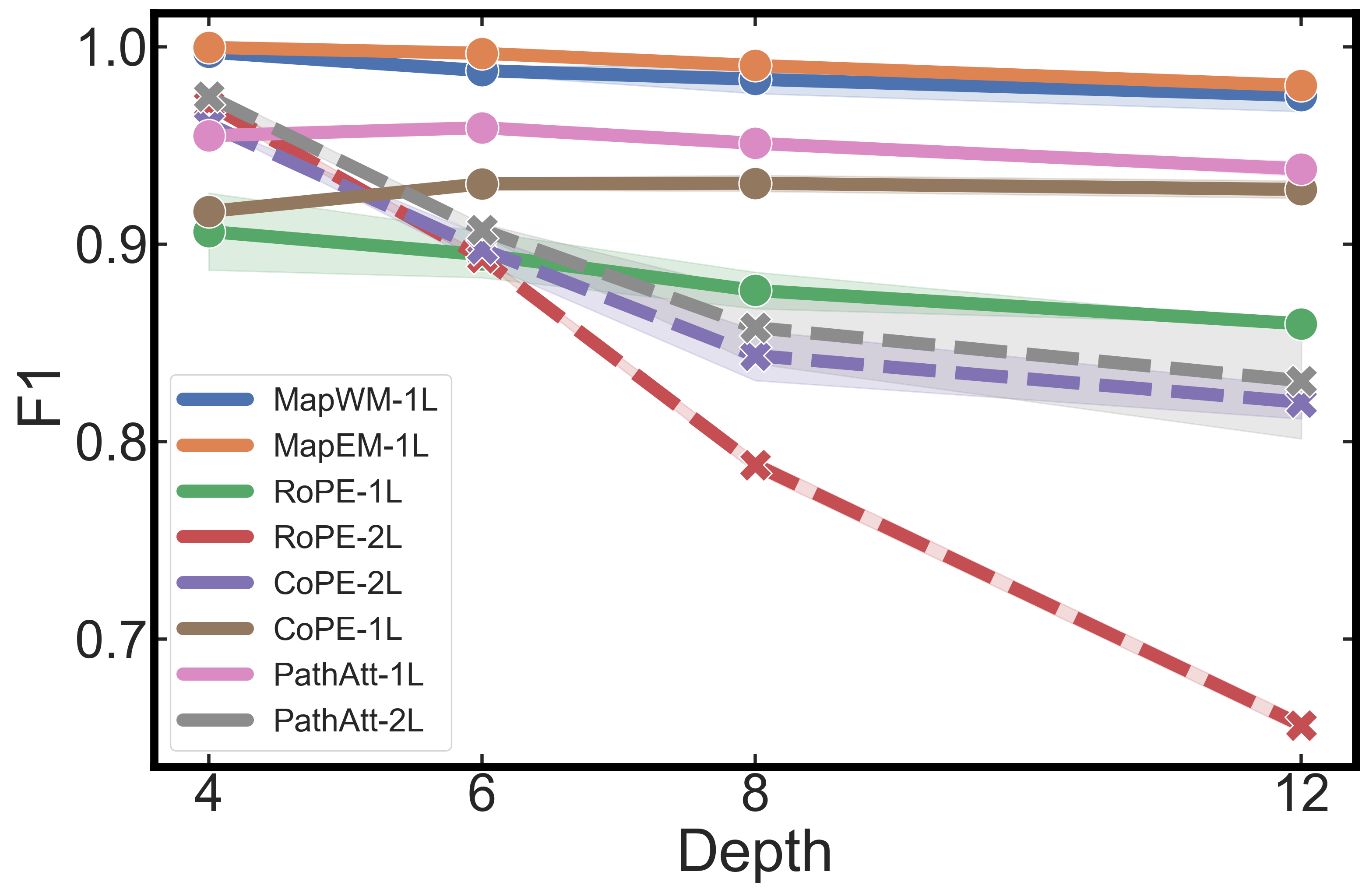}
        \caption{}
        \label{fig:dyck_gen:depth}
    \end{subfigure}
    \begin{subfigure}[c]{0.32\textwidth}
        \centering
        \includegraphics[width=\textwidth]{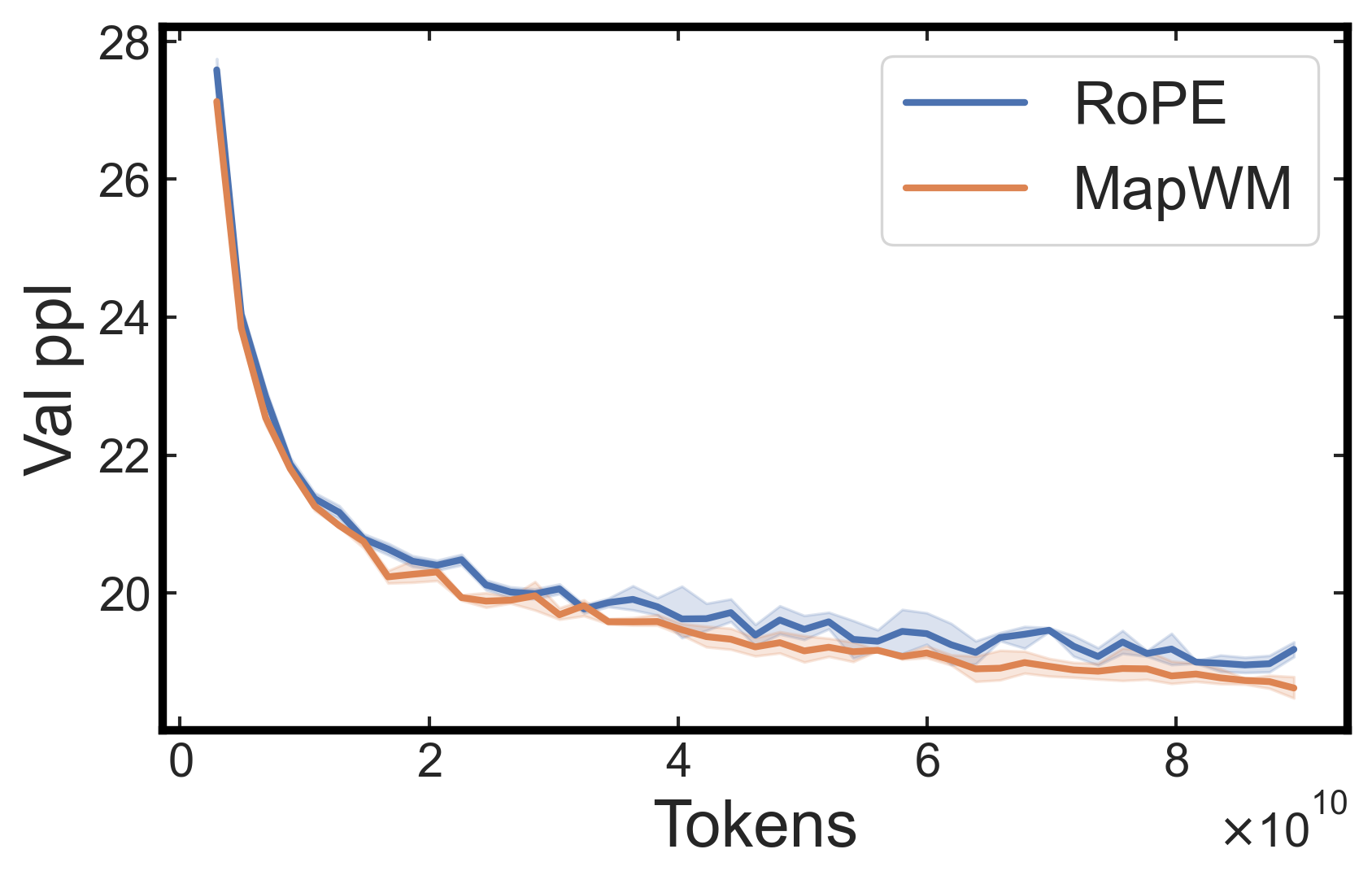}
        \caption{}
        \label{fig:dyck_gen:LM}
    \end{subfigure}
    \begin{subfigure}[c]{0.35\textwidth}
        \centering
        \includegraphics[width=\textwidth]{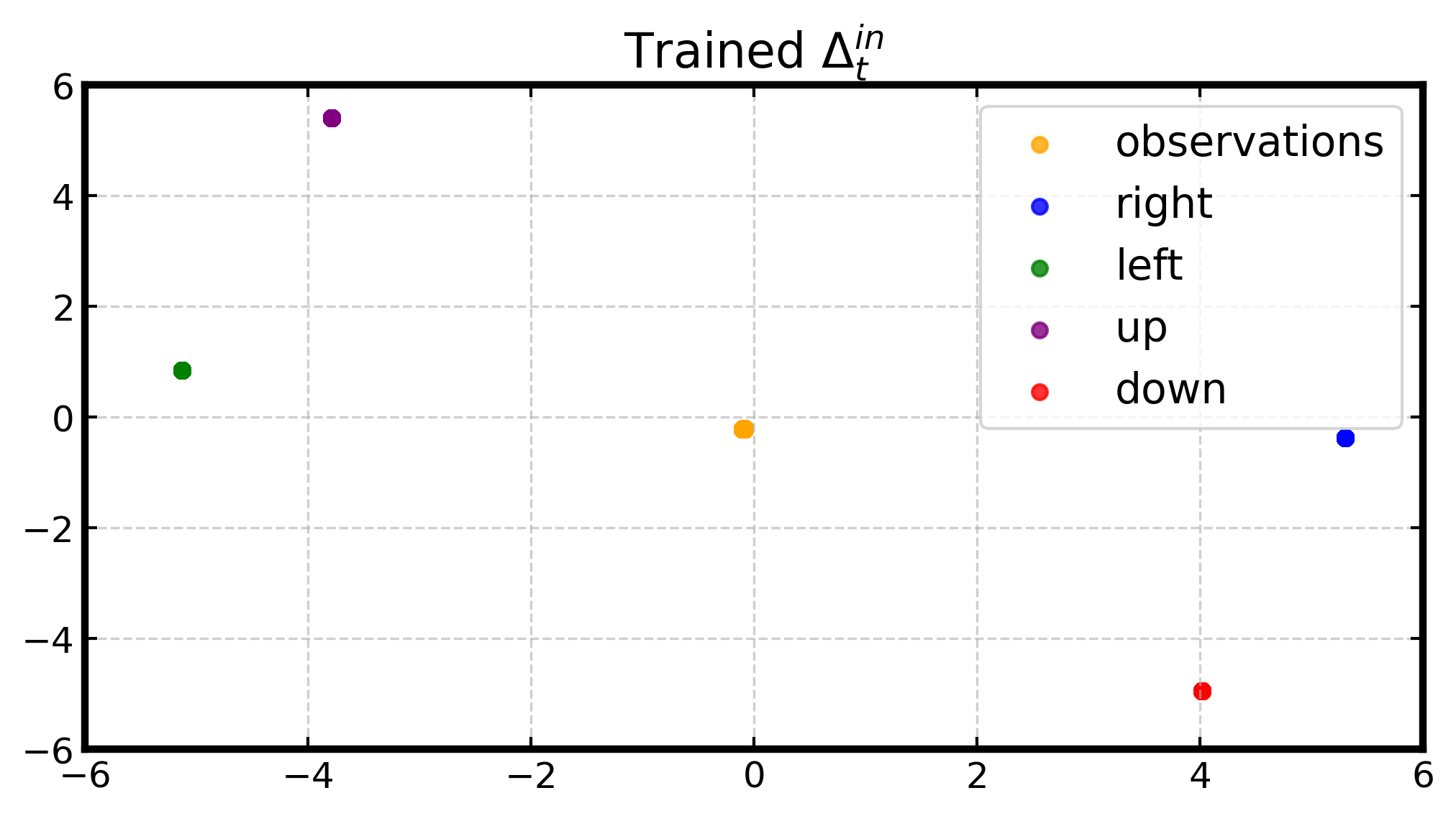}
        \caption{}
        \label{fig:dyck_gen:learndelta}
    \end{subfigure}
    \begin{subfigure}[c]{0.35\textwidth}
        \centering
        \includegraphics[width=\textwidth]{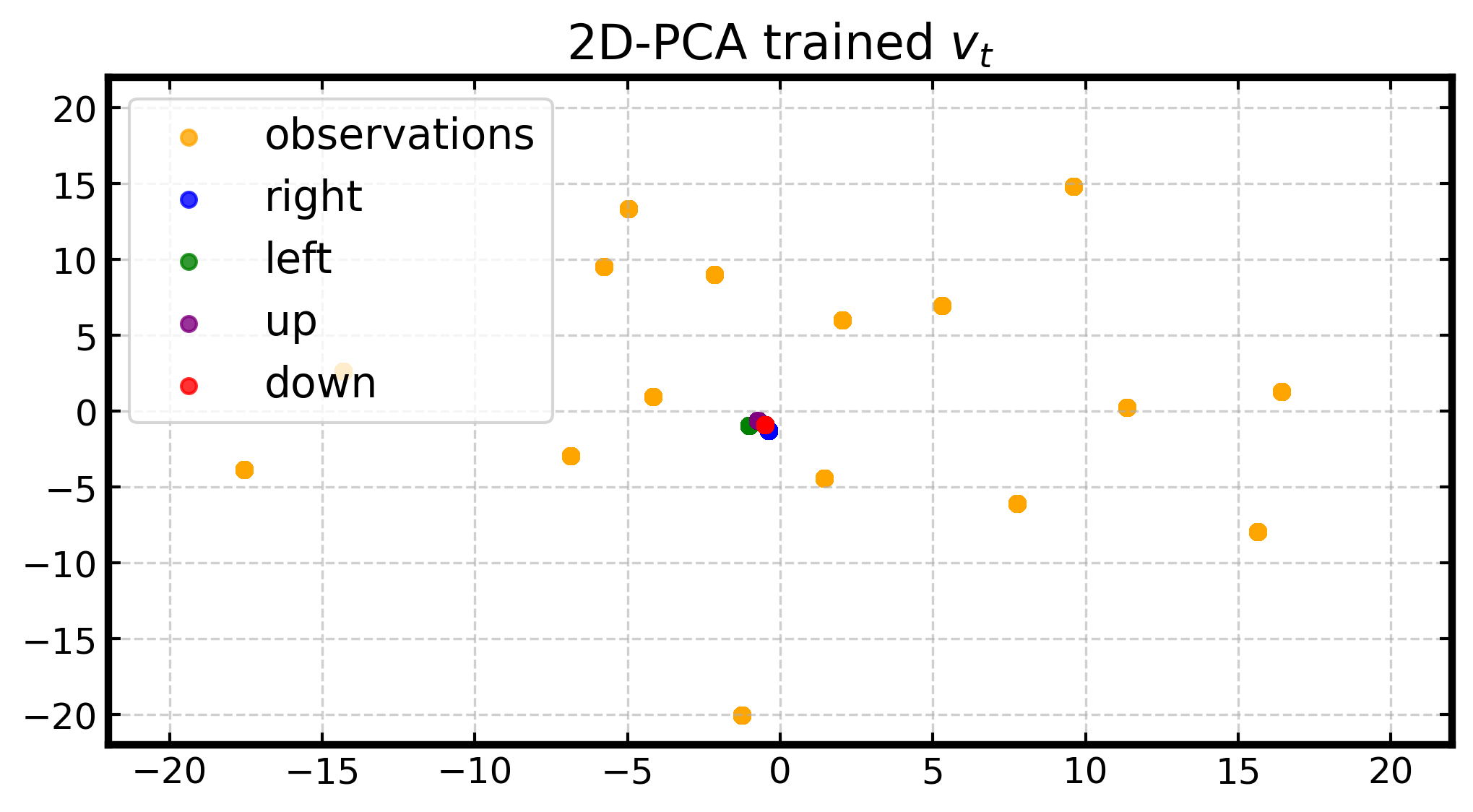}
        \caption{}
        \label{fig:dyck_gen:learnvalues}
    \end{subfigure}
    \begin{subfigure}[c]{0.26\textwidth}
        \centering
        \includegraphics[width=\textwidth]{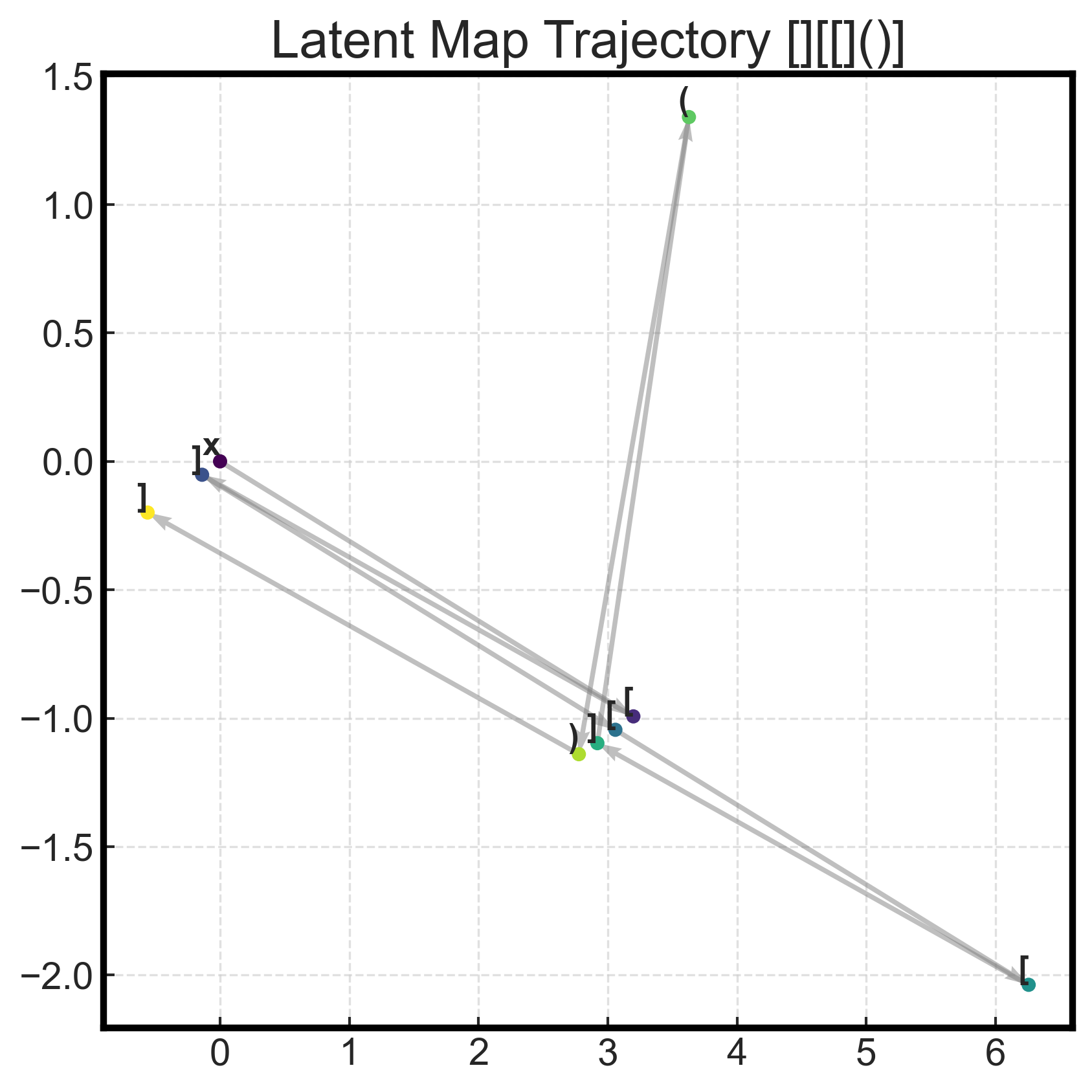}
        \caption{}
        \label{fig:dyck_gen:traj}
    \end{subfigure}

    \caption{\textbf{MapFormers outperform baseline models on Dyck-2 and natural language modeling.} \textbf{(a)} Length generalization: F1 as a function of sequence length; trained on length 32 (dashed line). \textbf{(b)} Depth generalization: F1 as a function of nesting depth; trained on depth 4 (dashed line). MapFormers solve the task with a single layer and generalize OOD, while baselines require a second layer to converge IID and fail OOD. \textbf{(c)} Validation perplexity on OpenWebText: 12-layer MapWM vs. RoPE, averaged over $5$ seeds. \textbf{(d-e)} Learned representations of MapWM in the 2D navigation task. \textbf{(d)} Inner projections $\Delta_t^{\mathrm{in}}\in \mathbb R^2$: observation tokens are all projected towards 0 (orange), while tokens representing opposite directions (left / right or up / down) are all projected to opposite directions. \textbf{(e)} PCA of learned values: action values $v_{a_t}$ are projected towards zero. \textbf{(f)} Dyck-2 latent trajectory $\text{cumsum} (\Delta^{\mathrm{in}}_t)\in \mathbb{R}^2$, for trajectory $[]([]())$, starting in symbol $\mathbf{x}$. The model uses opposite directions for open/close brackets of same type and orthogonal dimensions for each bracket type}
    \label{fig:dyck_gen}
\end{figure*}

Standard transformers require at least 2 layers to generate Dyck-2~\cite{yao-etal-2021-self}, but we hypothesized that \textit{MapFormers} can emulate a stack within their positional encoding and succeed with a single layer. We therefore compared 1 layer, 1 head to 2 layers and 2 heads ($h=64)$, training on sequences of length 32 and depth 4, and evaluating OOD along two axes: (1) by fixing depth and increasing length multiplicatively, up to $4\times$ the training length; (2) by fixing the sequence length and increasing depth additively up to training depth + 8. We report continuation quality using the F1-metric of \cite{goodale-etal-2025-meta}.

Fig.~\ref{fig:dyck_gen:len} \&~\ref{fig:dyck_gen:depth} show that  MapFormers with a single layer perfectly generalize OOD. In contrast, baseline models require a second layer to approach near-perfect training performance and, more importantly, fail to generalize beyond the training distribution. 



\subsection{Neural Analysis: Observations are Vectors, Actions are Matrices}
To complement the behavioral results, we analyze the learned embeddings in \textit{MapFormers}. A defining property of cognitive maps is the separation of content from structure, which can be directly probed in the model’s embeddings. In order to do so, we analyzed for each token the learned \textit{inner} movement $\Delta^{\mathrm{in}}_t$ on the cognitive map—where structure should be encoded—and the learned values $v_t$—which encodes the semantic content passed to the next layer. See further details in sec.~\ref{sec:neuralanaly_detail}.

In 2D navigation, the model learns with no supervision to disentangle structure from content, by projecting observation tokens to near zero movement on the map ($\Delta_{o_t}\approx 0$, \textit{i.e.} update the cognitive map with a near identity rotation), while opposite actions map to opposite directions (fig.~\ref{fig:dyck_gen:learndelta}). Conversely, since actions carry no semantic content in this task, their values are projected towards zero (fig.~\ref{fig:dyck_gen:learnvalues}).

On Dyck-2, open and closed brackets of the same type are projected toward opposite directions, while different bracket types use orthogonal dimensions (fig.~\ref{fig:dyck_gen:traj}). Unlike memory-augmented RNNs that require external memory modules \cite{suzgun2019memoryaugmentedrecurrentneuralnetworks}, \textit{MapFormers} embed the stack directly in positional encoding, enabling efficient parallel processing while achieving perfect generalization.

\subsection{Scalability to Natural Language}
To assess the scalability of our method (compatible with \textit{FlashAttention} \cite{dao2022flashattentionfastmemoryefficientexact} for \textit{Map}WM, tab.~\ref{tab:speed}), we pretrained a 12-layer \textit{Map}WM on OpenWebText \cite{Gokaslan2019OpenWeb} against a RoPE baseline, using the nanoGPT codebase~\cite{karpathy2022nanogpt} on 4 H100 GPUs for approximately $10^{11}$ tokens (fig.~\ref{fig:dyck_gen:LM}). Overall, \textit{Map}WM's training remained stable and yielded a consistent perplexity improvement over RoPE (RoPE $19.14_{\pm0.14}$ vs MapWM $18.79_{\pm0.15}$) and better length extrapolation (fig.~\ref{fig:length_extrapolation} in the appendix). However, matching the length-extrapolation of 1D path-integrating baselines like CoPE and PathAtt \cite{golovneva2024contextualpositionencodinglearning, yang2026pathattentionpositionencoding} would likely require more hyperparameter tuning. Notably, increasing the rank of $W^{\mathrm{in}}$ might create a bias for depth (as in Dyck-2, fig.~\ref{fig:dyck_gen:traj}) but reduce the precision on very long sequences.

Moreover, since \textit{MapFormers} disentangle structure from content, we further investigated whether our inductive bias would help to separate syntax from semantics, but preliminary results on BLIMP (\cite{warstadt-etal-2020-blimp-benchmark}, tab.~\ref{tab:blimp}) show no clear improvement, suggesting that the observed gains on OpenWebText do not come from better syntactic modeling, which might require another mechanism than (commutative) path-integration. See sec.~\ref{ssec:language_modelingdetails} for further details and discussion on language modeling.
\section{Discussion}
We introduced \textit{MapFormers}, a novel class of Transformer-based architectures that learn cognitive maps and perform path integration without supervision. By introducing the exponential map as a principled framework for dynamically building action-dependent matrices, \textit{MapFormers} address two key limitations of previous RNN-based models: (1) they enable parallel path integration on commutative groups (models in \cite{Whittington2025, Whittington770495, whittington2022relatingtransformersmodelsneural} are non-parallelizable even in 2D navigation), and (2) they remove the need to specify the underlying set of actions in advance, allowing to scale to multi-layer models that could learn \textit{context-dependent} cognitive maps.

We developed two variants, \textit{MapFormer}-WM and \textit{MapFormer}-EM, which
give a new perspective on the difference between absolute and relative positional encoding, by relating them to episodic and working memory, respectively. Because EM models scale better in recall tasks but are slower to run (sec.~\ref{ssec:recall_scaling} \& tab.~\ref{tab:speed}), future work could focus on hybrid architectures that alternate between absolute and relative positional encoding to improve episodic recall while maintaining competitive inference speed.

Empirically, \textit{MapFormers} achieved near-perfect OOD generalization on longer, sparser, and deeper sequences—baselines failed because they do not path-integrate with the right geometric prior—while our neural analysis in sec.~\ref{ssec:dyck_neural} demonstrates that the model does learns a cognitive map and allows us to interpret its step-by-step reasoning process. This framework enables gating (sec.~\ref{ssec:selective_copy}) and encoding of nested hierarchies (sec.~\ref{ssec:dyck_results}) and also extends to non-commutative, non-Euclidean cognitive maps (sec.~\ref{sec:non_commute}) suggesting that input-dependent matrices built from Lie-algebra generators might be a generic mechanism for structural learning on both smooth manifolds (infinitesimal motion in 2D navigation...) and symbolic structures (Dyck language, family trees...), by sampling small or large step sizes $\Delta_t$, respectively.

Finally, we used natural language as a scalability test, but reinforcement learning or world modeling might be more natural settings for studying the impact of cognitive maps in real-world scenarios.
\section{Limitations}

We highlight several limitations of our work that we leave for future explorations: (1) to compute \textit{path integration}, we sum the rotation angles along the temporal dimension, restricting our formalism to causal transformers, but the same methods used to adapt SSMs to vision \cite{zhu2024visionmambaefficientvisual} could be used for non-causal problems;
(2) although our analysis shows that non-commutative cognitive maps can be learned, the characterization of non-commutative path-integration could be broadened, and whether this is achievable in parallel remains an open question;
(3) our baselines focused on input-dependent positional encoding, but the comparison could be extended to other architectures such as memory-augmented networks;
(4) we trained language models at modest scale to validate that \textit{MapFormers} extend beyond synthetic tasks, but limited compute prevented hyperparameter tuning, longer training and broader benchmarking. Moreover, it is still unclear how limiting the commutative constraint will prove to be for language modeling.

\section*{Acknowledgments}
This work has benefited from public funding from the IA-Cluster PR[AI]RIE-PSAI project, managed by the French National Research Agency (ANR) as part of the France 2030 program, reference “ANR-23-IACL-0008”. Ecole Normale Supérieure-PSL’s Département d’Etudes Cognitives is supported by ANR grants, ANR-10-IDEX-0001-02 and FrontCog, ANR-17-EURE0017.

This work was performed using HPC resources from GENCI-IDRIS (Grant 2026-AD011016976).

\bibliography{references}
\bibliographystyle{unsrtnat}

\newpage
\appendix
\onecolumn
\section{Appendix}

\subsection{Action-Dependent Matrices Form a Lie Group}\label{ssec:groups}
The key idea to learn a cognitive map is to move beyond fixed matrices and learn action-dependent matrices, specific to each input action. 
In \cite{dorrell2023actionableneuralrepresentationsgrid}, the authors introduce \textit{actionable representations} using group and representation theory, as a formalism to map an input action $a \in \mathbb{R}^r$ (e.g. in 2D, $r=2$ and `move right’) to a neural transformation  $W_a\in M_n(\mathbb{R})$ (e.g., the corresponding weight matrix of a neural network that implements the effect of this action on the represented position in space $p$):

\begin{equation}
    \begin{aligned}
        \phi: \mathbb{R}^r &\to M_n(\mathbb{R}) \\
        a &\mapsto W_a
    \end{aligned}
\end{equation}

where $\phi_{a}$ acts on the representation $p \in \mathbb{R}^n$ (e.g., abstract position in space) such that $p(x+a) = W_a p(x)$. We refer to the matrices $W_a$'s as action-dependent matrices, since unlike standard RNNs, the weight matrix may depend on the specific action taken. To be useful for updating positions in cognitive maps, the action-dependent matrices need to form a group, and specifically, in addition to (1) closure and (2) associativity, it needs to satisfy the following properties: (3) include a matrix for the null action: $W_0=\mathbf{I}_n$, (4) $\forall a: W_a$ is invertible with inverse $W_{-a}$ (e.g., `one step right' has the exact opposite effect on the neural representation of `one step left' - $\forall x, a: p(x+a-a)=W_{-a} W_{a} p(x)=p(x)$). In summary, the requirements for a neural representation for a cognitive map to be consistently and reversibly updated by actions (including a null action and an inverse for every action) mean that the set of action matrices $\{W_a\}$ must constitute a subgroup of the General Linear Group, $\text{GL}(n, \mathbb{R})$, under matrix multiplication. Note that this results in various desired properties, such as making the actions $a$'s invariant with respect to the specific position $p$ they update -- In the absence of noise, the action has the same effect on any of the represented positions. Indeed, making the same step right should update the position in the same way whether we are in Paris or New York.

The insight that action-dependent matrices need to form a group, and in particular a Lie group over $\mathbb{R}$, allows the use of results from group and representation theory to understand what are the possible constraints on the form of $W_a$'s \cite{dorrell2023actionableneuralrepresentationsgrid}. Specifically, the Peter–Weyl theorem \cite{knapp1996lie} states that any compact group $\mathbf{G}$ can be represented in $M_n(\mathbb{R})$ as:

\begin{equation}\label{eq:irreps}
    W_a = M \begin{bmatrix}
    I_1(a) & 0 & \cdots & 0 \\
    0 & I_2(a) & \cdots & 0 \\
    \vdots & \vdots & \ddots & \vdots \\
    0 & 0 & \cdots & I_d(a)
    \end{bmatrix}M^{-1}
\end{equation}

where $I_k$ are called the \textit{irreducible representations} (\textit{irreps}) of group $\mathbf{G}$ and $M$ is an invertible matrix. This representation of action-dependent matrices, through the irreps of the group $\mathbf{G}$, provides insights into which core actions can be performed by $W_a$'s. For example, for navigation on a circle (1D) or torus (2D), the \textit{irreps} are rotations $R_{\theta_k}$:

\begin{equation}\label{eq:irreps_rotation}
    I_k(a): = R_{\theta_k} = \begin{bmatrix}
        \cos(\omega_k .\Delta_{t_k}) & -\sin(\omega_k .\Delta_{t_k})\\
        \sin(\omega_k .\Delta_{t_k}) & \cos(\omega_k .\Delta_{t_k})
    \end{bmatrix};\quad \theta_k = \omega_k .\Delta_{t_k}\in \mathbb{R},
\end{equation}

where $\omega_k, \Delta_{t_k} \in \mathbb{R}^r, r=1, 2$ are the angular frequencies and rotation duration, respectively, the former representing movement at different scales. That is, in this example, each $I_k$ performs a rotation of a subspace of the neural space. The matrix $M$ then performs a change of basis on this result, assuming $M$ is not the identity.


\subsection{Path Integration Can Be More Easily Computed in the Lie-Algebra Space}\label{ssec:algebra}
To learn a cognitive map, we saw that it is desirable to require action-dependent matrices to form a Lie group $\mathbf{G}$. A Lie group is a group $G$ that is also a smooth manifold, such that the group operations (multiplication and inversion) are smooth maps. For example, the group of 2D rotations $\mathrm{SO}(2)$ is a Lie group. To build a neural model that can learn cognitive maps, the goal would be to learn action-dependent matrices that form a Lie group.

To further characterize the group of action-dependent matrices, it is often simpler to analyze the corresponding Lie algebra. The associated Lie algebra $\mathfrak{g}$ of a Lie group $\mathbf{G}$ is the tangent vector space of $\mathbf{G}$ at the identity element. Lie-group theory shows the existence of an exponential map between the Lie algebra and the group $\exp(\cdot): \mathfrak{g} \to \mathbf{G}$, and in the case of a compact group, where the exponential map is surjective, any element $W_a \in \mathbf{G}$ of the group can be expressed as the exponential of an algebra element $A$, $W_a = \exp(A) \in \mathbf{G}$. This is useful since studying the structure of Lie groups (which involves, e.g., matrix multiplication) is often simpler from the point of view of the corresponding Lie algebra (which involves addition operations).

For the case of cognitive maps, we will think of elements of the Lie algebra $A$'s as action representations, and the corresponding group elements (expressed via the exponential map) as the action-dependent matrices $W_a$'s, which implement these actions in synaptic weights. The exponential map provides the link between the vector space of actions (the Lie algebra) and the space of weight matrices (the Lie group). Specifically, given the property of the exponential map (if the $A_i$'s commute, see sec.~\ref{sec:non_commute} for further details): $\exp\left(\sum_i A_i\right) = \prod_i \exp(A_i)$, the exponential mapping translates the additive structure of the Lie algebra (cumulation of actions in this vector space - e.g, `move right' then `move up') into the multiplicative structure of the Lie group (multiplication of the corresponding action-dependent weight matrices). This is useful for path integration, since the cumulative effect of taking actions $A_1, ..., A_n$ can be translated to the final action-dependent weight matrix $W_a=\prod_{i=1}^n \exp(A_i)$. As we will see, the exponential map allows in some cases to compute path
integration in parallel, in the vector space of the Lie algebra, by first computing the cumulative summation of subsequent actions and only then computing the exponent of the result.

For example, for spatial problems, such as navigation on a circle or a torus—and more generally, any 1-dimensional compact group, the Lie group of interest is $\mathrm{SO}(2)$ - the group of all 2D-rotations.

\begin{equation}\label{eq:rotation_matrix}
    R_{\theta} :=
    \begin{bmatrix}
        \cos(\theta) & -\sin(\theta)\\
        \sin(\theta) & \cos(\theta)
    \end{bmatrix}
\end{equation}

The corresponding Lie algebra of $\mathrm{SO}(2)$, denoted $\mathfrak{so}(2)$, consists of all $2 \times 2$ real matrices $A$ that are skew-symmetric ($A^\top = - A$). The Lie algebra $\mathfrak{so}(2)$ is a one-dimensional vector space, spanned by a single basis \textit{skew-symmetric} matrix $S:=-S^\top$ (2D-rotations have a single degree of freedom). Thus, all elements of the Lie algebra can be expressed by a single scalar (the angular velocity) $\omega$ as $A = \omega S$. A finite rotation $R_\theta$ can then be seen as generated by integrating the infinitesimal angular velocities $\omega$ over a duration $\Delta_t$:

\begin{equation}\label{eq:skew_sym_exp} 
S=- S^\top = \begin{bmatrix}
0 & -1\\
 1 & 0
\end{bmatrix}; R_\theta=\exp(\Delta_t A) = \exp((\omega \Delta_t) S) = \begin{bmatrix}
\cos(\omega \Delta_t) & -\sin(\omega \Delta_t)\\
\sin(\omega \Delta_t) & \cos(\omega \Delta_t)
\end{bmatrix}
\end{equation}

Intuitively, the skew-symmetric matrix $A=\omega S$ defines the instantaneous action direction $S$ and constant angular velocity $\omega$. The finite rotation $R_\theta \in \mathrm{SO}(2)$ is obtained by the exponential map, which performs a continuous integration of this velocity. Thus, the expression $R_\theta = \exp(\Delta_t A)$ shows that the final rotation yields a total angle of $\theta = \omega \Delta_t$, which defines the smooth path (geodesic) on the group manifold.

Note that the action group represented by this block-diagonal rotation matrices is the $k$-torus $\mathrm{SO}(2)^k$, where $k$ is the number of independent axis, each leaving in $\mathrm{SO}(2)$ and acting on orthogonal axes. Hence, the underlying matrix has a single generator per block, and $\mathfrak{so}(2)^k$ remains abelian, allowing additive path-integration. For this reason, we refer to $\mathrm{SO}(2)^k$ simply as $\mathrm{SO}(2)$—or for any other group, we mainly focus here on its block-wise representation, since it is what matters for parallelization—in the text.

This framework has important implications for learning a cognitive map. Given that all elements of a compact and connected Lie group (the action-dependent matrices in our case, like rotation groups) can be expressed by the exponential of a single linear combination of the associated Lie algebra generators, learning a cognitive map can be simplified: learning can be performed \emph{at the level of the algebra} rather than the non-linear group level. The algebra primarily involves linear operations and summation -- which are naturally compatible with modern deep learning frameworks and backpropagation -- while the Lie group involves matrix multiplication. This linearization is useful because the linear path integration required by the exponential map can be much more easily parallelized, as we will see further below.

More generally, given the group of action-dependent matrices $W_a$'s (which, as we saw, needs to be a subgroup of the general linear group $\mathrm{GL}(n, \mathbb{R})$), elements of the algebra, $A(t)$, can be expressed by a linear combination of generators $A_i$'s scaled by parameters $t_i$'s (the `integration times', which can be seen as coordinates along the different directions defined by the generators). If surjective, all group elements $W_a$'s can then be expressed using the exponential map:

\begin{equation}
\label{eq:lie_algebra_mapping}
\begin{aligned}
A(t) &= \sum_{i=1}^D t_i A_i \in \mathfrak{g}, \quad t_i \in \mathbb{R} \quad \text{(Linear combination in the Algebra)}\\
W_a &= \exp\left(\sum_{i=1}^D t_i A_i\right) \in G \quad \text{(Mapped to the Group via Exponential)}
\end{aligned}
\end{equation}

For the special case where the generators $A_i$ commute (i.e., the Lie bracket is zero), as in the simple $\mathrm{SO}(2)$ example, the exponential of the sum equals the product of the exponentials: $W_a = \prod_{i=1}^K \exp(t_i A_i)$. This shows again the benefit of the Lie algebra: it transforms the non-linear composition (multiplication) of the group into linear path integration (addition) within the algebra, thereby facilitating efficient learning and parallelization in the model.

\subsection{\textit{MAmPa}: Learning Cognitive Maps with block-diagonal Mamba Models}\label{ssec:PImba}
In this section, we propose a Lie-theoretic interpretation of State-Space Models (SSMs), which provides an interpretation of modern SSMs as models that implicitly perform path integration and can thus learn cognitive maps under certain conditions. This framework reveals a substantial limitation of current state-of-the-art SSMs, such as Mamba and Mamba2 \cite{gu2024mambalineartimesequencemodeling, dao2024transformersssmsgeneralizedmodels}, to meet these conditions for learning cognitive maps. We thus propose a simple modification of the Mamba architecture, which we call \textbf{\textit{MAmPa}}, for learning cognitive \textbf{Map}s with \textbf{Mamba} models. The link established in \cite{dao2024transformersssmsgeneralizedmodels} between SSMs and Transformers  is what will later allow us to understand the links between absolute and relative positional encoding with models of episodic and working memory, respectively.

Recently, structured state-space models (S4) \cite{gu2022efficientlymodelinglongsequences} have gained popularity over Transformers as they allow parallelized training of sequences while keeping a linear inference cost. These models are inspired by continuous state space models, where a continuous signal $x(t)\in \mathbb{R}$ is mapped to $y(t)\in \mathbb{R}$ via a hidden variable $h\in\mathbb{R}^n$:

\begin{equation}\label{eq:ssms_cont}
    \begin{aligned}
    h'(t) &= Ah(t) + Bx(t)\\
    y(t) &= Ch(t)
\end{aligned}
\end{equation}

These continuous systems can be applied to a sequence of discrete tokens, using the zero-order hold (ZOH) rule, which assumes that the input signal $x(t)$ is constant during a step size $\Delta$. This yields the corresponding discrete version of the dynamics:

\begin{equation}\label{eq:ssm_discrete}
    \begin{aligned}
        h_{t+1} &= \overline{A}h_t + \overline{B}x_{t+1}\\  
        y_{t+1} &= Ch_{t+1}
    \end{aligned} 
\end{equation}

where $\overline{A} = \exp (\Delta \cdot A)$ and $\overline{B} = (\Delta \cdot  A)^{-1} \left(\exp (\Delta \cdot A)-I\right)\Delta B$. \textbf{This discretization step is fundamental}, as it explains how the exponential map can arise from the integration of infinitesimal recurrent transformations over a time interval $\Delta\in \mathbb{R}$. The fact that recurrent computations are define with a single synaptic weight matrix $A$ implies that this exponential map can only generate 1-dimensional Lie-algebras, hence commutative groups.

In Mamba, an improved version of (S4), input selectivity is ensured via a per-token learnable step $\Delta_t$ and matrix $A$, that when combined defines a per-token transformation:

\begin{equation}
    \overline{A}_t:= \exp (\Delta_t A)
\end{equation}

The exponential relationship between $A$ (the continuous domain matrix) and $\overline{A}$ (the discrete domain matrix) directly parallels the relationship between the element $A$ of the Lie algebra and the weight matrix $W_a$ in the corresponding Lie group. In the rotation example, since the Lie-algebra $\mathfrak{so}(2)$ of \textit{skew-symmetric} matrices gives rise to the Lie-group of rotation matrices $\mathrm{SO}(2)$, $A:=\omega S$ is the infinitesimal generator and $W_a$ is the finite rotation over duration $\Delta_t$: $W_a=\exp(\Delta_t A)$. By identifying the discrete-time SSM matrix $\overline{A}$ with the action-dependent matrix $W_a$, we propose a new theoretical framework to analyze the capacity of Mamba to learn action-dependent matrices and cognitive maps. In this framework, Mamba can be interpreted as learning action-dependent matrices $\overline{A}_t$, which are dynamically created via an exponential map. As we saw, the exponential map integrates infinitesimal steps via $A$ (the direction and velocity in the Lie algebra space) over a varying time interval $\Delta_t$, giving a finite transformation $\overline{A}$ (the element of the Lie group). In other words, Mamba implicitly computes \textit{path integration} of action $A$ over a duration defined by the learned parameter $\Delta_t$.

Given this theoretical framework, it is evident that current versions of Mamba are incapable of learning cognitive maps that require, for example, learning rotations. In Mamba, $A$ was chosen as diagonal, to reduce computational cost, but a diagonal action-dependent matrix cannot learn rotations. This is since $\overline{A}$ needs to be similar to a block-diagonal matrix, with blocks of size 2 (eq. \ref{eq:irreps} \& \ref{eq:irreps_rotation}). Mamba was shown to be able to learn gating problems, which increases the performance of Mamba on various structure-dependent tasks by ignoring irrelevant information, however, Mamba with a diagonal constraint is not expressive enough to represent spatial cognitive maps as Lie-group theory shows. In this work, we modify $A$ in Mamba to be block-diagonal and, in particular, to blocks of size 2 (for learning rotations) and show that this modification increases the capacity of the model to learn cognitive maps.

\subsection{\textit{MAmPa} as a Model of Episodic Memory (EM) and Working Memory (WM)}\label{ssec:pimba_EMWM}

\begin{figure*}[ht!]
    \centering
    
    \begin{subfigure}[b]{0.42\textwidth} 
        \centering
        \includegraphics[width=\textwidth]{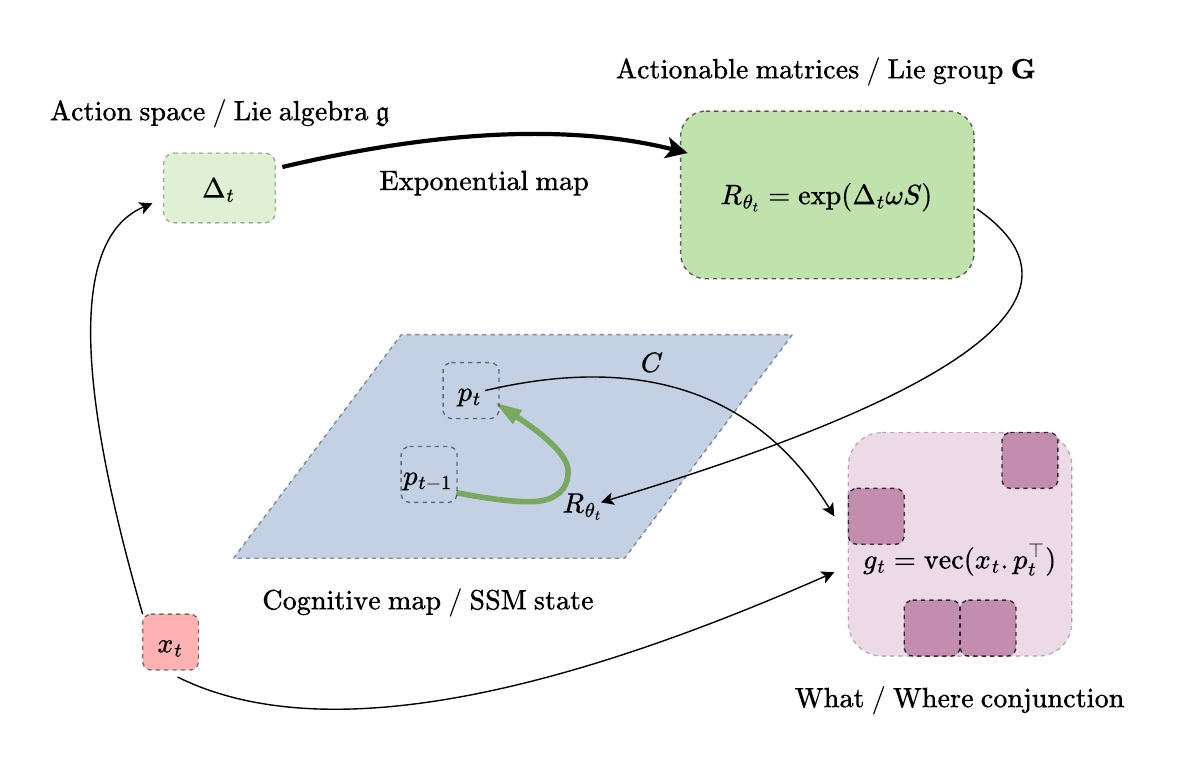}
        \caption{} 
        \label{fig:cogmap_ssm:a} 
    \end{subfigure}
    \hspace{0.2cm} 
    \begin{subfigure}[b]{0.50\textwidth}
        \centering
        \includegraphics[width=\textwidth]{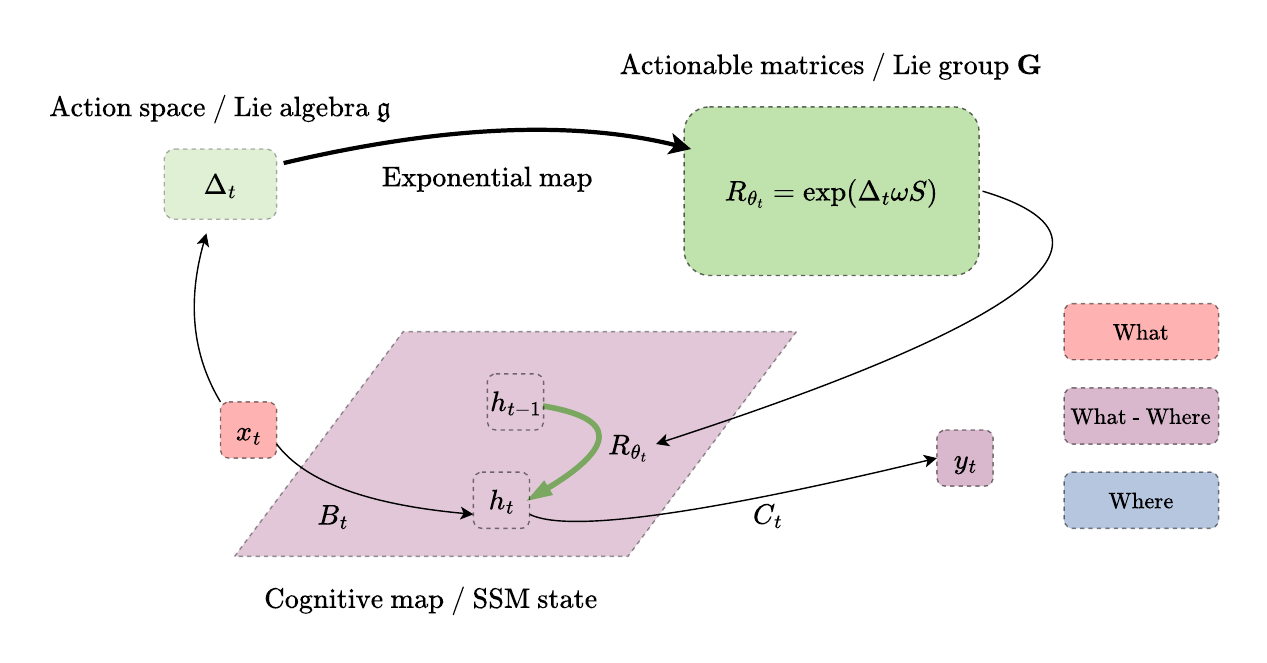}
        \caption{}
        \label{fig:cogmap_ssm:b} 
    \end{subfigure}
    
    \caption{\textbf{An illustration of the conceptual framework of \textit{action-dependent} matrices, and the differences between Episodic Memory (EM) and Working Memory (WM) models} \cite{Whittington2025}. (a) \textbf{Episodic Memory (EM) model}: Positions (or `states', more generally) in a cognitive map $p_t$'s (in blue), at times $t$'s, are encoded in a separate neural space by a dedicated neural population, which does not directly interact with the input $x_t$ (red), being updated via actions only. Actions $a_t$'s, when they occur in the input, can update the position via input-dependent matrices $W_{a_t}$ (green): $p_t=W_{a_t} p_{t-1}$. $W_a$'s are input-dependent matrices and are computed through an exponential map, which maps between abstract representations of actions $A_t:=\Delta_t \omega S=\theta_t S$ (in the Lie-Algebra) and the corresponding weight matrix $W_{a_t}:=\exp(A_t)=R_{\theta_t}$ (in the corresponding Lie group; section \ref{ssec:algebra}). The exponential mapping is computed based on the integration time $\Delta_t$ of the action (light green) -- roughly, how long the synaptic weights $S$ need to be applied. Finally, the conjunction (purple) of the content (what) and the structure (where) can be computed. (b) \textbf{Working Memory (WM) model}: In contrast to the EM model, which factorizes content (red) and position (blue) in two separate neural spaces, \textit{MapFormers}-WM represent position implicitly. Actions in the WM model directly update the conjunction of what and where (purple). The conjunction states $h_t$'s in the WM model are thus smaller by a quadratic factor compared to the EM model.
    }
    \label{fig:cogmap_ssm} 
\end{figure*}

In \cite{Whittington2025}, the authors introduce two related models of episodic (EM) and working memory (WM) that both learn cognitive maps using RNNs with \textit{input-dependent} matrices. Here, we explain why selective, discrete State-Space Models (SSMs) can be viewed as EM and WM models. First, note that discrete SSMs can be regarded as RNNs with no activation function. Second, selective SSMs learn \textit{input-dependent} matrices via an exponential map (eq.~\ref{eq:ssm_discrete}), meaning that their inner state captures the task's latent structure, as in EM/WM models. Therefore, EM and WM models can naturally be regarded as selective SSMs. 

\paragraph{SSM as a model of episodic memory} The EM-RNN model behaves exactly like TEM \cite{Whittington770495}, factorizing structure and content in two distinct neural populations, but only updates its position with an input-dependent RNN $p_t:=W_{a_t}p_{t-1}$. It then queries a Hopfield memory module \cite{ramsauer2021hopfieldnetworksneed} with its updated position to retrieve passed observations. The RNN in TEM can be seen as an SSM with subtle differences: (1) it uses no external input, \textit{i.e.} $\forall t,x_t=0$; (2) a learnable embedding $p_*$ representing the cognitive map's \textit{zero coordinate} must be introduced to avoid the state being always null; (3) SSM's inner state must be projected back to its final state ($y_t=Ch_t$, eq.\ref{eq:ssm_discrete}), which implies in this case that the final position $q_t:=Cp_t$ can be obtained with an optional learnable projection $C$; (4) as in TEM, the memory module is queried with the conjunction $g_t$ of \textbf{what} has been seen and \textbf{where} it was observed (fig.~\ref{fig:arch:EM}):
\begin{equation}
    \label{eq:EM_SSM}
    \begin{aligned}
        p_t&=\exp(\Delta_t A)p_{t-1}=\exp\left(\omega S \sum _{s=0}^t\Delta_s\right)p_*\\
        q_t&=C p_t = C R_{\theta_{0\to t}} p_*\\
        g_t&= \text{vec}(x_t^\top .q_t)
    \end{aligned}
\end{equation}
Setting $C:=\mathbf{1}_n$ implies that $q_t=p_t$, which is analogous to TEM/EM RNNs. Moreover, compared to RNN based models, this formulation allows to compute path integration directly in the Lie algebra ($\omega S\sum _0^t\Delta_s$) and avoid sequential matrix multiplication.

\paragraph{SSM as a model of working memory}
WM-RNNs directly updates the internal state with input-dependent matrices, which is precisely what selective SSMs do. As such, current selective SSMs with the appropriate matrix structure (\textit{skew-symmetric} in the case of navigation, eq.~\ref{eq:irreps_rotation}) naturally model working memory (fig.~\ref{fig:arch:WM}). 

Together, this allows us to view SSMs as both EM and WM models, where \textbf{the inner state of an SSM can be interpreted as a cognitive map}.

In the previous section, analyzing SSMs through Lie-group theory has allowed us to understand their current limitations, and that one only needs to switch to block-diagonal matrices to theoretically be able to learn a cognitive map. However, previous work decided to keep diagonal matrices because it dramatically reduces the computational load. By treating $A$ as a vector, one can avoid computing a matrix exponential and also dramatically reduce the memory footprint of a parallel scan, by computing vector-vector multiplications instead of matrix-vector's. In Mamba2 \cite{dao2024transformersssmsgeneralizedmodels}, it was further reduced to a scalar $A:=a\in \mathbb{R}$, which allows them to implement a fully parallelized version of selective SSMs but looses the ability to learn cognitive maps, as we have seen above. This highlights a fatal limitations of parallel-selective SSMs: they can be parallelized at the cost of loosing any hope of learning a cognitive map (see sec.~\ref{sec:limba} for further details). This is the reason why we decided to move away from SSMs and adapt Transformers to learn cognitive maps, as models of episodic (\textit{MapFormer}-EM) and working memory (\textit{MapFormer}-WM).

\subsection{Relationship between \textit{MapFormers} and \textit{MAmPa}: Linear \textit{Map}WM are selective SSMs}\label{ssec:map_are_ssms}

In sec.~\ref{sec:mapformers}, we have introduced the exponential map in Transformers (via the rotation matrices) to allow them to build \textit{input-dependent} matrices and learn cognitive maps. The Transformer formalism cannot naturally explain how an exponential map emerges like in SSMs, but in Mamba2 \cite{dao2024transformersssmsgeneralizedmodels}, the authors show that linear transformers \cite{wang2020linformerselfattentionlinearcomplexity} are actually SSMs. Starting from equation \ref{eq:ssm_discrete}:
\begin{equation}
    \begin{aligned}
    h_{t} &= A_th_{t-1} + B_tx_t\\
    &= A_t...A_1B_0x_0 + A_t...A_2B_1x_1 + ... + A_tA_{t-1}B_{t-2}x_{t-2} + A_tB_{t-1}x_{t-1} + B_tx_t\\
    &= \sum_{s=0}^t A^{\times}_{s\to t} B_s x_s; \quad A^{\times}_{s\to t} := \prod_{i=s+1}^t A_i
\end{aligned}
\end{equation}

after multiplying by $C_t$, we get the final equation:
\begin{equation}\label{eq:transssm}
    \begin{aligned}
        &y_t = \sum_{s=0}^t C_t^TA^{\times}_{s\to t} B_s x_s\\
        \Rightarrow& y = Mx;\quad m_{ij} := C_j^TA_j...A_{i+1}B_i
    \end{aligned}
\end{equation}

If we remove the matrices $A_t$, the matrix $M:=(m_{ij});\; m_{ij}=C_i^TB_j:= q_i^T k_j$ can be seen as the key-query similarity matrix $QK^T$ without softmax normalization, as in linear transformers. Equation~\ref{eq:transssm} is fundamental as it characterizes exactly how actionable representations can be implemented in transformers, where matrix $A_{i\to j}^{\times}=A_i...A_{j+1}$ represents the model's path integration between tokens $i$ and $j$.\newline

If we choose matrix $A:=S$ to be a $2\times2$ block-diagonal skew-symmetric matrix and $R_t:= \exp(\theta_t S)=\exp(\Delta_t \omega S)$ as in equation~\ref{eq:skew_sym_exp}, we retrieve the definition of RoPE \cite{su2023roformerenhancedtransformerrotary}:
\begin{equation}\label{eq:ssm_rope}
    \begin{aligned}
    m_{ij} &= q_iR^{\times}_{j\to i}k_j\\
    &= q^\top_iR^{\times}_{0\to i}\left(R^{\times}_{0\to j}\right)^{-1}k_j\\
    &= \left(\overline{R}_{i}q_i\right)^T \left(\overline{R}_{j}k_j\right);\; \overline{R}_i:= \left(R^{\times}_{0\to i}\right)^{-1}= \exp\left(-\sum_{s=0}^i \Delta_s \omega S\right):=\exp\left(\sum_{s=0}^i \theta_s S\right)\\
    &= \text{RoPE}(q_i, k_j)
\end{aligned}
\end{equation}

\textbf{This proves that linear RoPE transformers are SSMs with a $\mathbf{2\times 2}$ \textit{skew-symmetric} recurrent matrix}, where $\theta_t$ controls the angle of rotation. In Mamba2, the authors define $A:=a\in\mathbb{R}$ as a scalar, giving $M_{ij}=\exp(-a\sum_{t=j+1}^i \Delta_t)q_i^\top k_j$, which is analogous to the cumulative gating of CoPE (Contextualized Positional Encoding) \cite{golovneva2024contextualpositionencodinglearning}, with a multiplicative and not additive binding operation. CoPE is an enhanced Transformer architecture with context-dependent positional encoding, that like Mamba compresses the relative key-query distances with a cumulative gating mechanism, and in both models, input selectivity allows them to focus or ignore certain tokens and solve structure-dependent tasks, but should not allow them to learn spatial cognitive maps.

This extends the unification of transformers with models of episodic memory (EM) started in TEM-t \cite{whittington2022relatingtransformersmodelsneural}, showing that the difference between absolute and relative positional encoding can be better understood via the SSM formulation. EM models use \textit{action-dependent} matrices on a dedicated neural population (absolute positional encoding), before querying an external memory module, \textit{i.e.} modeling memories in synaptic activity, while WM models directly update the whole neural activity with \textit{action-dependent} matrices, directly

\subsection{\textit{MAmPa}: block-diagonal matrices improve performance of Mamba models in 2D navigation}\label{sec:limba}

As explained in sec.~\ref{ssec:PImba}, one only needs to modify the recurrent matrix $A:=S=-S^\top$ of selective SSMs to be block-diagonal skew-symmetric in order to learn spatial cognitive maps. As a proof of concept, we implemented an extended version of Mamba with $2\times 2$ block-diagonal matrices to verify that it can learn spatial maps like our theory predicts. Because the parallel scan on square matrices dramatically increases the computational load, we kept the inner state small ($N=32$) and only trained models on sequences of size $l=32$. We report the results in tab.~\ref{tab:limba}, where OOD dense and OOD sparse use sequences of size $16$ and $64$ respectively.

\begin{table}[h]
    \centering
    \begin{tabular}{l c c c} 
        \toprule
        \textbf{} & \textbf{IID} & \textbf{OOD-d} & \textbf{OOD-s} \\
        \midrule
        Mamba & {$0.38_{\pm 0.02}$} & {$0.66_{\pm 0.03}$} & {$0.30_{\pm 0.01}$} \\
        \textbf{\textit{MAmPa}} & {$\textbf{0.84}_{\pm 0.03}$} & {$\textbf{0.96}_{\pm 0.03}$} & {$\textbf{0.71}_{\pm 0.04}$} \\
        \midrule
        \textbf{\textit{Map}WM} & {$\textbf{1.00}_{\pm 0.00}$} & {$\textbf{1.00}_{\pm 0.00}$} & {$\textbf{1.00}_{\pm 0.00}$} \\
        \textbf{\textit{Map}EM-os} & {$\textbf{1.00}_{\pm 0.00}$} & {$\textbf{1.00}_{\pm 0.00}$} & {$\textbf{1.00}_{\pm 0.00}$} \\
        \textbf{\textit{Map}EM-s} & {$\textbf{1.00}_{\pm 0.00}$} & {$\textbf{1.00}_{\pm 0.00}$} & {$\textbf{1.00}_{\pm 0.00}$} \\
        \bottomrule
    \end{tabular}
    \vspace{5pt}
    \caption{\textbf{SSM 2D grid navigation - Accuracy.} As expected, \textit{MaMPA} offers substantial improvements over Mamba, but fails to reach performances a par with \textit{MapFormers}.}
    \label{tab:limba}
\end{table}

Tab.~\ref{tab:limba} confirms that our formalism extends to SSMs, since our SSM using SO(2) matrices significantly outperforms the Mamba variant and future work could focus on implementing a hardware efficient SO(2) SSM for appropriate scaling.

We also compared our SSMs to MapFormers, the latter reaching a perfect accuracy on all datasets. Many things could explain this difference: (1) the convolution kernel used in Mamba-SSMs might add noise before computing the projection $\Delta=f(\mathrm{conv1d}(X))$, since an action token will be averaged with nearby observation—irrelevant to the cognitive map—and conversely; (2) transformers use a softmax—known to increase the amount of memories that a model can remember \cite{ramsauer2021hopfieldnetworksneed}—which might explain the increased performances of transformers on recall tasks compared to SSMs.

\subsection{Implementation details}\label{sec:implem_details}

We chose rotation matrices because they are well suited to represent most bounded commutative structures (like movement on a finite grid or integers modulo $N$), but also because they have useful numerical properties: (1) rotations and their inverse are orthogonal and therefore ensure numerical stability, not present in Mamba, since the inverse of the exponential gating matrix they use diverges towards infinity, making it impossible to represent both an action and its inverse in a numerically stable way; (2) their transpose is also their inverse, implying that the key-query dot product $\overline{q}_i^\top.\overline{k}_j=q_i^\top R_{PI_{j\to i}}k_j$ represents path integration between tokens $i$ and $j$, necessary to learn cognitive maps in parallel.

Rotations can be ensured by enforcing $S:=\overline{S}-\overline{S}^\top$ to be skew-symmetric, but since a single generator matrix $S$ can only generate rotations along a single axis (see sec.\ref{sec:non_commute} for further details), choosing blocks of size $b>2$ is useless as it increases the computational load while effectively reducing the amount of rotations being performed. Using blocks of size $2$ also allows us to use the explicit \textit{irreps} representation of $2\times 2$ rotations and avoid computing a matrix exponential (eq.\ref{eq:irreps_rotation}). This also allows, as in RoPE \cite{su2023roformerenhancedtransformerrotary}, to rewrite matrix multiplication of $R_{\theta}x$ in \textit{MapFormers} as:

\begin{equation}
    R_{\theta}x=\begin{pmatrix} 
        x_1 \\ 
        x_2 \\ 
        \vdots \\
        x_{d_h-1}\\
        x_{d_h}\\
    \end{pmatrix}
    \odot
    \begin{pmatrix} 
        \cos\left(\Delta_{t_1}.\omega_1\right) \\ 
        \cos\left(\Delta_{t_1}.\omega_1\right) \\ 
        \vdots \\
        \cos\left(\Delta_{t_{n_b}}.\omega_{n_b}\right) \\ 
        \cos\left(\Delta_{t_{n_b}}.\omega_{n_b}\right)
    \end{pmatrix}
    +
    \begin{pmatrix} 
        -x_1 \\ 
        x_2 \\ 
        \vdots \\
        -x_{d_h-1}\\
        +x_{d_h}\\
    \end{pmatrix}
    \odot
    \begin{pmatrix} 
        \sin\left(\Delta_{t_1}.\omega_1\right) \\ 
        \sin\left(\Delta_{t_1}.\omega_1\right) \\ 
        \vdots \\
        \sin\left(\Delta_{t_{n_b}}.\omega_{n_b}\right) \\ 
        \sin\left(\Delta_{t_{n_b}}.\omega_{n_b}\right)
    \end{pmatrix}
    \label{eq:rot_vec_prod}
\end{equation}
leaving for each head only angular velocities $\omega\in \mathbb{R}^{n_h\times n_b}$ (see sec.~\ref{ssec:freqs} for a discussion about their initialization) as learnable parameters instead of matrix $S\in \mathbb{R}^{n_h \times n_b\times b\times b}$.

The actual rotation angle $\theta_t=\omega \Delta_t$ is dynamically computed from the input features $X$ using a low-rank projection $W_\Delta=W_\Delta^{\mathrm{out}} W_\Delta^{\mathrm{in}}$. The goal of this two-step process is to ensure the rotation relies only on the most salient action features of the input. First, for each head, the internal projection $W_\Delta^{\mathrm{in}} \in \mathbb{R}^{d \times n_h\times r}$ ($r \ll d$) maps the high-dimensional input $X$ to a low-dimensional $\Delta_t^{\mathrm{in}} \in \mathbb{R}^{t \times n_h \times r}$ (for instance, in 2D, $\Delta_t^{\mathrm{in}}$ is the 2D movement vector $\Delta_{t_k}\in \mathbb{R}^r$ in eq.~\ref{eq:irreps_rotation}, where $r=2$). Second, this extracted action is projected by $W_{\Delta}^{\mathrm{out}}$ into the final set of path increments $\Delta_t \in \mathbb{R}^{t \times n_h \times n_b}$, which controls the rotation angle for all attention heads ($n_h$) and all diagonal blocks ($n_b$), where $n_b = d_h / b$ is the number of blocks per head dimension.

Finally, compared to TEM and TEM-t that use a single neural population $p_t$ to encode position, our \textit{MapFormers} use two separate initial vectors $k_0^p$ and $q_0^p$. This distinction is optional and mimics the formalism of EM-SSMs defined in eq.~\ref{eq:EM_SSM}, meaning that we could set $k_0^p=q_0^p=p_0$ without loss of generality. However, we suspect this separation to be beneficial because it would create sparser attention values. Indeed, our positions $p_t$ vary smoothly through path-integration, implying that nearby positions would not be strongly separated, but introducing distinct neural populations for keys and queries could allow the model to find the optimal balance between separation/matching of distinct/similar positions.

Recall that the conjunctive query $g^q_i:=\text{vec}({q^x_j}^\top.q^p_i)$, the conjunction of what has been seen and where. Since $\forall i, j:\langle g^q_i, g^k_j \rangle = \langle {q^x_j}^\top.q^p_i, {k^x_i}^\top.k^p_j \rangle = \langle {q^x_j}, k_j^x \rangle\langle q^p_i, k^p_j \rangle$, we can avoid the outer products $Q_X^\top Q_P$ and $K_X^\top K_P$ by computing content and position attentions $A_X=\textbf{Att}(Q_X, K_X)$ and $A_P=\textbf{Att}(Q_P, K_P)$ separately (still in parallel by concatenating the heads), such that $A_P$ acts as an attention mask on $A_X$

The fact that \textit{MapFormer}-EM models rely on observations (content) or structure (position) attention $A_X$ and $A_P$, allows to balance between the two attention maps to compute similarities. To test the impact of each modality, we introduce three types of EM models: (1) \textit{Map}EM-os relying on both observation and structure to compute attention; (2) \textit{Map}EM-s relying on structure alone and (3) \textit{Map}EM-o relying solely on observation, equivalent to a Transformer without positional embeddings, serving as a control. Note that ideally, the model should learn to balance between observation or structure depending on context, but we decided to leave it for future work. This forced separation is primarily used to highlight the benefits of focusing on different modalities depending on the task.

\subsection{Angular velocities initialization}\label{ssec:freqs}

Like \textit{grid cells} that fire at different frequencies \cite{Hafting2005}, our formalisms uses rotations at different scales, scale whose initialization has a major impact on position representations.

\begin{figure}[h]
    \centering
    \begin{minipage}{0.4\textwidth} 
        \centering
        \includegraphics[width=\linewidth]{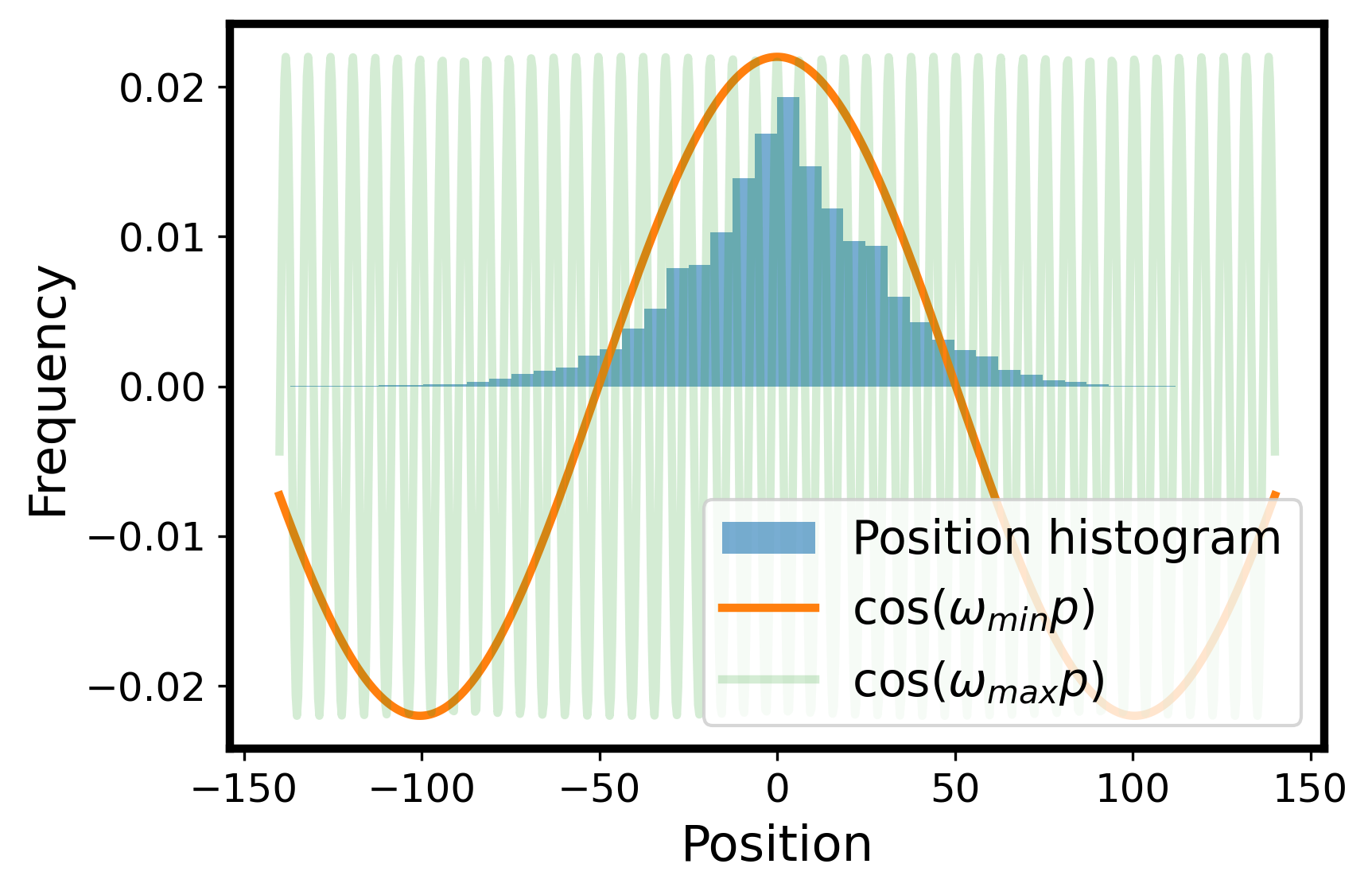} 
        \vspace{-0.5em}
        \label{fig:freqs:dist}
        \centering (a)
    \end{minipage}%
    \hfill%
    \begin{minipage}{0.28\textwidth} 
        \centering
        \includegraphics[width=\linewidth]{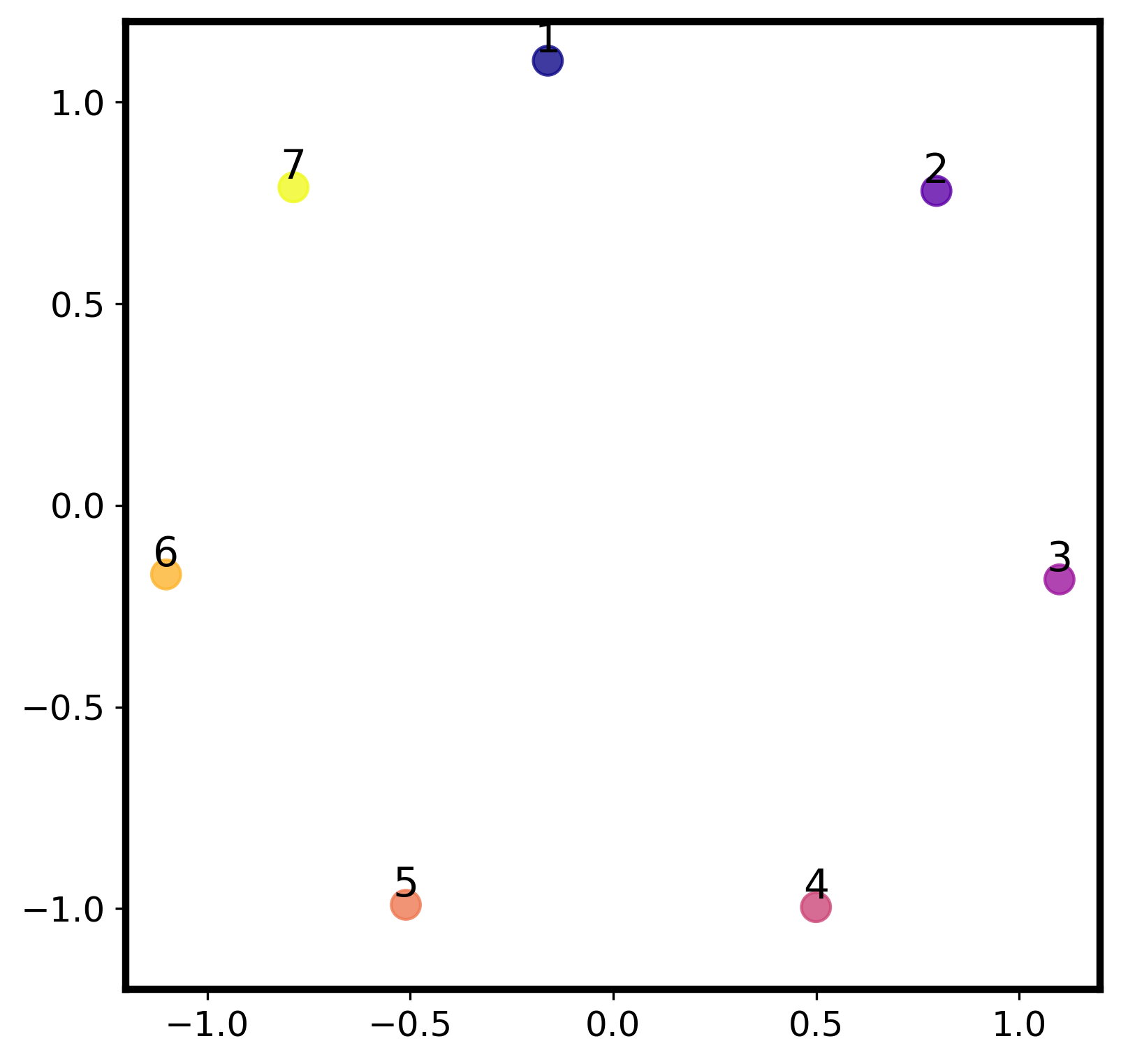} 
        \vspace{-0.5em}
        \label{fig:freqs:circ1}
        \centering (b)
    \end{minipage}
    \hfill%
    \begin{minipage}{0.28\textwidth} 
        \centering
        \includegraphics[width=\linewidth]{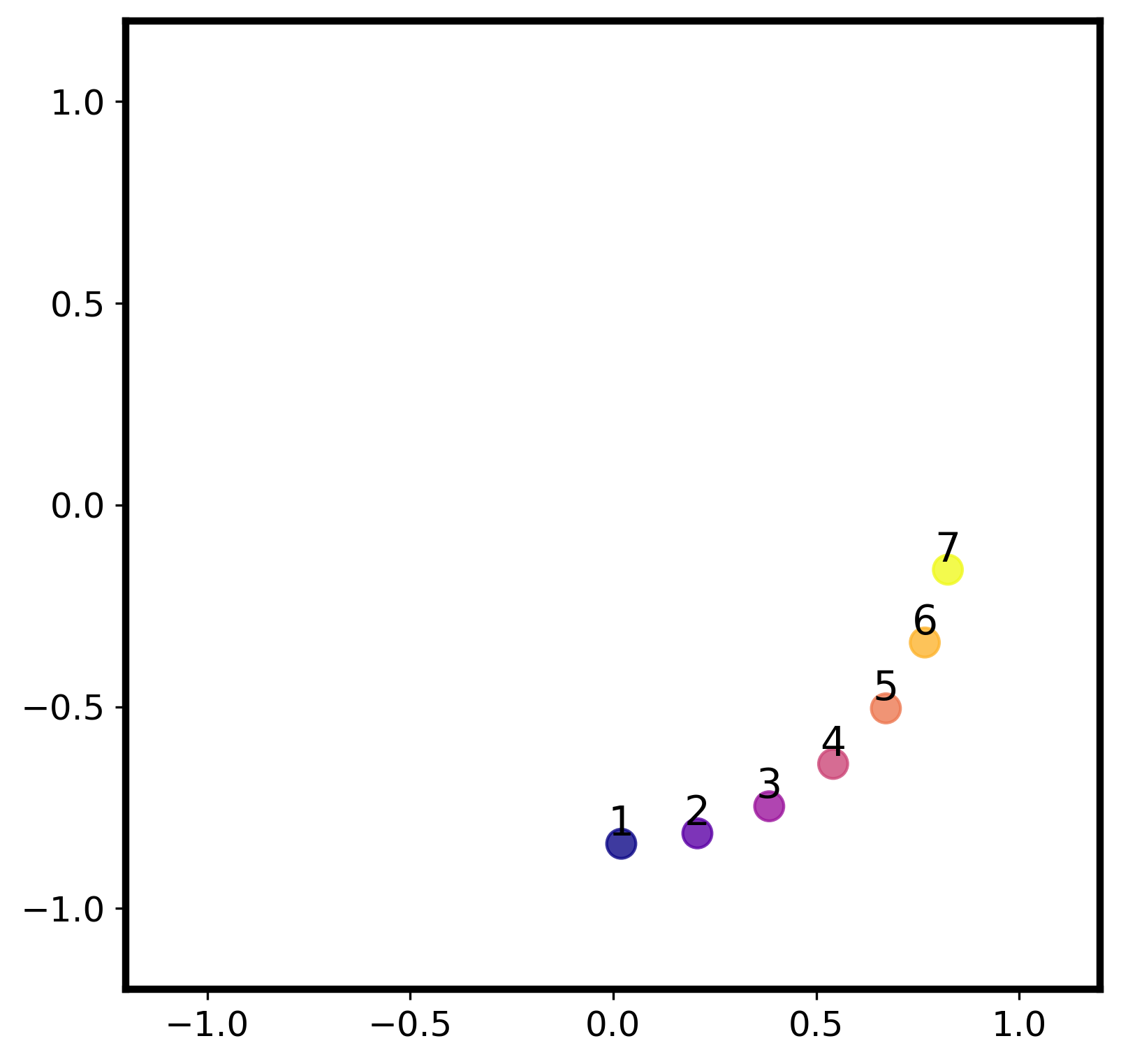} 
        \vspace{-0.5em}
        \label{fig:freqs:circ2}
        \centering (c)
    \end{minipage}

    \vspace{0.5em} 
    
    \caption{\textbf{(a)} Distribution histogram of position in 1D navigation vs oscillation at the highest (green) and lowest (orange) angular velocities. The longest distances reached by the model are comparable with the length of $\omega_{min}$ cycles.\textbf{(b-c)} Rotation blocks at high and low frequencies.}
    \label{fig:freqs}
\end{figure}

The highest angular velocity (or frequency $f$, assuming $\omega = 2\pi f$) $\omega_{max}$ defines the finest granularity $\Delta_{min}=\frac{2\pi}{\omega_{max}}$ at which the signal is encoded, while low frequencies encode long range dependencies. We would like the lowest angular velocity $\omega_{min}:=\frac{2\pi}{\Delta_{max}}$ to complete its cycle once the the model has covered the longest distance $\Delta_{max}$ it possibly can, where $\Delta_{max}=n$ on a $n$-size grid (fig.~\ref{fig:freqs}).

Angular velocities should encode signal at different scales, and a geometric initialization like the one used in RoPE \cite{su2023roformerenhancedtransformerrotary} spreads the scales uniformly between 1 and $\Delta_{min}=\frac{1}{base}$ with: 
$$\omega_i = \left( \frac{1}{\Delta_{max}} \right)^{-\frac{i}{n_b}},\quad 1\leq i \leq n_b$$
where $base\approx 10000$ in RoPE, dictates how many tokens the model has to see before completing a full circle. Compared to RoPE that always rotates by a constant rotation duration $\Delta = 1$, our input-dependent rotation durations $\Delta_t$ can take values bellow $1$, and as such we might want our highest angular velocity $\omega_{max}$ to be greater than $1$. We extend the logic such that:

\begin{equation}
\label{eq:frequencies} 
\forall i: \omega_i = \omega_{\max} \left( \frac{1}{\Delta_{\max}} \right)^{-\frac{i}{n_b}} \implies
    \left\{
    \begin{aligned}
        \omega_{\max} &= \frac{2\pi}{\Delta_{\min}} \\
        \omega_{\min} &= \frac{\omega_{\max}}{\Delta_{\max}} = \frac{2\pi}{n} \\
        \implies \Delta_{\max} &= \frac{n \times \omega_{\max}}{2\pi}
    \end{aligned}
    \right.
\end{equation}

Leaving only $\omega_{\max}$ and $base:=n$ as hyperparameter-parameters and find that initializing $1 <\omega_{max}\leq 2\pi$ to work well in practice.

When modeling position in dimension $n$ ($n>1$), we devise the embedding space into $n$ orthogonal subspaces, and initialize each subspace's frequencies as a 1-dimensional space.

\subsection{MapFormers achieve competitive speed over baseline Transformers}

In order to validate the scalability of our approach, we evaluated in tab.~\ref{tab:speed} throughput of 12 layer models MapFormer vs a baseline RoPE transformer, on forward, backward and optimizer steps, averaged over 100 calls. As expected, our Working Memory MapFormer, compatible with \textit{FlashAttention}, shows speed on-par with baseline. The Episodic Memory model, which needs two neural populations while being incompatible with \textit{FlashAttention}, is slower, perfectly highlighting the tradeoff between processing speed and recall capacities highlighted in fig.~\ref{fig:model_scaling}. We also tested non-commutative models (see sec.~\ref{sec:non_commute}), which raised OOMs at that scale.

\begin{table}[h!]
    \centering
    \begin{tabular}{l c c c} 
        \toprule
        \textbf{Architecture} & \textbf{Sequence length} & \textbf{Batch size} & \textbf{Throughput} \\
        \midrule
        RoPE (FlashAttention) & 256 & 64 & \textbf{1939} samples/s \\
        MapWM (FlashAttention) & 256 & 64 & \textbf{1869} samples/s \\
        MapEM & 256 & 64 & 1210 samples/s \\
        \bottomrule
    \end{tabular}
    \vspace{5pt}
    \caption{\textbf{Hardware Throughput.} Benchmarks conducted on a single NVIDIA H100 GPU using bfloat16 precision. Model configuration: $L=12$, $d_{model}=768$, $n_{heads}=12$, $d_{head}=64$. Throughput is averaged over 100 iterations, for forward, backward and optimizer steps.}
    \vspace{-0.5cm}
    
\end{table}\label{tab:speed}



\section{Task details and baseline models}\label{sec:task_details}

For all tasks beyond natural language, models were trained using a single H100 GPU.

\subsection{Baseline models}\label{ssec:baseline_models}

\textbf{Rotary Position Embedding (RoPE) } \cite{su2023roformerenhancedtransformerrotary} is used as our baseline Transformer since it is the most popular (fixed) position encoding architecture, and is analogous to our \textit{Map}WM model without input selectivity, theoretically incapable of learning input-dependent matrices and structure-based problems.

\textbf{Contextualized Equivariant Position Embedding (TAPE) } \cite{zhu2025rethinkingaddressinglanguagemodels} updates positional encodings after attention via input-dependent MLPs, which has two drawbacks: (1) updates rely on structure from the \emph{previous} layer rather than the current one, and (2) they are driven by attention scores rather than path integration, with no guaranty of long-range structural coherence.

\textbf{Contextualized Positional Encoding (CoPE) } \cite{golovneva2024contextualpositionencodinglearning} compresses relative distances with a cumulative gating mechanism, akin to path integration. Since the sigmoid gate can only take positive values, the model cannot represent an action and its inverse. Moreover, since CoPE only compresses relative ordinal distance, it should not be able to represent structure in higher dimensions.

\textbf{Path Attention (PathAtt) } \cite{yang2026pathattentionpositionencoding} uses rank-1 Householder matrices to sequentially update positional encoding. This model path integrates by learning possible inverse of actions, but rank-1 Householder matrices cannot learn complex structures such as 2D rotations.

\textbf{MapEM-s and MapEM-o } can choose to balance between observation $o$ and structure $s$, giving rise to \textit{Map}EM-s, \textit{Map}EM-o or \textit{Map}EM-os by choosing to use either structure, observation or both in eq.~\ref{eq:absolute_pe}, respectively.

\textbf{MapWM-r1 and MapWM-r2 } when specified, MapWM-r designate the rank of the low-rank projection of our learned velocity signals $\Delta_t$ defined in sec.~\ref{sec:implem_details}.

\subsection{Selective copy}\label{ssec:selec_copy}

The selective copy task was introduced in Mamba \cite{gu2024mambalineartimesequencemodeling} and used in \cite{golovneva2024contextualpositionencodinglearning} to show that standard backbones are unable to dynamically ignore irrelevant tokens. Specifically, given a blank token $B$ and a sequence $CFEABBCBF$, the model must copy it in order, without the blanks $CFEACF$. If the proportion of blank tokens remains fixed during training, the model is able to learn a form of heuristic by attending to tokens at a fixed position offset. However, solving it OOD requires the model to attend to a specific token, regardless of how many distractors lie in between, therefor requiring a form of gating.

Mamba SSMs \cite{gu2024mambalineartimesequencemodeling, dao2024transformersssmsgeneralizedmodels} are able to do so by multiplying the state with a negative diagonal matrix $h_{t+1}=e^{-\Delta_t A}h_t$. CoPE \cite{golovneva2024contextualpositionencodinglearning} solves it by compressing position count between tokens, and we posit that \textit{MapFormers} will be able to do so by incrementing positions on a circle.

Following the setup of \cite{golovneva2024contextualpositionencodinglearning}, we train 2 layer models with 2 heads of size $h=64$, on sequences of length $l=128$ and $128$ blanks. Non blank tokens were sampled among a vocabulary of $K=16$. OOD dense and OOD sparse are generated by lowering or increasing the amount of blank tokens to 64 or 256, while keeping the number of non blank tokens fixed.

We train all models on 100K sequences, with a batch size of 64, a learning rate $\mathrm{lr}=2e^{-4}$, AdamWM with a weight decay $0.05$. Moreover, we found that using a temperature of 0.90 to be beneficial for stabilizing training on all models.

\subsection{Forced Grid Navigation}\label{ssec:navi_detail}

    

\begin{figure}[ht!]
    \centering

    \begin{subfigure}[c]{0.30\textwidth}
        \centering
        \includegraphics[width=\textwidth]{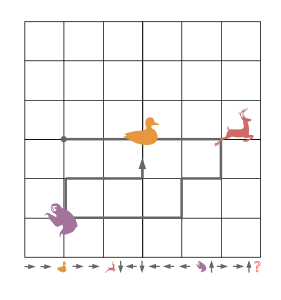}
        \caption{}
        \label{fig:taskdet:a}
    \end{subfigure}
    \hfill
    \begin{subfigure}[c]{0.30\textwidth}
        \centering
        \includegraphics[width=\textwidth]{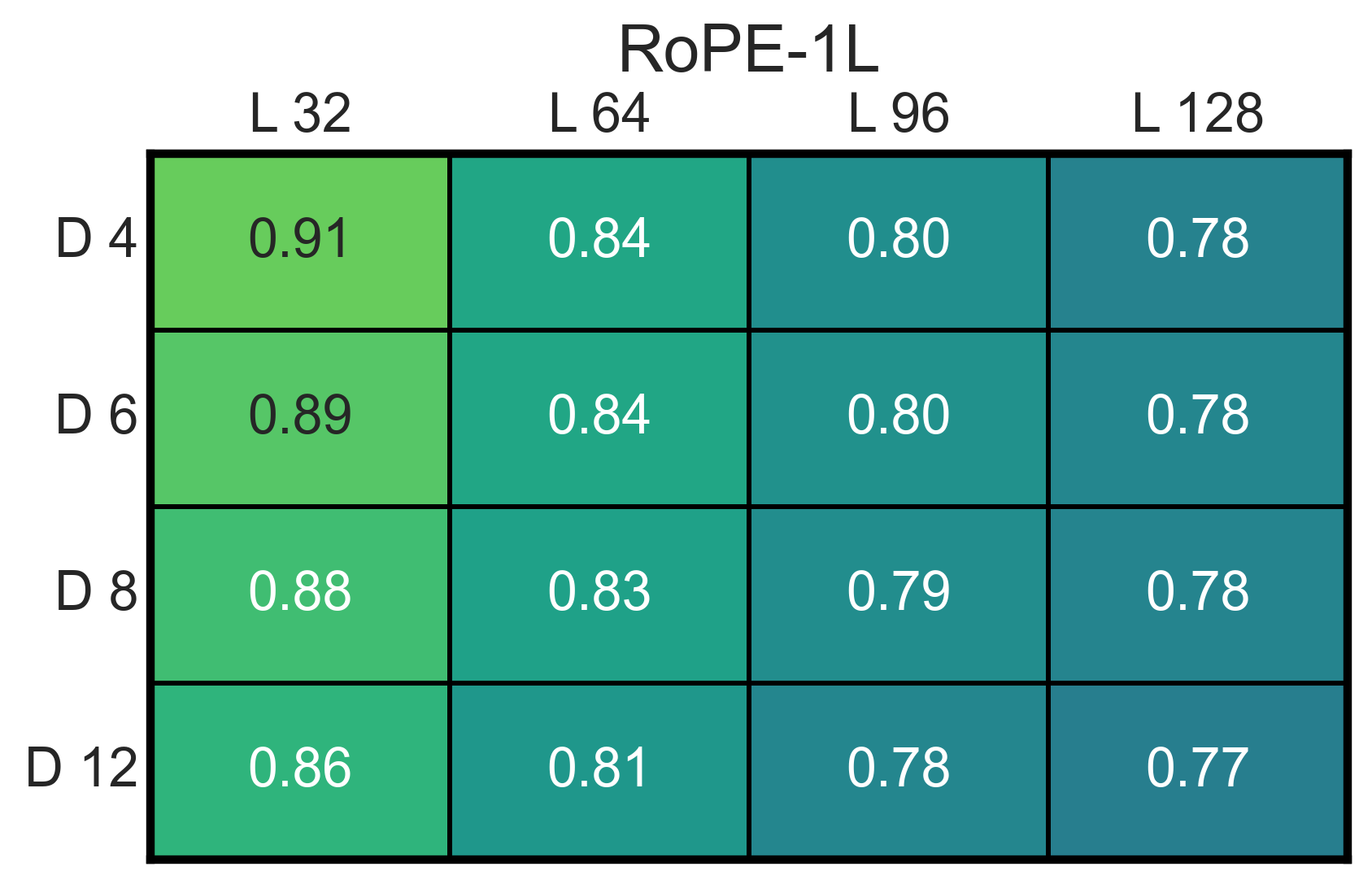}
        \caption{}
        \label{fig:taskdet:b}
    \end{subfigure}
    \hfill
    \begin{subfigure}[c]{0.30\textwidth}
        \centering
        \includegraphics[width=\textwidth]{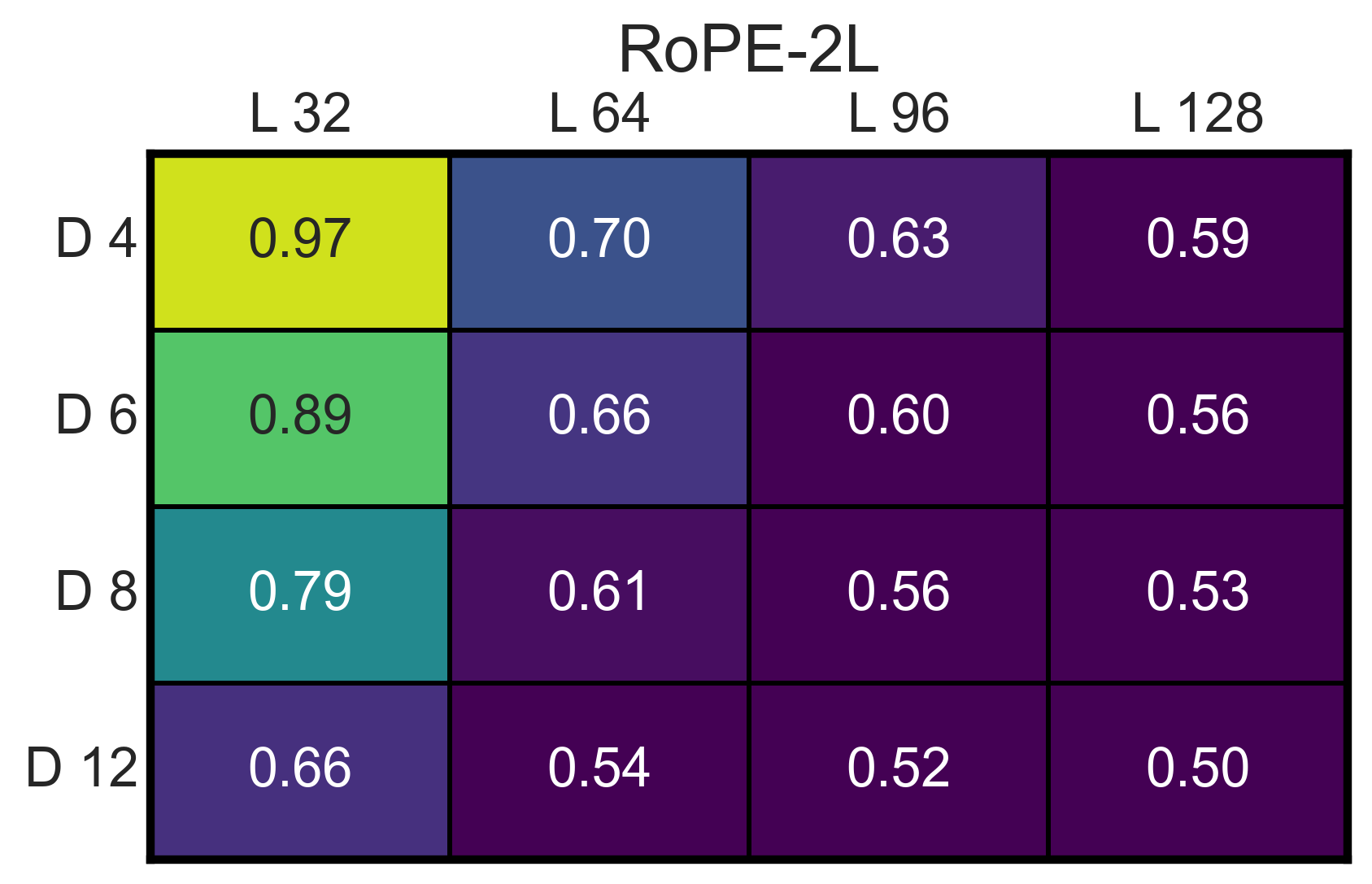}
        \caption{}
        \label{fig:taskdet:c}
    \end{subfigure}
    \hfill
    \begin{subfigure}[c]{0.30\textwidth}
        \centering
        \includegraphics[width=\textwidth]{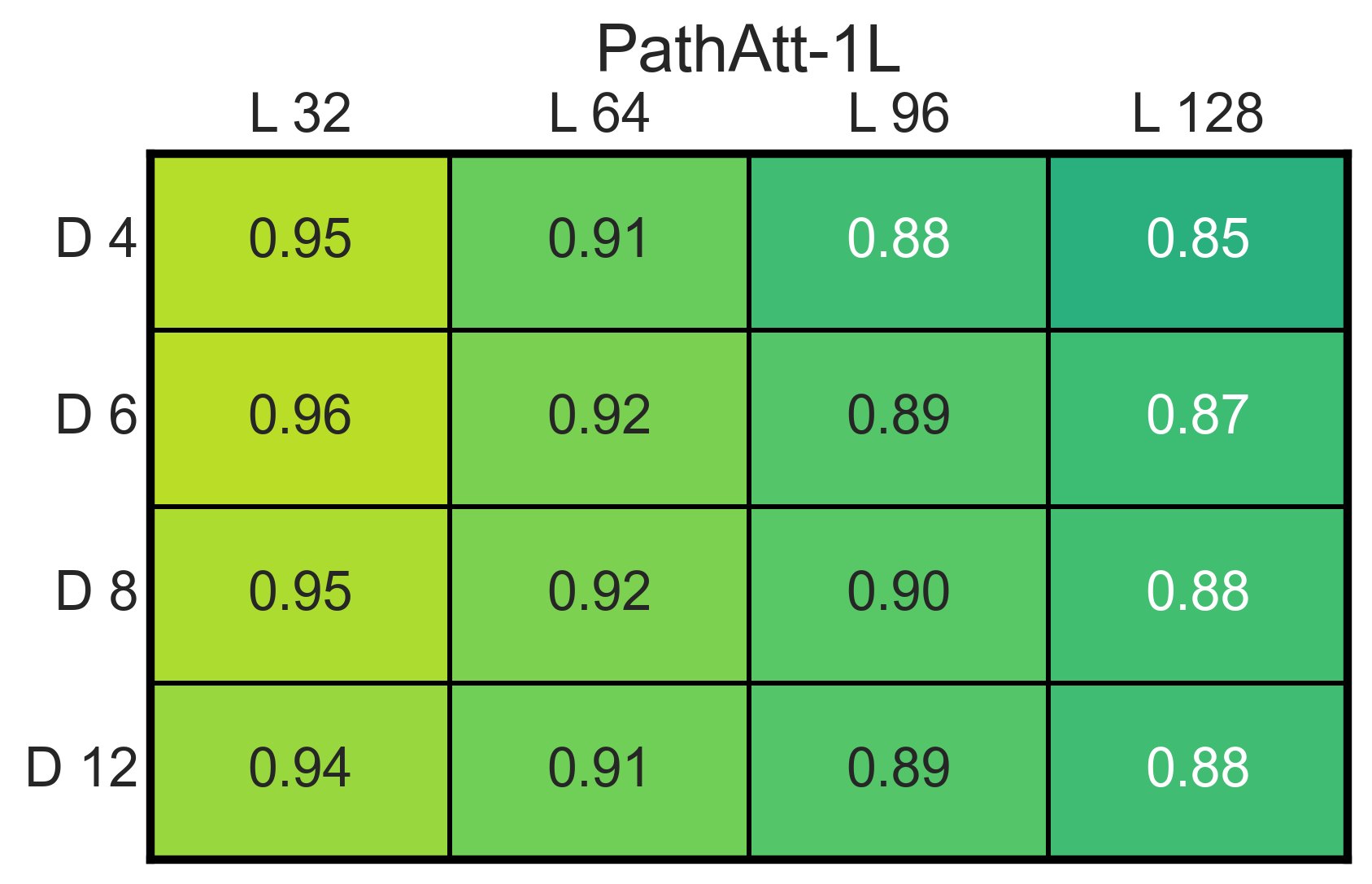}
        \caption{}
        \label{fig:taskdet:b}
    \end{subfigure}
    \hfill
    \begin{subfigure}[c]{0.30\textwidth}
        \centering
        \includegraphics[width=\textwidth]{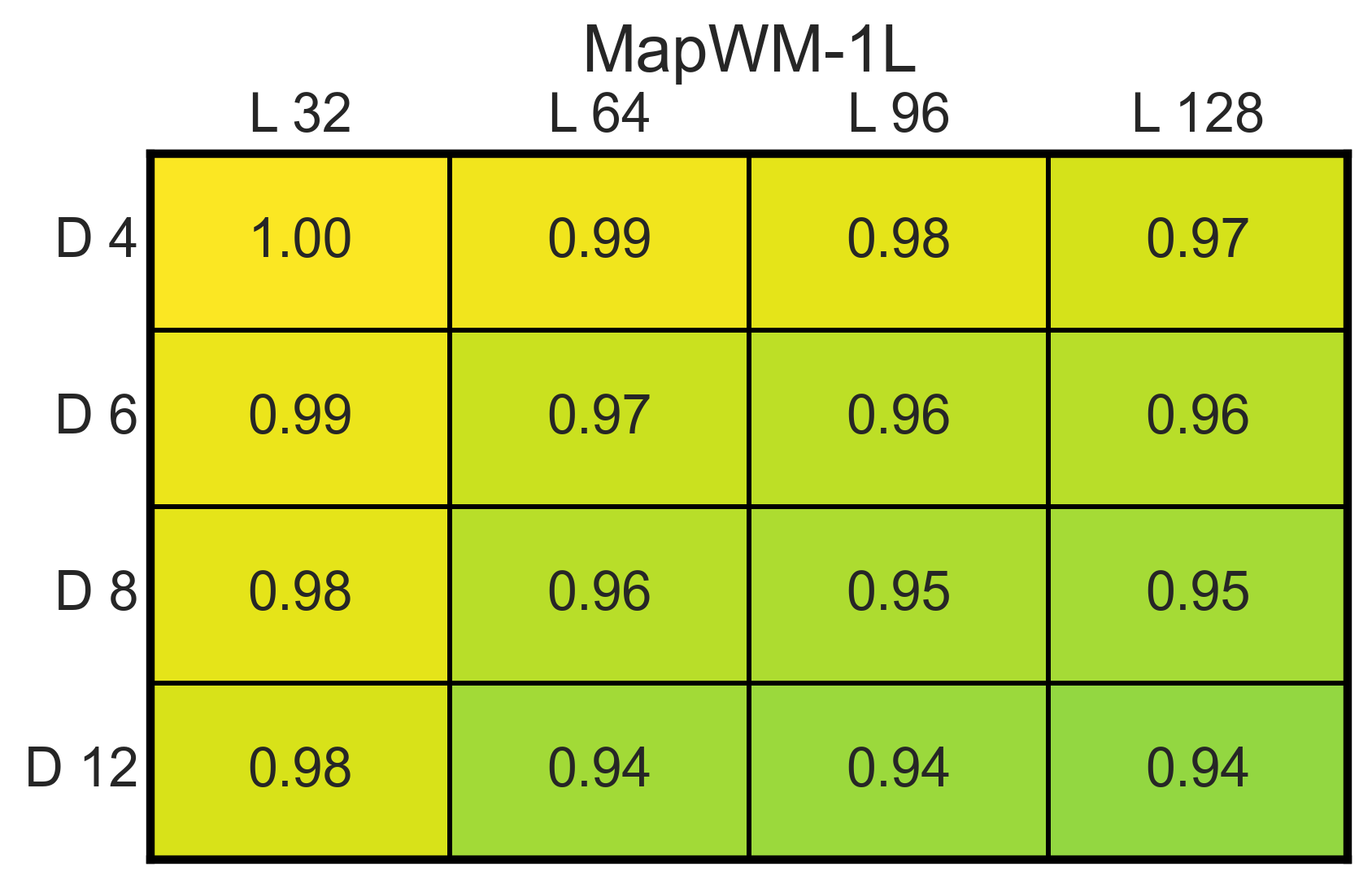}
        \caption{}
        \label{fig:taskdet:c}
    \end{subfigure}
    \hfill
    \begin{subfigure}[c]{0.30\textwidth}
        \centering
        \includegraphics[width=\textwidth]{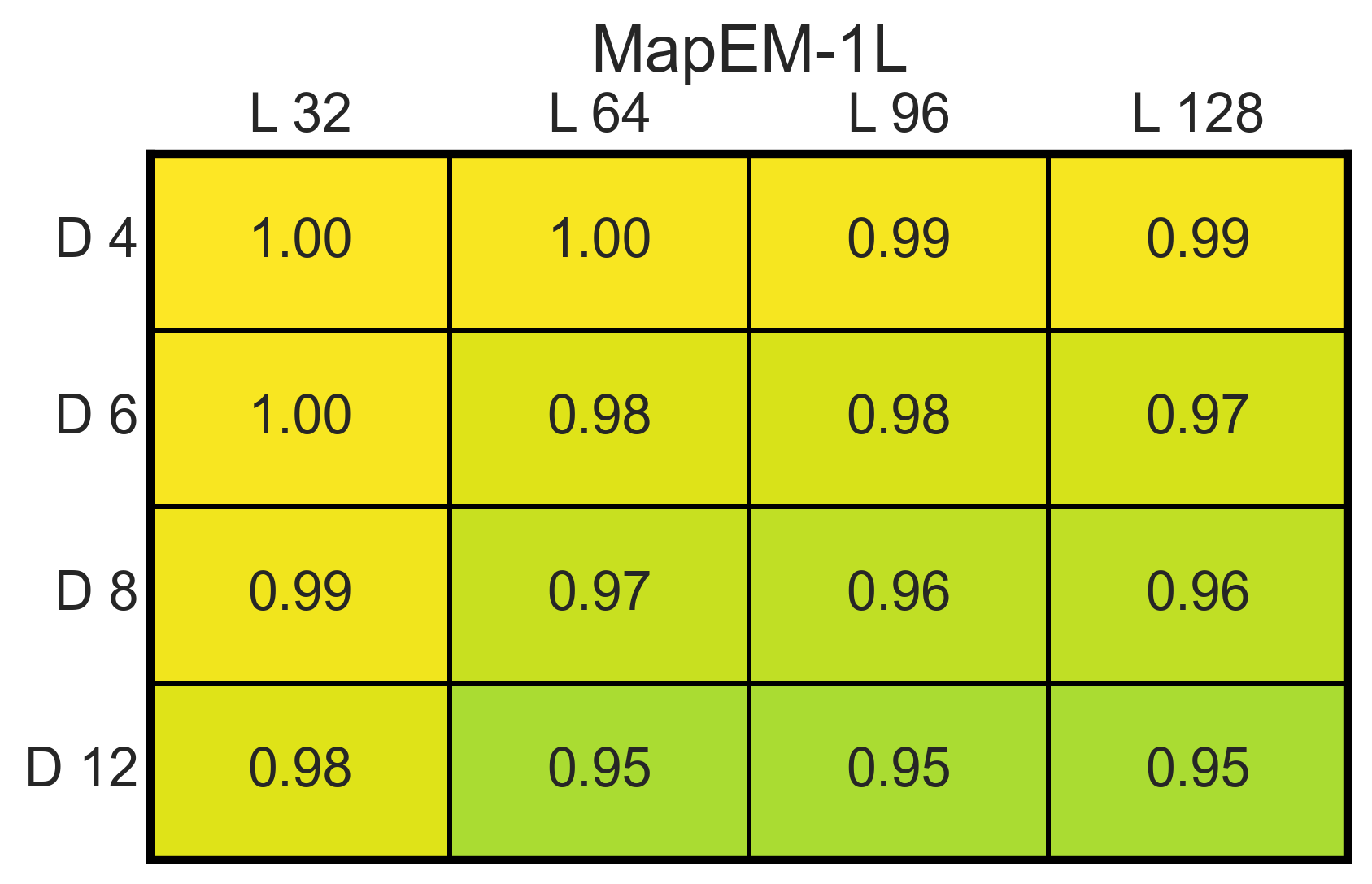}
        \caption{}
        \label{fig:taskdet:c}
    \end{subfigure}
    
    \caption{\textbf{Navigation intuition and Dyck-2 generalization performances} \textbf{(a)} The model only receives a sequence of symbols (tokens) and must understand their meaning. Notably, it needs to understand that only action tokens (arrows) update the cognitive map. Next-token prediction loss is computed on all tokens, but accuracy is computed only when the model comes back to a previously visited location. \textbf{(b-f)} Model's length and depth generalization on Dyck-2 generalization. Only MapFormers reach perfect performances on the training set (L 32, D 4) and maintain strong performances in the hardest OOD setting (L 128 D 12).}
    \label{fig:taskdet}
\end{figure}

At each step, the model receives one token at a time and must predict the next one. The full sequence consists of alternating actions and observations:
\[
s = (s_1, s_2, \dots, s_{2n}) = (a_1, o_1, \dots, a_{n}, o_n)
\]
where $a_t$ is an action drawn from a finite set of cardinal directions — $\{\rightarrow, \leftarrow\}$ in 1D; $\{\uparrow, \downarrow, \rightarrow, \leftarrow\}$ in 2D — and $o_t$ an observation associated with the agent's current location, drawn from a finite vocabulary of $K=16$ objects. Like in the selective-copy task, we add a blank token $B$ for empty grid locations.

Crucially, the model is not told which tokens are actions and which are observations—it sees a flat sequence of symbols and must learn this distinction on its own from the sequence alone. The model is trained to predict the next token at each step, but accuracy is computed only when the agent revisits a location, since these are the only positions whose observations are predictable with certainty (fig.~\ref{fig:taskdet:a}).

Because observations are statistically uncorrelated, the only way to solve this task is to understand its hidden, stable topology. For this, the model needs to:

\begin{enumerate}
    \item \textbf{Disentangle actions from observations } the model isn't told which tokens should update the cognitive map (actions) or leave it untouched (observations)
    \item \textbf{Learn the meaning of each action } the model must for example understand that opposite actions cancel each other
    \item \textbf{Path integrate } the model must sequentially compose these actions and update the state accordingly
    \item \textbf{Bind observation and position } the model must \textit{bind} an observation to a precise position to accurately recall this observation on revisits
\end{enumerate}

When generating trajectories, to favor grid exploration, we uniformly sample a direction and the number of steps $k$ to pursue in that direction $(1\leq k \leq 10)$ in 1D, $(1\leq k \leq 3)$ in 2D. Otherwise, agents would be statistically stuck in small environments.

For 1D/2D grid navigation, we trained all models on sequences of $l=128$ steps and a grid size of $l_{grid}=64$, and a probability of sampling an empty token $p_{empty}=0.5$—which should influence the sampling probabilities of models unable to build a cognitive map. We tested length generalization on two OOD datasets, by either reducing or increasing sequence size and world size: OOD-d (dense): $l=64, p_{empty}=0.2, l_{grid}=32$ and OOD-s (sparse): $l=512, p_{empty}=0.8, l_{grid}=128$.

Given the simplicity of the task, we assume that our models won't need more than one layer and one head to solve it, and as such, all our models are single head, single layer transformers, with a head size of $h=64$. To validate that without the appropriate structural bias, models cannot build a cognitive map, we also train deeper models for all baselines, with up to 4 layers and two heads. Since deeper models take longer to train and baselines are longer to converge, we first trained the deepest models until convergence, and used their training hyperparameters to train all other models.

All models were therefore trained on 700 000 sequences, with AdamW optimizer \cite{loshchilov2019decoupledweightdecayregularization}, a linear learning rate decay, a base learning rate $lr=3e^{-4}$, weight decay $0.05$ and a batch size of $128$ on a single GPU. Loss is computed on all tokens, but accuracy is only computed when coming back to a previously visited location (see fig.~\ref{fig:taskdet:a} for conceptual explanation).

\subsection{Dyck-2 valid continuations}\label{ssec:dyck_explained}

Dyck-2 is a context-free formal language over two bracket types, () and [], where every valid string corresponds to a correctly nested sequence of brackets. For example, ([]) and [([])()] are valid, but ([)] isn't, since only the last opened bracket can be closed. Dyck-2 is designed to test a model's ability to handle nested hierarchies, which is a core feature of natural language.

To solve this task, the model must predict for each token the set of \textit{valid continuations}. Since a new bracket can always be opened, ( and [ are always valid continuations. When depth is non-zero, the model is also allowed to close the last opened bracket. For example, in the following sequence ([]([][]?, valid continuations include (, [ and ). In other words, the model must implicitly maintain a stack of unclosed brackets through the sequence: each opening token pushes onto the stack, each closing token pops one out of it, and the top of the stack determines which closing bracket is currently legal.

The external memory module of Memory-augmented RNNs allows them to emulate a stack and generalize to longer and deeper sequences unseen during training~\cite{suzgun2019memoryaugmentedrecurrentneuralnetworks}, while standard transformers and LSTMs fail to generalize beyond their training distribution~\cite{bhattamishra-etal-2020-practical}, showing that even gating is insufficient to emulate a stack.

To control for length and depth, and therefore carefully tune our OOD samples, we generate sequences by first fixing a target length $L$ and maximum depth $D$ that must be reached. To achieve this, we modify the sampling distribution at each step to ensure that depth $D$ is reached, and that the stack comes back to depth 0 at the end of the sequence.

Since there are more than one possible valid continuation for each token, we measure valid continuations using the F1 valid continuation metric for formal languages introduced in \cite{goodale-etal-2025-meta}. This metric is the harmonic mean of valid and "better-than" continuations, the latter controlling for degenerate models. Specifically, the "better-than" metric counts the amount of valid tokens which have a higher probability than the sum of all invalid tokens, at each time step:
\begin{itemize}
    \item \textbf{Valid continuations}: $P_{Val}(s) = \sum_{x\in Val(s)}P(x|s)$
    \item \textbf{Better than continuations}: $BT(s) = \frac{\sum_{x\in Val(s)}\left[ P(x|s) > \sum_{c\notin Val(s)}P(c|s)\right]}{|Val(s)|}$
    \item \textbf{F1}: $2\frac{P_{Val}(s).BT(s)}{P_{Val}(s)+BT(s)}$
\end{itemize}

Since previous work have demonstrated the ability of transformers to generate Dyck-2 using 2 layers and 2 heads, we defined this setup as our baseline configuration. However, since \textit{MapFormers} can use their positional encoding mechanism to emulate a stack (by moving forward or backwards), we also train models with one layer and one head of size $h=64$.

Like in the navigation task, we first optimized hyperparameters for our bigger models, and used these parameters to train all smaller ones. We trained all models on sequences of size $L=32$ and depth $D=4$ on 560 000 sequences, using a learning rate $\mathrm{lr}=1e^{-4}$, AdamWM optimizer with weight decay 0.01 and a cosine scheduler with warmup.

OOD sequences were generated by varying length up to $4\times$ the training length, and depth up to $D+12$. Generalization performances along these two dimensions are highlighted for a 2 layer RoPE and single layer \textit{Map}EM in fig.~\ref{fig:taskdet:b} and fig.~\ref{fig:taskdet:c}, highlighting the generalization capacities of MapFormers over baseline transformers.

\subsection{Language modeling}\label{ssec:language_modelingdetails}

\begin{figure*}[ht!]
    \centering
    
    \begin{minipage}[t]{0.45\textwidth}
        \centering
        \setlength{\tabcolsep}{3pt}
        \begin{tabular}{
            l
            S[table-format=1.2(1)]
            S[table-format=1.2(1)]
        }
        \toprule
         & \textbf{RoPE} & \textbf{MapWM} \\
        \midrule
            wh island                          & {$0.83_{\pm0.01}$}          & {$0.81_{\pm0.02}$} \\
            sentential subject island          & {$\textbf{0.39}_{\pm0.02}$} & {$0.32_{\pm0.03}$} \\
            left branch island simple q.       & {$0.52_{\pm0.10}$}          & {$\textbf{0.64}_{\pm0.06}$} \\
            left branch island echo q.         & {$0.55_{\pm0.02}$}          & {$0.54_{\pm0.03}$} \\
            complex NP island                  & {$0.59_{\pm0.03}$}          & {$\textbf{0.64}_{\pm0.00}$} \\
            adjunct island                     & {$\textbf{0.90}_{\pm0.02}$}          & {$0.87_{\pm0.01}$} \\
            \midrule
            overall island                     & {$0.63_{\pm0.03}$}          & {$0.64_{\pm0.02}$} \\
            overall BLIMP                      & {$0.78_{\pm0.03}$}          & {$0.79_{\pm0.02}$} \\
            \midrule
            \midrule
            OpenWeb val ppl                      & {$19.14_{\pm0.14}$}          & {$\textbf{18.79}_{\pm0.15}$} \\
        \bottomrule
        \end{tabular}
        \subcaption{}
        \label{tab:blimp}
    \end{minipage}
    \hfill
    \begin{minipage}[t]{0.45\textwidth}
        \centering
        \setlength{\tabcolsep}{3pt}
        \begin{tabular}{l c c}
            \toprule
             & \textbf{RoPE} & \textbf{MapWM} \\
            \midrule
            n layers      & 12       & 12 \\
            embed size    & 768      & 768 \\
            head size     & 64       & 64 \\
            rank size     & —        & 4 \\
            base freq     & —        & 1024 \\
            \midrule
            lr            & $5 e^{-4}$ & $5 e^{-4}$ \\
            batch size    & $12\times 4$       & $12 \times 4$ \\
            context size    & 1024     & 1024 \\
            weight decay  & 0.1      & 0.1 \\
            warmup iters  & 4000     & 4000 \\
        \bottomrule
        \end{tabular}
        \subcaption{}
        \label{tab:hparams}
    \end{minipage}
    
    \caption{\textbf{Language modeling.} (a) Comparing RoPE and \textit{Map}WM accuracies on BLIMP after pretraining on OpenWeb for $10^{11}$ tokens. Results on OpenWebText are averaged across 5 seeds ($\mathrm{p_{value}}<0.005)$. For BLIMP, results are reported on each Island sub-tasks, known to be challenging for language models. (b) Pretraining hyperparameters for both models.}
    \label{fig:language_modeling}

\end{figure*}

To assess the scalability of our method beyond controlled formal tasks, we pretrained \textit{Map}WM on the OpenWebText dataset and compared it to a RoPE baseline, using the nanoGPT repository \cite{karpathy2022nanogpt}. We trained both models over approximately $10^{11}$ tokens on 4 H100 GPUs, using a base learning rate of $5e^{-4}$, AdamWM optimizer with a weight decay of $0.1$. We averaged both model's performances over 5 runs and reported the validation perplexity through training. To reduce noise in the final validation perplexity, we report the mean of the last 5 validation values before averaging these across seeds in tab.~\ref{tab:blimp}. As we can see, \textit{MapWM} yields consistent perplexity improvements compare to the RoPE baseline on OpenWebText $19.14_{\pm0.14}$ vs $18.79_{\pm0.15}$ ($\mathrm{p_{value}}<0.005$).

\begin{wrapfigure}{r}{0.45\textwidth}
    \centering
    \includegraphics[width=0.45\textwidth]{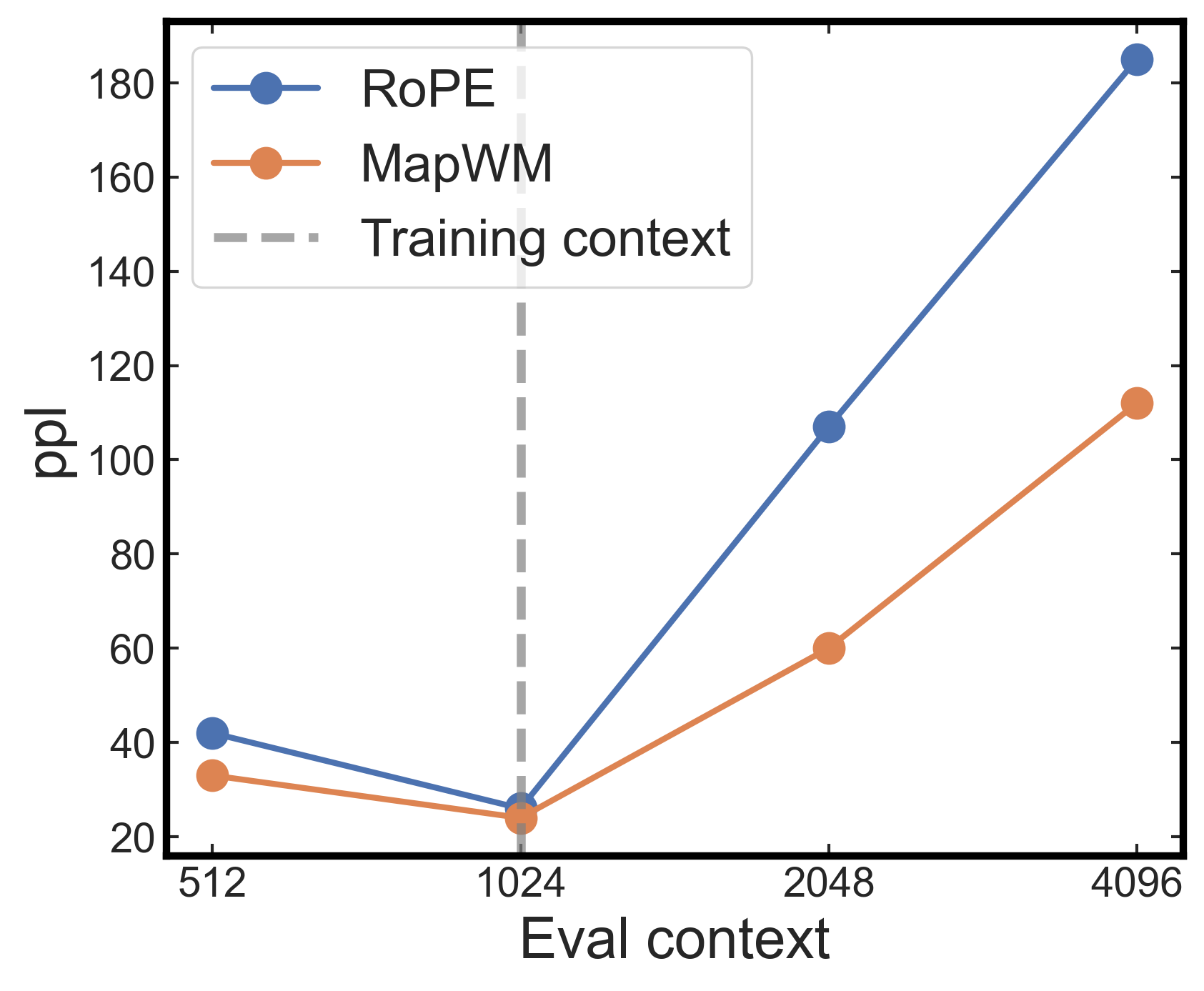}
    \caption{\textbf{Validation perplexity on Narrative QA for longer context length.} \textit{Map}WM improves over the RoPE baseline, though nos as much as other path-integrating models like CoPE or PathAtt \cite{golovneva2024contextualpositionencodinglearning, yang2026pathattentionpositionencoding}.}
    \label{fig:length_extrapolation}
    \vspace{-12pt}
\end{wrapfigure}

Because path-integrating models like CoPE and PathAtt \cite{golovneva2024contextualpositionencodinglearning, yang2026pathattentionpositionencoding} show strong generalization beyond their training length, we tested whether the same effect holds for \textit{Map}WM by measuring perplexity at varying context lengths on the NarrativeQA dataset \cite{kocisky-etal-2018-narrativeqa}, a standard benchmark for long-context understanding. As we can see in fig.~\ref{fig:length_extrapolation}, \textit{Map}WM improves over baseline at all tested lengths, but its perplexity still degrades sharply compared to the numbers reported for CoPE and PathAtt. We identify several design choices that likely contribute to this gap and that would require further experimentations:
\begin{enumerate}[noitemsep, topsep=0pt]
    \item In contrast to CoPE and PathAtt, our model can encode structures beyond 1D (tab.~\ref{tab:navigation_perf}), which enables length and depth generalization on Dyck languages (tab.~\ref{fig:dyck_gen:len} \& \ref{fig:dyck_gen:depth}), by encoding each bracket type on orthogonal dimensions. This expressivity comes at the cost of reducing the amount of rotation blocks available to encode each of the $r$ inner dimensions of $\Delta_t^{\mathrm{in}}\in \mathbb{R}^r$, which may limit precision in very long contexts.
    \item Since different brackets use orthogonal inner dimensions in Dyck languages (fig.~\ref{fig:dyck_gen:traj}) setting $r=4$ likely biases the model towards depth generalization, which is not directly probed here and may be of limited use for language models with a context length of $1024$, since multi-layer RoPE transformers already match the training distribution on Dyck. However, code modeling, where deep nested structures are more common, could be a natural setting to test this further.
    \item The angular velocities $\omega$ are learned, which might bias the model towards its training context length and keeping the angular velocities fixed during training could create an inductive bias towards long context. However, the fact that \textit{Map}WM outperforms RoPE in all tested contexts suggests that its positional encoding didn't just overfit the training context length, but learn to interpolate (path-integrate) between positions.
    \item Other inductive biases could be introduced in the architecture, such as a negative per-block diagonal matrix akin to Mamba SSMs \cite{gu2024mambalineartimesequencemodeling, dao2024transformersssmsgeneralizedmodels}, combined with the rotation matrix used in \textit{MapFormers}. This will remain compatible with parallel computation and the rotation prior could encode depth-wise structures, while the diagonal scaling could favor long contexts by mitigating the wrap-around effect of rotations.
\end{enumerate}

We additionally evaluated both models on BLIMP \cite{warstadt-etal-2020-blimp-benchmark}, a benchmark targeting 67 syntactic phenomena such as agreement, binding or island effects. Each sub-task consists of 1000 minimal pairs—sentences that differ by a single feature, where the model must assign a higher probability to the grammatical sentence. We selected this benchmark to test whether commutative path-integration might help disentangle syntax from semantic in language models. The results in tab.~\ref{tab:blimp} show that both RoPE and \textit{Map}WM perform on-par, with a 0.78 vs 0.79 overall accuracy, and no strong generalization gap is observed as in our synthetic tasks, which suggests that syntactic modeling likely requires more than path-integration.

One limitation of this analysis is that due to compute constraints, both RoPE and \textit{Map}WM are undertrained. Moreover, the angular velocity initialization likely requires more careful tuning to be properly understood at this scale, but doing so would require extensive pretraining experiments that were beyond the scope of this paper. Specifically, understanding the tradeoff between length and depth generalization in different tasks appears as an interesting research direction. BLIMP, on the other hand, suggests that syntax modeling probably requires more than input-dependent positional encoding—the impacts that non-commutativity (sec.~\ref{sec:non_commute}) would have on syntax is still unclear. More broadly, understanding when cognitive maps help language modeling needs further investigation.

\section{Neural analysis}\label{sec:neuralanaly_detail}

\subsection{2D navigation interpretability Analysis: Observations are Vectors, Actions are Matrices}\label{ssec:obs_vectors}

\begin{figure*}[ht!]
    \centering

    \begin{subfigure}[b]{0.38\textwidth}
        \includegraphics[width=\textwidth]{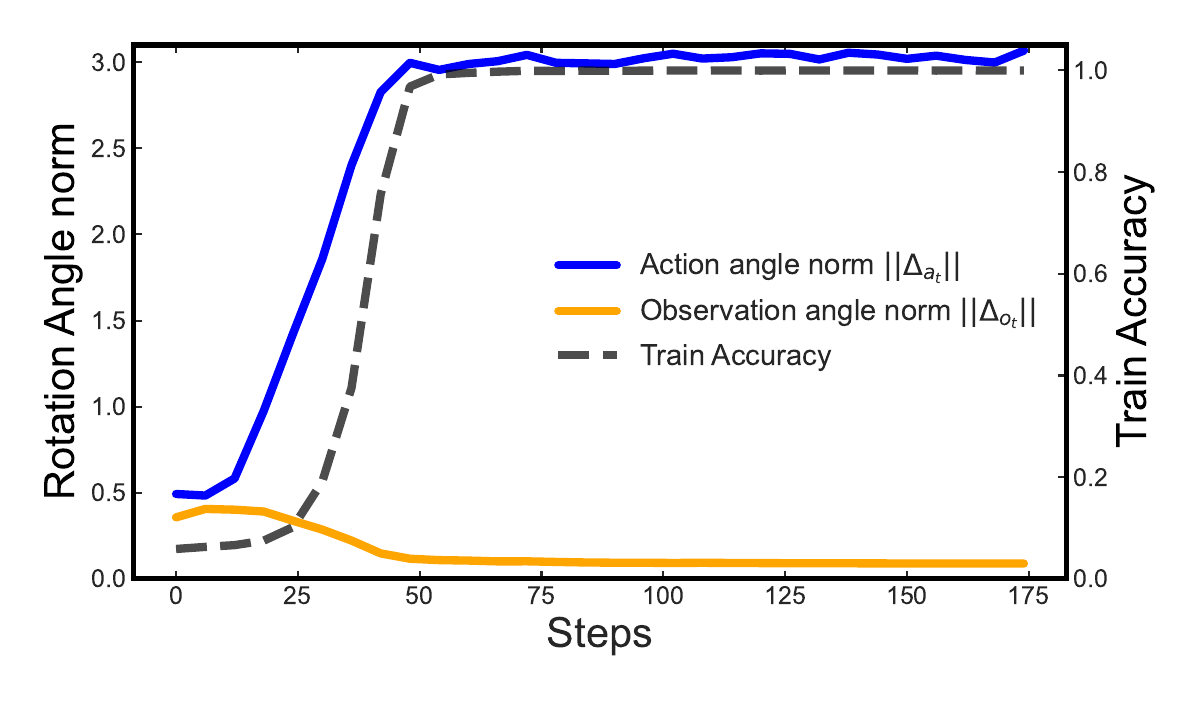}
        \caption{}
        \label{fig:learn_map:training}
    \end{subfigure}
    \hspace{1cm}
    \begin{subfigure}[b]{0.38\textwidth}
        \centering
        
        \setlength{\tabcolsep}{3pt}
        \begin{tabular}{l c c c c}
            \toprule
            & right & left & up & down \\
            \midrule
            right & 1.0 & \textbf{-1.0} & -0.6 & 0.7 \\
            left  & \textbf{-1.0} & 1.0 & 0.7 & -0.8 \\
            up    & -0.6 & 0.7 & 1.0 & \textbf{-1.0} \\
            down  & 0.7 & -0.8 & \textbf{-1.0} & 1.0 \\
            \midrule
            \midrule
            & & $||v_{o_t}||$ & $||v_{a_t}||$ & \\
            \midrule
            untrained & & 1.3 & 1.2 &\\
            trained & & \textbf{18.2} & 3.1 &\\
            \bottomrule
        \end{tabular}
        \caption{}
        \label{fig:learn_map:tab}
    \end{subfigure}

    \vspace{1em} 

    \begin{subfigure}[c]{0.38\textwidth}
        \centering
        \includegraphics[width=\textwidth]{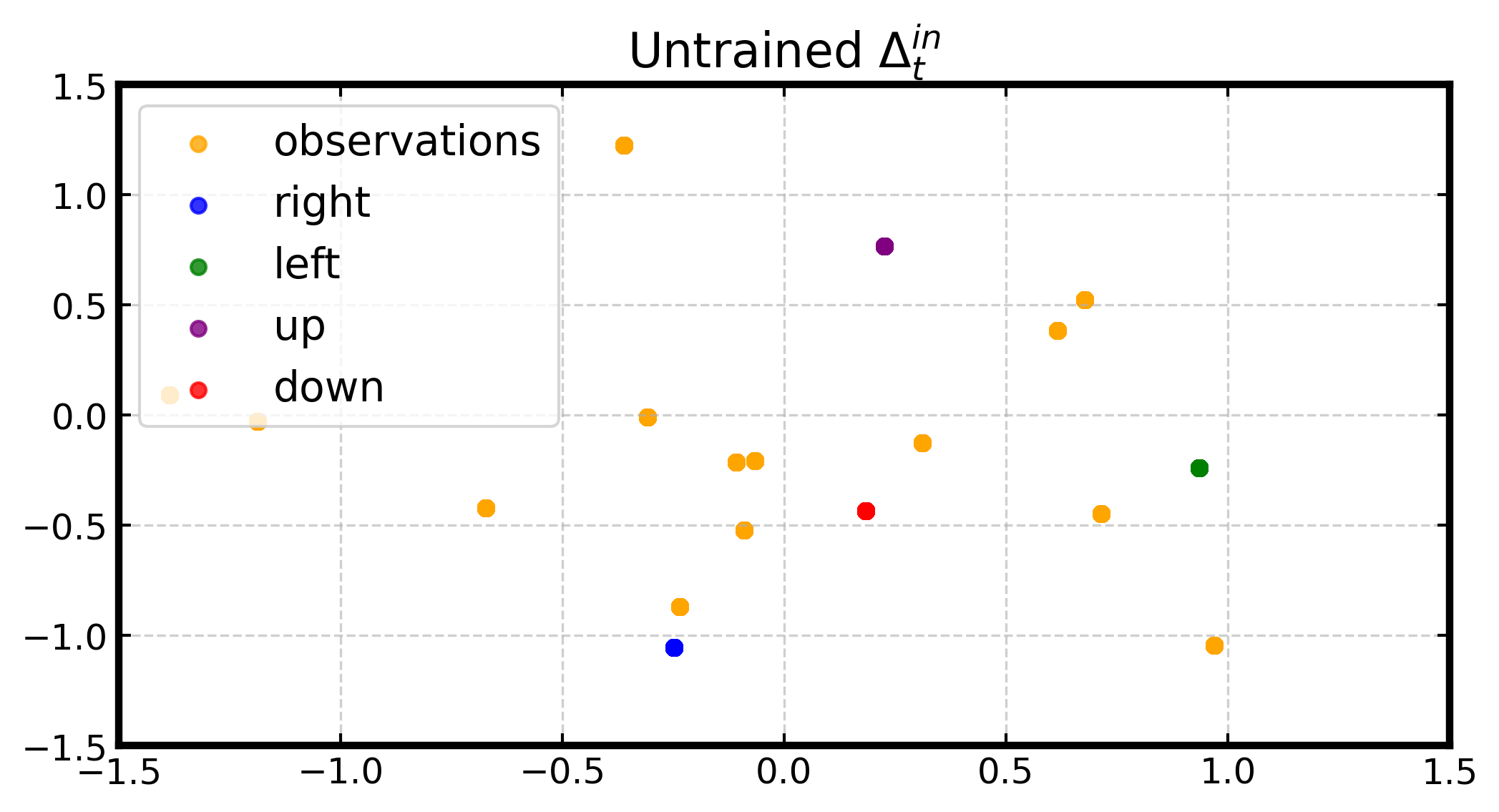}
        \caption{}
    \end{subfigure}
    \hspace{1cm}
    \begin{subfigure}[c]{0.38\textwidth}
        \centering
        \includegraphics[width=\textwidth]{figures/results/traineddelta2D.png}
        \caption{}
        \label{fig:learn_map:trained}
    \end{subfigure}
    
    \vspace{1em} 

    \begin{subfigure}[c]{0.38\textwidth}
        \centering
        \includegraphics[width=\textwidth]{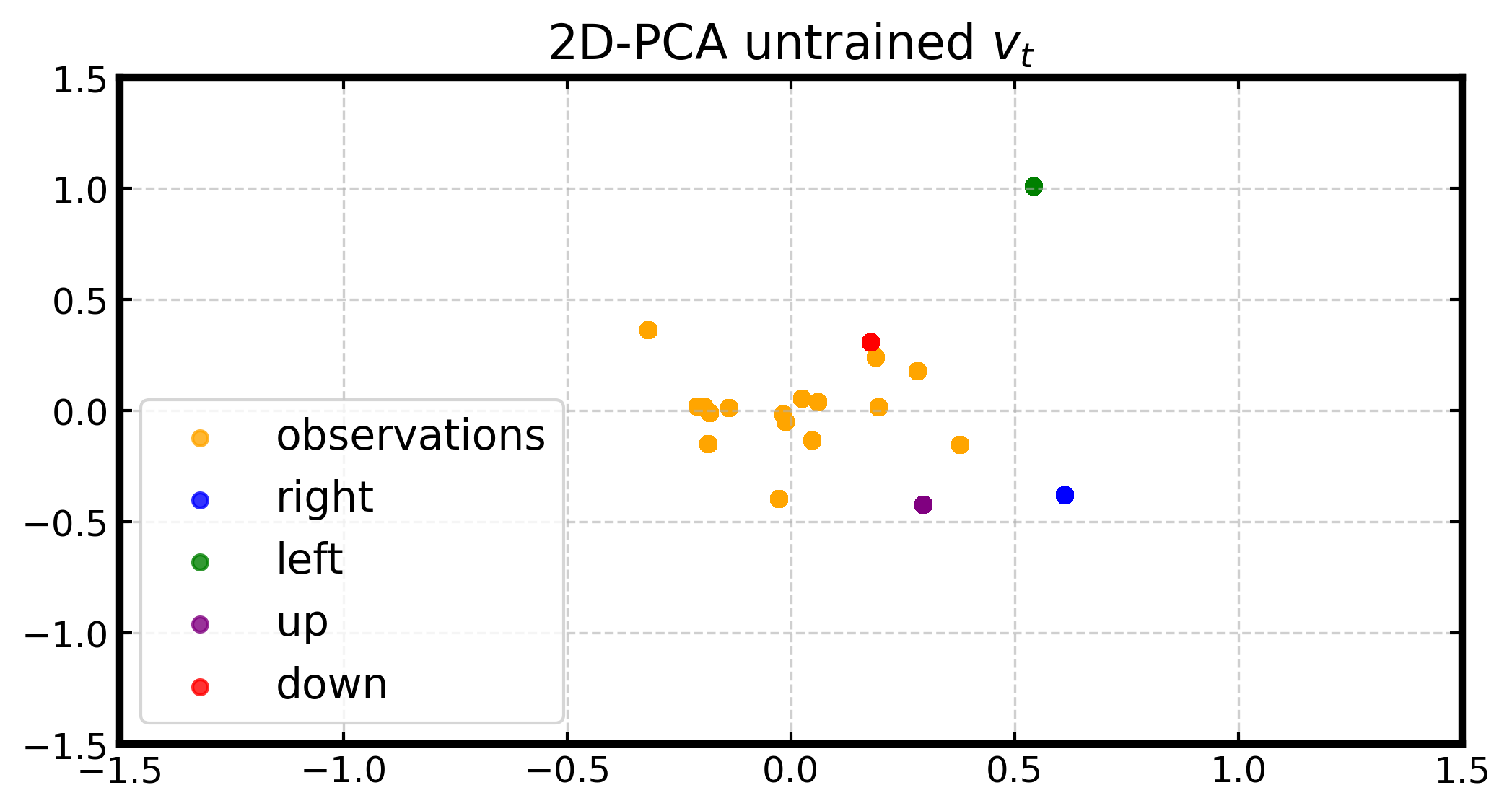}
        \caption{}
    \end{subfigure}
    \hspace{1cm}
    \begin{subfigure}[c]{0.38\textwidth}
        \centering
        \includegraphics[width=\textwidth]{figures/results/trained2Dvalues.png}
        \caption{}
        \label{fig:learn_map:trained_v}
    \end{subfigure}

    \caption{\textbf{Neural Analyses: Actions are Matrices and Observations are Vectors.} \textbf{(a)} Norm of rotation duration $||\Delta_t||$ vs Accuracy through training. \textit{MapFormers} reaches perfect accuracy as soon as it learns that action tokens $a_t$ (blue curve) update the agent's position while observation tokens $o_t$ (orange curve) leave it untouched via $0$-angle rotations, \textit{i.e.} $R_{\theta_o} \approx \mathrm{I}_n$ \textbf{(b)} (top table) Action's movement in the Lie algebra $\Delta_{a_t}^{in}$ cosine similarities. Opposite actions (right v left / up v down) cancel each other ($\cos(\Delta^{in}_{left},\Delta^{in}_{right})=-1$, but orthogonal dimensions are not projected onto orthogonal axes ($|\cos(\Delta^{in}_{left},\Delta^{in}_{up})|\gg 0$). Other constraints, such as bounded energy \citep{whittington2023disentanglementbiologicalconstraintstheory}, could be added to force disentanglement (bottom table). Once trained, the norm of value embeddings $v_t$ in the attention layer becomes much bigger for observations than actions $\left(||v_{o_t}||\gg ||v_{a_t}||\right)$, implying that only observations contribute in updating the state's content. 
    \textbf{(c)} An untrained \textit{Mapformer} projects observations and actions randomly \textbf{(d)} After training, observation tokens are all projected towards 0 (orange), while tokens representing opposite directions (left / right or up / down) are all projected to opposite directions. \textbf{(e)} An untrained \textit{Mapformer} projects values $v_t$ randomly. \textbf{(f)} After training, action tokens are projected towards zero.}
    \label{fig:learn_map}
\end{figure*}

In the tasks, some symbols (actions $s_t \in \{\uparrow, \downarrow, \rightarrow, \leftarrow\}$) must update the model's position on its cognitive map, while others should leave position the same (observations) and instead modify its content. The model must learn by itself which ones are actions and which ones are observations. How can we tell whether \textit{MapFormers} have learned this and therefore have learned a cognitive map?

For this, we first compared the learned integration durations $\Delta_t$, since they define how the cognitive map is updated. Fig.~\ref{fig:learn_map:training} compares model performance and integration durations during training. It shows that perfect generalization is achieved as soon as the model learns that action tokens $a_t$ trigger state rotations while observation tokens $o_t$ leave it untouched. 
Next, we studied the representation of action tokens, after training. Tab.~\ref{fig:learn_map:tab}-top \& fig.~\ref{fig:learn_map:trained} show that opposite actions (left / right or up / down) cancel each other, pointing to opposite direction, and representations of observations are all projected to zero. That is, only action tokens $a_t$ can update the cognitive map. Conversely, only observations need to update the state's content, since $||v_{o_t}||\gg ||v_{a_t}||$, as fig.~\ref{fig:learn_map:trained}-bottom $\&$ fig.~\ref{fig:learn_map:trained_v} confirm.

From a point of view of Lie-group theory, actions are represented within the space of the Lie algebra  $\mathfrak{g}$, whereas observations $v_t$
belong to the content space, the traditional latent space of transformers. Therefore, in \textit{MapFormers}, actions are matrices and observations are vectors, which is reminiscent of ideas presented by \citet{baroni-zamparelli-2010-nouns}, where nouns are vectors and adjectives are matrices, giving the model stronger OOD generalization compared to baseline models.

\subsection{MapFormers learn a continuous stack to solve Dyck-2}\label{ssec:dyck_neural}

\begin{figure*}[ht!]
    \centering

    \vspace{1em} 

    \begin{subfigure}[c]{0.45\textwidth}
        \centering
        \includegraphics[width=\textwidth]{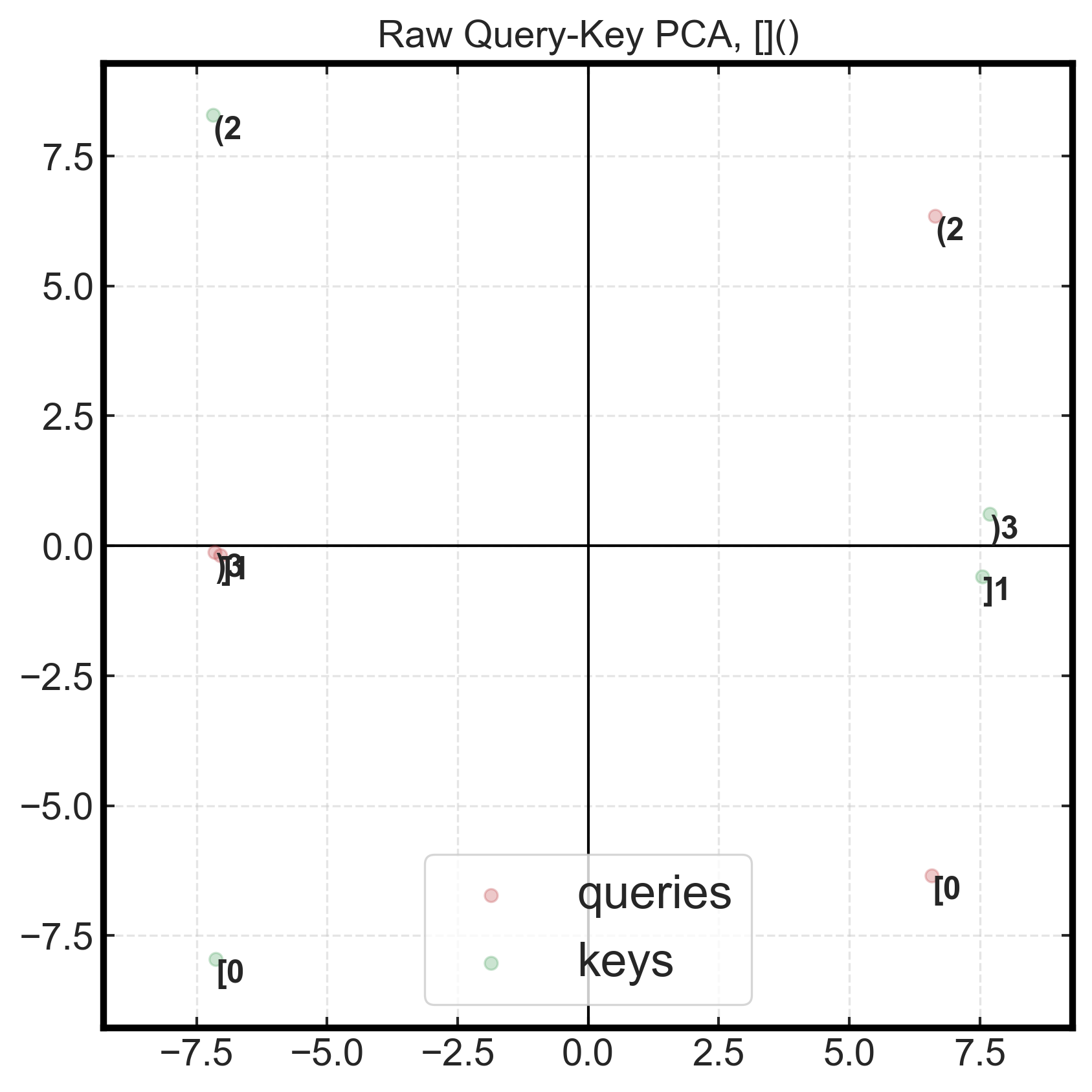}
        \caption{}
        \label{fig:dyck:pcaraw}
    \end{subfigure}
    \hspace{1cm}
    \begin{subfigure}[c]{0.45\textwidth}
        \centering
        \includegraphics[width=\textwidth]{figures/results/latent_traj.png}
        \caption{}
        \label{fig:dyck:trajectory}
    \end{subfigure}
    \vspace{1em} 

    \begin{subfigure}[c]{0.45\textwidth}
        \centering
        \includegraphics[width=\textwidth]{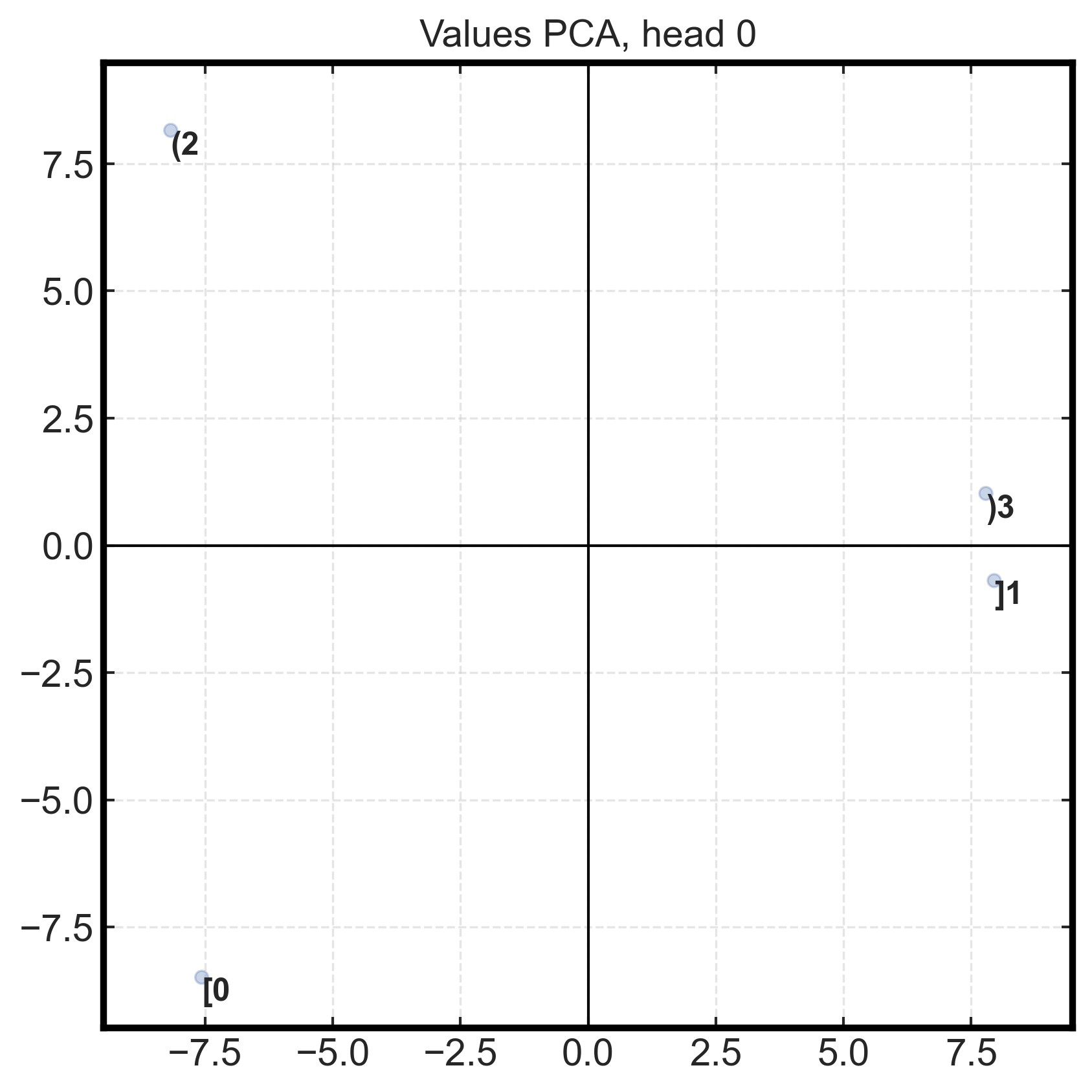}
        \caption{}
        \label{fig:dyck:pcavalues}
    \end{subfigure}
    \hspace{1cm}
    \begin{subfigure}[c]{0.45\textwidth}
        \centering
        \includegraphics[width=\textwidth]{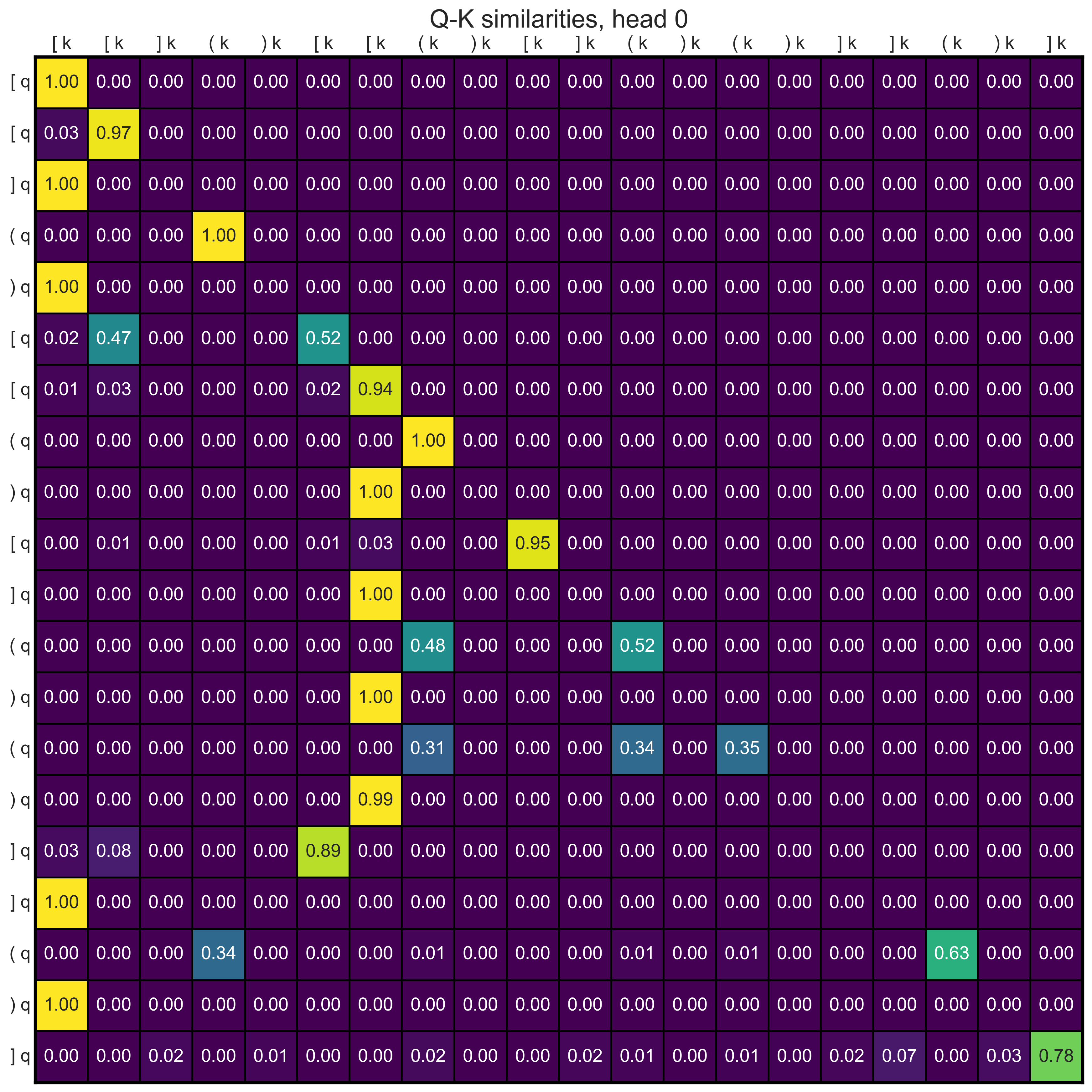}
        \caption{}
        \label{fig:dyck:attn}
    \end{subfigure}

    \caption{\textbf{Rotations drive closed brackets towards the closest opened one} \textbf{(a)} Before being rotated, closed queries $q_)$ or $q_]$ (red) attend indifferently to open keys $k_($ or $k_[$ (green), while open queries $q_($ or $q_[$ attend to their type, opened or close. \textbf{(b)} Latent trajectory $\text{cumsum} (\Delta_t)\in \mathbb{R}^2$, for trajectory $[]([]())$, starting in symbol $\mathbf{x}$. The model uses opposite directions for open/close brackets of same type and orthogonal dimensions for each bracket type—the latter facilitates the identification of last opened bracket at a given depth. \textbf{(c)} PCA of value representations for all symbols. The value space is organized along two dimensions, one encoding bracket identity, the other one open/close. \textbf{(d)} This mechanism allows MapFormers to produce sparse attention maps, where closed queries always attend to the last opened key, aligning the model's next token prediction with the true valid continuations.}
    \label{fig:dyck}
\end{figure*}

In Dyck-2, any bracket can open a new one, which means that opening a bracket of any type is always a valid continuation. The real task therefore consists of identifying if a closed bracket is considered a valid move and if yes, of what type. Without context, only a bracket of type \textit{a} can be closed by a bracket of same type, which means that valid continuations of an opened bracket of type \textit{a} are any new open bracket, or a closed bracket of type \textit{a} and conversely for a bracket of type \textit{b}.

Conversely, \emph{without context}, closed brackets ')' or ']' cannot be followed by another closed one, and we should expect this to be reflected within the model's valid continuations of closed symbols.


\begin{table}[h!]
    \centering
    \begin{tabular}{l c c c c}
        \toprule
        & {(} & {[} & {)} & {]} \\
        \midrule
        $\text{Val}_\theta( \,{(}\, )$ & 1 & 1 & 1 & 0 \\
        $\text{Val}_\theta( \,{[}\, )$ & 1 & 1 & 0 & 1 \\
        $\text{Val}_\theta( \,{)}\, )$ & 1 & 1 & 0 & 0 \\
        $\text{Val}_\theta( \,{]}\, )$ & 1 & 1 & 0 & 0 \\
        \bottomrule
    \end{tabular}
    \vspace{5pt}
    \caption{\textbf{Dyck-2 learned continuations.} Once trained, models learn that only opened brackets can produce a closed one of corresponding type $\text{Val}_\theta(\mathrm{close_a}|\mathrm{open_a})=1$. This will force attention to retrieve the value of last opened bracket in order to produce the next token.}
    \vspace{-0.5cm}
    
\end{table}\label{tab:dyck_continu}

To compute the valid continuations of each \emph{uncontextualized} symbol, we fed the model $f_\theta$ with a sequence of size 1 containing only the symbol $s\in\left\{ (, [, ), ]\right\}$ of interest, forcing the model to only attend to this specific token and return its raw value. We then approximated its valid continuations $\text{Val}_\theta(s)$ under model $f_\theta$ as follows:
\[
\text{Val}_\theta(s) := f_\theta(s) > 0 \in \llbracket 0, 1 \rrbracket ^4
\]
As expected, in tab.~\ref{tab:dyck_continu} we can see that the model returns positive logits only on valid continuations of uncontextualized symbols. Since the values carry this information, we also computed in fig.~\ref{fig:dyck:pcavalues} a PCA of the learned values $V$ of the model for each symbol in the vocabulary. The model splits them on two dimensions, the first dimension focusing on open/close, while the second dimension focuses on bracket identity.

At all depths but $D=0$, closed brackets are allowed to close the last open bracket. The attention allows this by rotating a closed query $q_{)}$ or $q_{]}$ towards the last open key $k_{(}$ or $k_{[}$ (fig.~\ref{fig:dyck:attn}). Since attention maps are perfectly sparse, the retrieved value will match the identity of the last opened bracket, and so will the valid continuations for the next token.

To allow for such a mechanism, the model \textit{recycles} its cognitive map in order to represent a continuous stack. The model uses two orthogonal dimensions for bracket identity (fig.~\ref{fig:dyck:trajectory}), with open/close of the same bracket identity mapping to opposite directions.

\subsection{\textit{Map}EMs are computationally more demanding but scale better than \textit{Map}WMs}\label{ssec:recall_scaling}

Compared to \textit{MapFormer}-WM, \textit{MapFormer}-EM queries its memories with the conjunction of observational and structural features, which requires a second pool of neurons $p_t\in \mathbb{R}^d$ as well as computing two attention matrices $A_X$ and $A_P$ separately. This offers sparser and higher dimensional features $g_t = \text{vec}(x_t.p_t^\top)\in\mathbb{R}^{d^2}$ (fig.~\ref{fig:cogmap_ssm:a}) but doubles the computational load of the attention operation, which is already the main bottleneck in scaling transformers.

\begin{figure*}[ht!]
    \centering

    \begin{subfigure}[c]{0.30\textwidth}
        \centering
        \includegraphics[width=\textwidth]{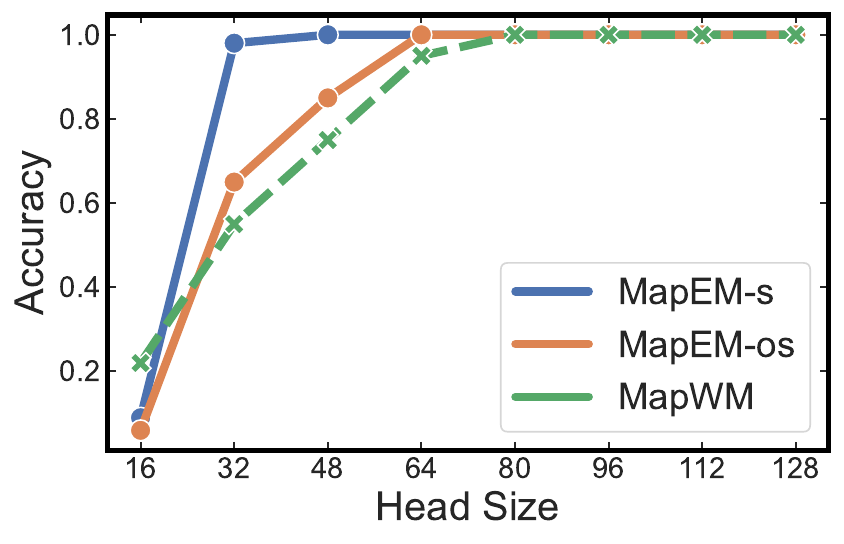}
        \caption{}
        \label{fig:model_scaling:a}
    \end{subfigure}
    \hfill
    \begin{subfigure}[c]{0.30\textwidth}
        \centering
        \includegraphics[width=\textwidth]{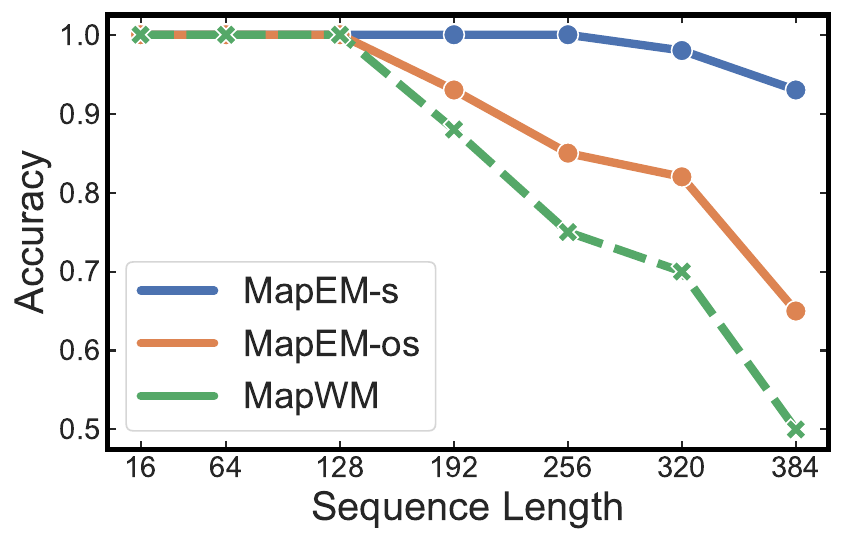}
        \caption{}
        \label{fig:model_scaling:b}
    \end{subfigure}
    \hfill
    \begin{subfigure}[c]{0.30\textwidth}
        \centering
        \includegraphics[width=\textwidth]{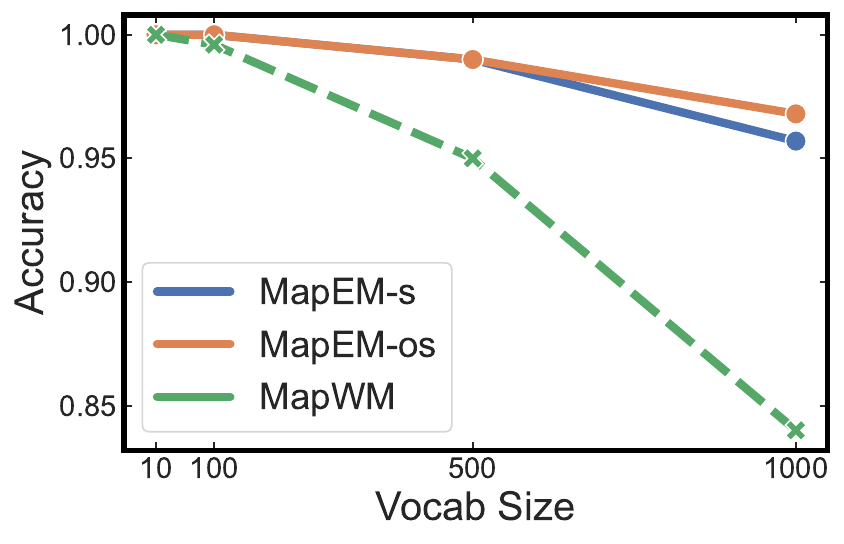}
        \caption{}
        \label{fig:model_scaling:c}
    \end{subfigure}
    
    \caption{\textbf{Behavioral Analyses: \textit{Map}EM scales better than \textit{Map}WM.} Comparison of \textit{Map}EM-s, \textit{Map}EM-os, and \textit{Map}WM robustness across model size, sequence length, and vocabulary size. \textbf{(a)} Train accuracy vs. head size (fixed sequence length $l=256$, vocabulary size $v=16$). \textit{Map}EM requires fewer units than \textit{Map}WM to achieve high performance. \textbf{(b)} Train accuracy vs. sequence length (fixed head size $h=48$, vocabulary $v=16$). EM models exhibit greater robustness to increasing sequence lengths. \textbf{(c)} Train accuracy vs. vocabulary size (fixed head size $h=32$, sequence length $l=16$). EM models outperform WM models in retaining an increasing number of items.}
    
    \label{fig:model_scaling}
\end{figure*}

This factorization in two separate pools of neurons should allow EM to be more efficient than WM, as in the former, neurons specialize for either position or observation. To test this, we conducted three experiments to understand the effect of sequence length, model size and vocabulary size on recall performance in 2D navigation, by varying each time one parameter while keeping the others fixed. Fig.~\ref{fig:model_scaling} shows that, indeed, EM models scale better than WM, consistently with previous results \cite{Whittington2025}. Also, \textit{Map}EM-s (blue curves), which relies on position alone, is more robust to both sequence length (fig.~\ref{fig:model_scaling:a}) and model size (fig.~\ref{fig:model_scaling:b}), since sensory features only add noise in this task. Last, having a sufficiently large number of neurons is crucial for both learning a cognitive map (fig.~\ref{fig:model_scaling:a}, models fail to converge with $h=16$) and memorizing large sets of objects (fig.~\ref{fig:model_scaling:c}). In sum, these results demonstrate the superior capacity of \textit{Map}EM models in recall tasks, at the cost of doubling the attention computation.


\section{High-dimensional Navigation and Non-Commutativity}

\subsection{3D and 5D navigation}\label{sec:3D}
Because translations in $k$ dimensions are just $k$ independent $1$D-translations on orthogonal axes ($\mathrm{SO}(2)^k$), our models should theoretically be able to track position on these structures -- with rotation blocks representing movement on one specific dimension -- but should require a higher number of neurons to represent movement on all these orthogonal axes.

We train all models with a head size of 64. In 3D, we choose a sequence length of 256 and a grid size of 16. The task being particularly hard in 5D, we decided to reduce the sequence length to 16 and the grid size to 5 and report the results in tab.~\ref{tab:newref3D5D}. Moreover, for each model, we set the dimension of its inner rank to the dimension of the world, \textit{i.e.}: $\Delta^{\mathrm{in}}\in \mathbb{R}^k,\, k=3,5$.

\begin{table}[h]
    \centering
    \begin{tabular}{l | c c c | c c c} 
        \toprule
        \textbf{} & \multicolumn{3}{c|}{\textbf{3D navigation}} & \multicolumn{3}{c}{\textbf{5D Navigation}} \\
        \cmidrule(lr){2-4} \cmidrule(lr){5-7} 
        
        \textbf{} & \textbf{IID} & \textbf{OOD-d} & \textbf{OOD-s} & \textbf{IID} & \textbf{OOD-d} & \textbf{OOD-s} \\
        \midrule
        
        RoPE (1L) & {$0.39_{\pm0.02}$} & {$0.36_{\pm 0.03}$} & {$0.35_{\pm 0.01}$} & {$0.74_{\pm 0.01}$} & {$0.45_{\pm 0.01}$} & {$0.33_{\pm 0.02}$} \\
        
        CoPE (1L) & {$0.78_{\pm 0.03}$} & {$0.82_{\pm 0.02}$} & {$0.74_{\pm 0.04}$} & {$0.94_{\pm 0.01}$} & {$0.80_{\pm 0.01}$} & {$0.69_{\pm 0.01}$} \\
        
        \textbf{\textit{Map}WM} & {$\textbf{1.00}_{\pm 0.00}$} & {$0.95_{\pm 0.03}$} & {$0.94_{\pm 0.02}$} & {$0.75_{\pm 0.06}$} & {$0.50_{\pm 0.05}$} & {$0.35_{\pm 0.08}$} \\
        \textbf{\textit{Map}EM-os} & {$\textbf{1.00}_{\pm 0.00}$} & {$\textbf{0.99}_{\pm 0.00}$} & {$\textbf{0.99}_{\pm 0.01}$} & {$\textbf{1.00}_{\pm 0.00}$} & {$\textbf{0.98}_{\pm 0.01}$} & {$\textbf{0.84}_{\pm 0.04}$} \\
        \textbf{\textit{Map}EM-s} & {$\textbf{1.00}_{\pm 0.00}$} & {$\textbf{1.00}_{\pm 0.00}$} & {$\textbf{0.98}_{\pm 0.01}$} & {$\textbf{1.00}_{\pm 0.00}$} & {$\textbf{1.00}_{\pm 0.00}$} & {$\textbf{0.87}_{\pm 0.02}$} \\
        \bottomrule
    \end{tabular}
    \vspace{5pt}
    \caption{\textbf{3D - 5D grid navigation accuracy.} $2\times 2$ block-diagonal rotation matrices can represent translations in 3D and 5D, since it equivalent to (3 or 5) 1D orthogonal translations. However, performances quickly degrade, since the amount of dimensions to represent with a fixed neuron budget grows exponentially. Notably, in 5D, \textit{Map}WM fails to learn the task and perform a par with RoPE, while \textit{Map}EM, which scale better in recall tasks, solve it even though performances start to degrade on longer sequences.}
    \label{tab:newref3D5D}
\end{table}

As we can see, all of our models converge in dimension 3, confirming that 2D-rotations are enough to learn simple translations in higher dimension. However, our WM models never manage to learn a cognitive map in 5D, and actually performances comparable to RoPE and much worse than CoPE. This is due to the difficulty of encoding each axis accurately in high dimension, given a fixed neural budget. EM models do converge, showing once again their increased robustness compared to WM on recall tasks.

Note that in 5D, because we keep the world size and sequence length small, CoPE asymptotically approaches a perfect score on the training set but fails to generalize OOD, while EM models solve it with $6$ times less training data and generalize, highlighting the difference between learning of a cognitive map versus statistical heuristics. Models learning a cognitive map achieve robust generalization with much fewer data points, something that has been observed when comparing learning efficiency in humans versus AI models.

\subsection{Non-Commutative Cognitive Maps}\label{sec:non_commute}

The ability to learn non-commutative relations is fundamental for most cognitive maps. For instance, family relations like \textit{mother} and \textit{father} do not commute: applying \textit{father} then \textit{mother} gives the maternal grandfather, while \textit{mother} then \textit{father} gives the paternal grandmother. This implies that the learned matrices must satisfy: 
\[\mathbf{W}_{\mathrm{mother}}\mathbf{W}_{\mathrm{father}}\neq \mathbf{W}_{\mathrm{father}}\mathbf{W}_{\mathrm{mother}}\]
Aside from non-commutativity, we want neural activity to remain bounded under sequential matrix multiplication. This favors orthogonal matrices in $\mathrm{O}(n)$, which are norm-preserving and therefore constrain the representation to a compact (bounded) portion of space. Akin to trajectories on 2D grids which wrap around on the torus, non-commutative structures like family trees will therefor be encoded accurately up to a finite depth.

To do so, we can approximate non-commutative cognitive maps with $M$ generators using the compact group $\mathrm{SO}(n)$, which has $K=\frac{n(n-1)}{2}\geq M$ \textit{skew-symmetric} generators $\{S_i\}_{1\leq i\leq K}$. Intuitively, each canonical relation ("mother", "father", ...) will be represented by a distinct generator $S_i$, but since the generators do not commute in general ($S_iS_j \neq S_jS_i$), we loose the ability to perform additive path-integration:
\begin{equation}
    R^n_{\theta} = \exp \left(\sum_{i=1}^{n(n-1)/2} \theta_{i} S_{i} \right) \neq \prod_{i=1}^{n(n-1)/2}\exp \left( \theta_{i} S_{i} \right)
\end{equation}
This implies that path integration cannot be performed via the exponential of a sum anymore and can only be achieved via a sequential matrix product, scaling linearly with sequence length $L$. Consequently, we introduce non-commutative models (EM-NC), with a basis of $K=n(n-1)/2$ canonical \textit{skew-symmetric} matrices $\{S_i\}_{1\leq i \leq K}\in \mathbb{R}^{n_b \times b\times b}$—these matrices are not learned, only the angular velocities $\omega \in \mathbb{R}^{n_h \times K \times n_b}$ are.

Input $x_t$ is projected towards $\Delta_t \in \mathbb{R}^{n_h\times K \times n_b}$ and combined with $\omega$ to build the final rotation angle $\theta :=\omega \Delta_t \in \mathbb{R}^{n_h\times K \times n_b}$ and rotation matrix $R^n_{\theta}$. The non commutativity forces us to compute path-integration sequentially and use a parallel scan to speed-up computation, since our models cannot leverage the parallel processing abilities of Transformers anymore, making them analogous to TEM-t \cite{whittington2022relatingtransformersmodelsneural}, where path integration is computed sequentially before computing attention.

Moreover, because of non-commutativity, the formulation of eq.\ref{eq:transssm} that allowed us to prove the link between RoPE/MapWM and SSMs doesn't hold anymore, since attention cannot be reduced to a bilinear operation:
\begin{equation}
    a_{ij} = q_i^\top R^n_{j\to i}k_j \neq \langle R^n_{\theta_{0\to i}}q_i, R^n_{\theta_{0\to j}}k_j \rangle
\end{equation}

This means that for WM models, only the SSM formulation can still be used, requiring a custom parallel scan that was beyond the scope of this paper. Since these experiments were designed as proof of concepts, we restricted ourselves to TEM-t like models to prove the further improvements that can be made.

\subsection{Non-Commutative \textit{MapFormers} Learn to Navigate Family-Trees}

\begin{table}[h!]
    \centering
    \begin{tabular}{l c c} 
        \toprule
        \textbf{} & \textbf{IID} & \textbf{OOD sparse} \\
        \midrule
        RoPE & $0.49_{\pm0.03}$ & $0.40_{\pm0.04}$ \\
        MapWM & $0.69_{\pm0.05}$ & $0.58_{\pm0.04}$ \\
        MapEM & $0.72_{\pm0.04}$ & $0.61_{\pm0.03}$ \\
        \midrule
        MapEM-NC-3 & $0.81_{\pm0.02}$ & $0.73_{\pm0.01}$ \\
        MapEM-NC-4 & $0.88_{\pm0.01}$ & $0.80_{\pm0.01}$ \\
        MapEM-NC-6 & $0.97_{\pm0.01}$ & $0.85_{\pm0.00}$ \\
        MapEM-NC-8 & $\textbf{1.00}_{\pm0.00}$ & $\textbf{0.91}_{\pm0.02}$ \\
        \bottomrule
    \end{tabular}
    \vspace{5pt}
    \caption{\textbf{Family tree navigation accuracy.} IID sequences of length $L=64$ (32 observations and 32 actions), $G=5$ generations, $p_{\mathrm{out}}=0.3$, $F=3$ founding families, and $K=10$ names. OOD: $L=128$, $G=5$, $G=10$, $p_{\mathrm{out}}=0.3$, $K=10$. RoPE performs worth than commutative \textit{MapFormers}, than can still infer commutative relations like older and younger sibling. Since the graph is non-commutative it requires a higher number of generators to learn the map: $K=n(n-1)/2$.}
    \vspace{-0.5cm}
    \label{tab:family}
\end{table}

To test whether our formalism allows to learn non-commutative, \textit{generic} cognitive maps without supervision, we constructed a synthetic environment in which an agent navigates family trees by following relations. Crucially, this dataset isn't embedded in an Euclidean space and is purely relational and non-commutative:  $\mathrm{father}\circ\mathrm{mother}\neq \mathrm{mother}\circ \mathrm{father}$.

We initialize generation 0 with $F$ founding couples, each consisting of one male and one female with no parents. For each subsequent generation $g=1, \dots,G$, every couple in generation $g-1$ produces a number of offspring drawn from $\{1, \dots, K\}$, with sex assigned randomly. Offsprings within generation $g$ are then paired into couples and with a probability $p_{\mathrm{out}}$ marries a newly introduced individual with no parents. Otherwise, we pair a man and a woman from the previous generation.

To navigate the graph, the agent can choose from the following set of actions: $\{\textit{mother}, \textit{father}, \textit{spouse}, \textit{older sib}, \textit{younger sib}, \textit{oldest child}\}$. This set of actions is defined to remain deterministic—the relation \textit{child} is non-determenistic in general—but remains entirely navigable via transitivity. Moreover, when sampling the trajectories, we only sample actions that are possible within the tree (we don't sample the action \textit{mother} when arriving at a node with no parents, like when hitting a wall on a grid).

Like in our grid navigation task, we alternate by sampling observations and actions, and compute accuracy when coming back to a previously visited location. We train our models on sequences of length $L=64$ (32 observations and 32 actions), $G=5$ generations, $p_{\mathrm{out}}=0.3$, $F=3$ founding families and evaluate on sequences of $L=128$, $G=5$, $G=10$, $p_{\mathrm{out}}=0.3$.

Since there are 7 actions to choose from, we want to ensure that our number of generators $K=n(n-1)/2$ matches this number. Therefor, along our RoPE and commutative baselines, we evaluate four \textit{Map}EM-NC-$n$ models, where $n=3, 4, 6, 8$ designate the dimension of $\mathrm{SO}(n)$, with corresponding block size. We train all models with a single layer, a single head of $d=192$ neurons (which is divisible by all block sizes $n$).

As we can see in tab.~\ref{tab:family}, all MapFormer models significantly outperform RoPE, but only our non-commutative models manage to reach a perfect accuracy and strong OOD generalization. Even though commutative \textit{MapFormers} are unable to solve the task, they still outperform RoPE. This can be explained by the fact that some relations within the latent structure are commutative (\textit{e.g.} "younger sibling" and "older sibling"), implying that a commutative bias can still help by partly modeling the underlying structure.

These numbers also demonstrate the inherent tradeoff between the number of generators (hence increasing the block size, which increases the cost of matrix exponential and matrix-vector products) and computational efficiency.

\end{document}